gsave
newpath
  20 20 moveto
  20 220 lineto
  220 220 linetop
  220 20 lineto
closepath
2 setlinewidth
gsave
  .4 setgray fill
grestore
stroke
grestore
\RequirePackage{fix-cm}
\documentclass{svjour3}                     
\smartqed  
\usepackage{graphicx}
\usepackage{textcomp}
\usepackage{cite}
\usepackage{amsmath,amssymb,amsfonts}
\usepackage{algorithmic}
\usepackage{graphicx}
\usepackage{caption}
\usepackage{textcomp}
\usepackage{color}
\usepackage[dvipsnames]{xcolor}
\usepackage{subfigure}
\usepackage{multirow}
\usepackage{rotating}
\usepackage{diagbox}
\usepackage{stfloats}
\usepackage[backref]{hyperref}
\usepackage{float}
\usepackage{bm}
\usepackage{booktabs}
\usepackage{float}
\usepackage{multirow}
\usepackage{longtable}
\usepackage{lscape}
\usepackage{array}
\begin{document}

\title{A Comprehensive Review of Computer-aided Whole-slide Image Analysis: 
from Datasets to Feature Extraction, Segmentation, Classification and Detection Approaches}


\author{Chen Li         \and
        Xintong Li  \and
        Md Rahaman \and
        Xiaoyan Li \and
        Hongzan Sun \and
        Hong Zhang \and
        Yong Zhang \and
        Xiaoqi Li \and
        Jian Wu \and
        Yudong Yao  \and 
        Marcin Grzegorzek      
}


\institute{Chen Li, Xintong Li, Md Rahaman, Xiaoqi Li and Jian Wu \at
              Microscopic Image and Medical Image Analysis Group, 
              College of Medicine and Biological Information Engineering, 
              Northeastern University, 110169, Shenyang, PR China \\
              Chen Li \email{lichen201096@hotmail.com}           
            \and
            Xiaoyan Li, Hongzan Sun, Hong Zhang and Yong Zhang \at
              China Medical University, 110122, Shenyang, China 
            \and
              Yudong Yao \at
              Department of Electrical and Computer Engineering, 
              Stevens Institute of Technology, Hoboken, NJ 07030, USA
            \and
              Marcin Grzegorzek \at
              Institute of Medical Informatics, University of Luebeck, 
              Luebeck, Germany
}

\date{Received: date / Accepted: date}

\maketitle

\begin{abstract}
With the development of computer-aided diagnosis (CAD) and image scanning technology, 
\emph{Whole-slide Image} (WSI) scanners are widely used in the field of pathological diagnosis. 
Therefore, WSI analysis has become the key to modern digital pathology. Since 2004, WSI has 
been used more and more in CAD. Since machine vision methods are usually based on semi-automatic 
or fully automatic computers, they are highly efficient and labor-saving. The combination of WSI and CAD technologies for segmentation, classification, and detection helps 
histopathologists obtain more stable and quantitative analysis results, save labor costs and improve diagnosis objectivity. This paper reviews the methods of WSI analysis based on 
machine learning. Firstly, the development status of WSI and CAD methods are introduced. Secondly, we discuss 
publicly available WSI datasets and evaluation metrics for segmentation, classification, and 
detection tasks. Then, the latest development of machine learning in WSI segmentation, classification, 
and detection are reviewed continuously. Finally, the existing methods are studied, the applicability 
of the analysis methods are analyzed, and the application prospects of the analysis methods in this 
field are forecasted.

\end{abstract}

\section{Introduction}
\label{s:int} 
\subsection{Brief Knowledge of Whole-slide Imaging Technique}
\label{ss:int:Purposes}
\emph{Whole-slide Image} (WSI), also generally mention as ``virtual microscopy'', 
purposes to imitate typical light microscopy in a computer-generated 
model~\cite{Farahani-2015-WSIP}. People usually think of whole-slide imaging as 
an image acquisition method. It is possible to transform the whole glass slide into a digital form~\cite{Janabi-2012-WSIA}. Furthermore, the 
``digital slides'' are used for humans observation or performing them to 
automated image analysis~\cite{Pantanowitz-2011-RTCS}.

The processing of whole-slide imaging is performed by the WSI system. A WSI 
system has a scanner, networked computer(s), and possibly a server or cloud 
solution for storage, display (e.g. tablet, etc.), and compatible software for 
image creation, management, and 
analysis~\cite{gabry-2014-WSIW}~\cite{saco-2016-CSWS}~\cite{pantanowitz-2013-VWSI}. 
The first part applies technical hardware (scanner) to digitize glass slides, 
generates a sizable classical digital image (so-called ``digital slide'') accordingly. 
The second part exploits technical software (ie, virtual slide viewer) to view and/or 
analyze these huge digital images~\cite{Weinstein-2009-OTVM}. WSI devices have 
different looks and performance, but overall, the WSI scanner includes the following 
parts: an optical microscope system with a camera, an acquisition system, computer 
hardwares/softwares, scanning softwares, and a digital slide viewer. Supplemental 
components include the feeder or image processing systems~\cite{Farahani-2015-WSIP}.
\begin{figure}[htbp!]
\centerline{\includegraphics[width=0.55\linewidth]{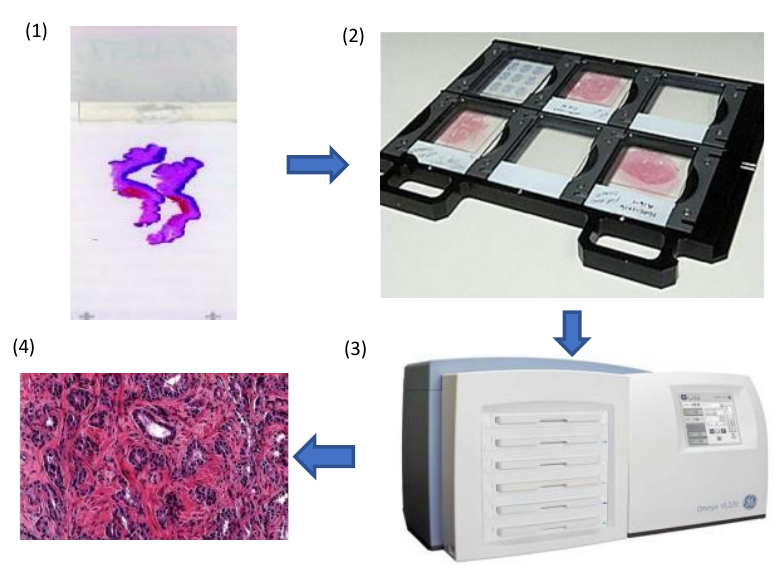}}
\caption{The workflow of the proposed whole-slide imaging. 
(1) is the glass slide~\cite{Amin-2008-AWSI}. 
(2) whole mount glass slides. 
(3) whole slide imaging scanner. 
(4) digital whole  image~\cite{Farahani-2015-WSIP}. }
\label{fig:equip}
\end{figure}

As shown in Fig.~\ref{fig:equip}, (1) is the pathologic biopsy, (2) is the whole 
mount glass slides, (3) is whole slide imaging scanner, (4) is the obtained WSIs. 
The optical microscope system is the essential part of the WSI scanner, especially 
the lens optics and the camera because it can determine the quality of the images. 
Charged coupled device (CCD) sensors on cameras that can convert analog signals into 
digital signals. There are two major methods of slide acquisition. One is area scanning, 
the other is line scanning. The area scanner moves on the sample block by block and 
section by section, that is, after stopping at each position to capture an image, 
it is repositioned to the next position. The line scanner is smooth and continuous 
movement and fast scanning~\cite{Higgins-2015-ACDP}. After choosing a range of interest 
on a slide, adjust focus, and scan the slide~\cite{Amin-2008-AWSI}. If WSI scanners 
have a Z-stacking facility (scan slides at different focal planes along the vertical 
Z-axis and stack images on top of each other to produce composite multi-planar 
images~\cite{Farahani-2015-WSIP}), they can better center on particular areas of 
interest~\cite{gabry-2014-WSIW}. Owing to the images generated by the WSI systems 
are large, the visual field of a computer should be bigger than the visual field of 
a traditional microscope over four times~\cite{Rojo-2006-CCCA}.

With the accelerating development of science and technology, the WSI system has 
progressed rapidly. WSI offers higher quality and resolution images with 
annotation~\cite{Pantanowitz-2011-RTCS}. The scanner with fast scanning speed has 
improved image quality and reduced storage costs~\cite{Janabi-2012-WSIA}. The digital 
approach also can reduce the time of transporting glass slides and the risk of breakage 
and fading~\cite{Ghaznavi-2013-DIPW}~\cite{Camparo-2012UWSI}~\cite{Webster-2014-WSIA}. 
Moreover, the digital slides do not deteriorate over time~\cite{saco-2016-CSWS}.

WSI infuses into many fields such as E-education, virtual workshops, and pathology aspect. 
Now, there is a growing need for pathology to improve quality, patient safety, and 
diagnostic accuracy. These causes and economic pressures to consolidate and centralize 
diagnostic services~\cite{Ghaznavi-2013-DIPW}. Moreover, WSI can boost distinct pathology 
practices, so it is generally used in pathology~\cite{Cornish-2012-WSIR}. Digital pathology 
networks based on WSI systems can solve some difficult problems with pathology. For example, 
WSI can be explored  by several observers from different areas at the same time. 
Discussions using WSI can save the time needed for transferring glass slides to distant 
places for attaining second minds and 
teleconsultation~\cite{Janabi-2012-WSIA}~\cite{Webster-2014-WSIA}~\cite{Janabi-2012-WSIF}. 
WSI equivalently broadens the scope of cytopathology where virtual slides are used for 
numerous intents like telecytology, quality activities (e.g. archiving and proficiency testing), 
and education (e.g. virtual atlases)~\cite{gabry-2014-WSIW}. It will also let pathologists 
become more efficient, precise, and creative at quantifying prognostic biomarkers like 
HER2/neu (c-erbB-2). But also, crucially, WSI develops CAD in combination with the 
continually developing computer artificial intelligence, big data, and cloud technology. 
Nowadays, WSI technology is very advance and offers the pathology community novel clinical, 
nonclinical, and research image-related applications~\cite{Farahani-2015-WSIP}.

\subsection{The Development of WSI Analysis}
\label{ss:int:current development}
The traditional pathological section analysis method requires specially trained pathologists 
to look for areas of interest under the microscope one by one, and then analyze and diagnose 
based on professional knowledge. Traditional manual analysis of pathological images has many 
drawbacks and problems. There are no quantitative indicators, so the qualitative analysis results cannot be reproduced~\cite{Singh-2010-BCDC}. Moreover, most doctors have tight 
working conditions, heavy workload, and time pressure. In this case, the human cognitive 
process is easily disturbed, leading to incomplete diagnosis and 
misdiagnosis~\cite{Goggin-2007-CDSS}. Although traditional slide analysis is accurate, it can 
be deeply personal. It is available for the same person to evaluate a slide one day and to 
get different conclusions the following week. Besides, the procedure is a challenging and 
time-consuming task~\cite{Higgins-2015-ACDP}. Therefore, CAD is a more efficient, accurate, 
and intuitive method.

The computer-aided reading slide can help pathologists improve diagnosis accuracy and detection 
rate and reduce the overall misdiagnosis rate. Moreover, the computer is not affected by fatigue 
and human error and provides better assistance to doctors~\cite{Goggin-2007-CDSS}~\cite{Fazal-2018-PPFR}. 
It is also a valuable tool to reduce the workload of clinicians~\cite{Lisboa-2006-UANN}. While reducing pathologists' workload and improving efficiency, it can also perform intuitive quantitative analysis of pathological conditions. These are better than manual reading slides. 
Computer-aided viewing of WSI is now rapidly developing. WSI provides the pathology field unique 
clinical, nonclinical, and analysis of image-related applications~\cite{Farahani-2015-WSIP}.

In recent years, the pathological WSI analysis performed by CAD doctors, it has been widely 
used in different cancer fields (ie, breast cancer, prostate cancer, gastric cancer, neuroblastoma). 
The scope of applications focuses on disease classification, early screening, tissue localization, 
and benign and malignant diagnosis. Common tasks with CAD include classification, segmentation, 
and detection.

\begin{figure}[htbp!]
\centerline{\includegraphics[width=0.98\linewidth]{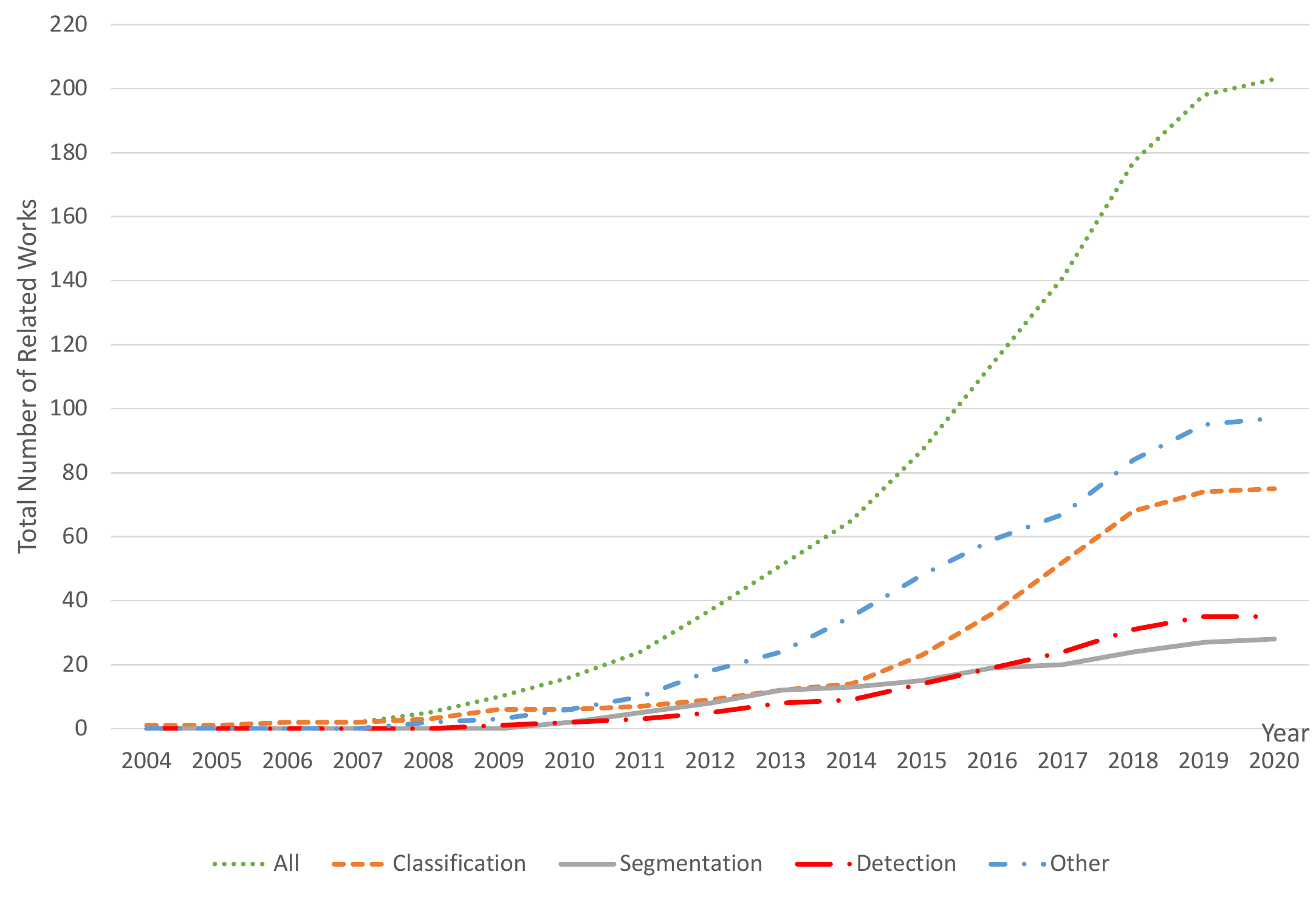}}
\caption{The development trend of computer-aided viewing of WSIs. 
The horizontal direction shows the time. 
The vertical direction shows the cumulative number of related works in each year. }
\label{fig:trend}
\end{figure}

For example, in the work of~\cite{Huang-2017-AHGP}, automatic 
detection and sequencing system based on Gleason pattern recognition is proposed for the automatic 
detection of high-grade prostate cancer. In the field of breast cancer, the work 
of~\cite{Mehta-2018-LSBB} makes the segmentation of WSI images of breast biopsy with biologically 
significant tissue markers. The study of~\cite{Korbar-2017-DLCC} trains a modified version of 
the residual network (ResNet) to classify different types of colorectal polyps on WSIs. 
At present, the development trend of computer-aided viewing of WSIs is shown in Fig.~\ref{fig:trend}. 

As shown in Fig.~\ref{fig:trend}, as the years are getting closer and closer, technology 
continues to advance. There are more and more cases of using computers to assist diagnosis. 
The number of cases in the three main applications of classification, segmentation, and 
detection has increased year by year. The number of cases in other applications 
are growing, such as retrieval~\cite{Ma-2016-BHIR}, localization~\cite{Alomari-2009-LTHR}. 
Beginning in 2008, CAD viewing WSI has helped pathologists begin to realize it significantly. 
Since 2014, there has been an increasing trend in the number of computer-aided pathologists 
diagnosed with WSI. Gradually by 2020, the growth rate of CAD has increased, reflecting the 
vigorous development of this technology.

\begin{figure*}[htbp!]
\centerline{\includegraphics[width=0.98\linewidth]{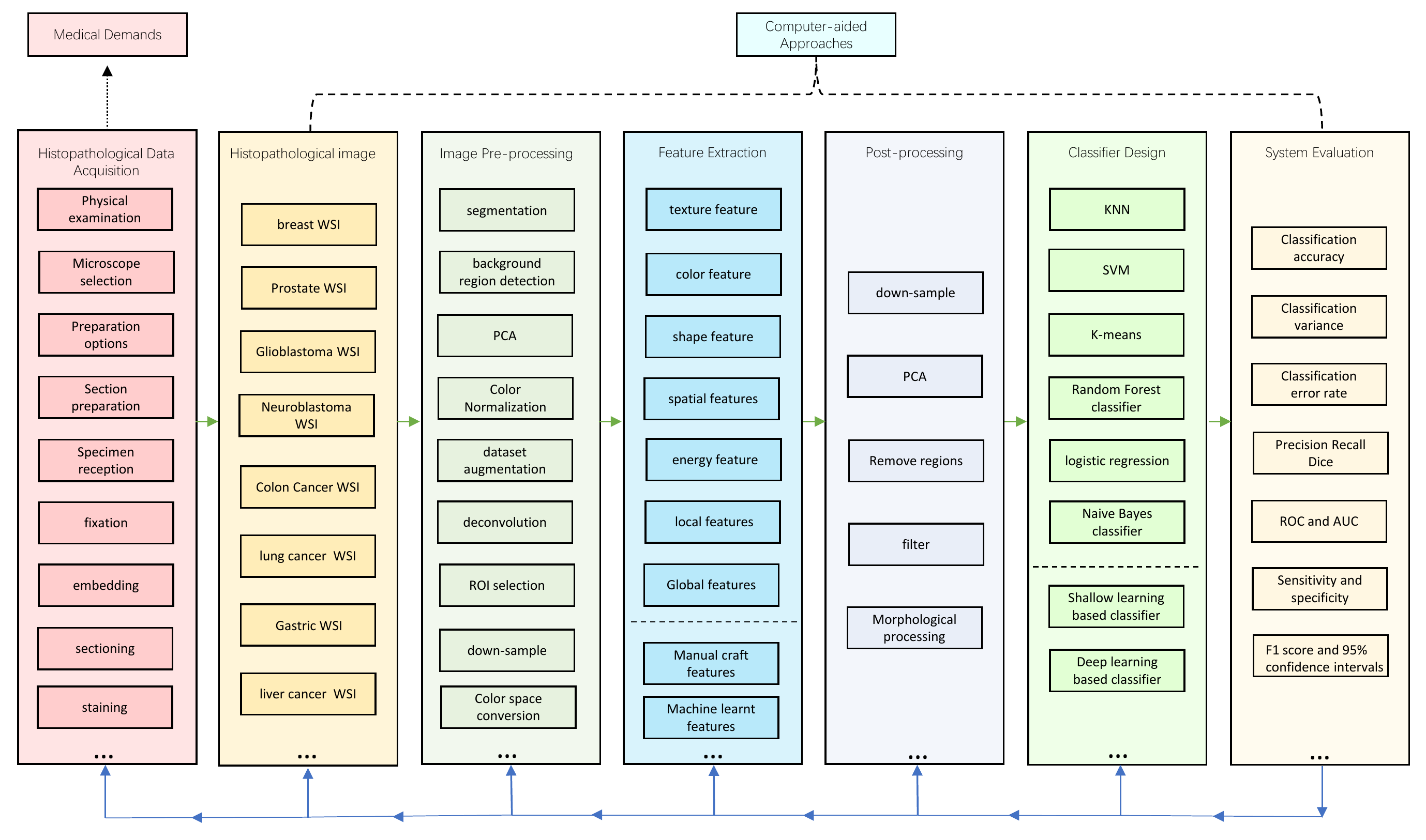}}
\caption{The organization chart of histopathology WSI analysis using computer-aided analysis 
approaches in this paper. }
\label{fig:flow}
\end{figure*}

Besides, to explain and clarify computer-aided pathologists' work context in viewing WSI, an organization chart is shown in  Fig.~\ref{fig:flow}. The figure shows the general process 
of CAD and processing WSI. It shows seven important steps in the histopathology image analysis 
system, including data acquisition, image presentation, image preprocessing, feature extraction, 
data post-processing, classifier design, and system evaluation. 

In Fig.~\ref{fig:flow}, histopathological data initially obtained from the 
medical field, 2-D or 3-D digital microscopic images are first captured by various imaging 
equipment (e.g., optical microscopy), and then saved in a specific color space (e.g., Red, 
Green and Blue (RGB) color space). The 3rd step is the  image pre-processing step, the 
properties of images are improved by dataset augmentation, segmentation, and so on, which 
is an importation preparation for the feature extraction step. In the next step, feature 
extraction is implemented, where the image can be represented by its attributes (shape, 
texture, and color features), or layout features (global and local features), or extraction 
style (manual or automatic). These feature extraction categories are not separate, but can 
be converted into other categories using appropriate methods. After that, the post-processing 
step takes the responsibility to enhance the extracted features, where filter, morphological 
processing, normalization are always used. Also, the classifier can be classified 
as shallow or deep according to its learning structure. Finally, various numerical and 
intuitive methods are used to evaluate the classification system, such as classification 
accuracy, classification error rate, sensitivity, and specificity. Besides, each 
step is not independent, but is closely connected with other steps through information 
feedback. Therefore, the entire CAD viewing WSI system is an organic whole~\cite{C.Li-2016-CBMI}.

\subsection{Motivation of This Review}
\label{ss:int:motivation}
Now WSI technology has applications in many fields. For example, to perform preliminary diagnosis 
of surgical pathology, and perform intraoperative frozen section diagnosis through remote 
consultation~\cite{huang-2018-TCFS}~\cite{boyce-2017-AUVW}, and seek expert advice without 
incurring international transportation costs or delays~\cite{Boroujeni-2016-WSIH}. WSI also 
provides advantages in tumor diagnosis, prognosis, and targeted therapy. It can also facilitate 
teachers and students in teaching~\cite{Ghaznavi-2013-DIPW}. Therefore, the research field of WSI analysis through CAD systems is significant. To the best of our knowledge, there exist some survey papers that summarize WSI analysis (e.g., the reviews 
in~\cite{Pantanowitz-2011-RTCS}, ~\cite{Gurcan-2009-HIAA,Kothari-2013-PIIQ,Veta-2014-BCHI,Sharma-2015-ARGB,Li-2018-LSRM,Komura-2018-MLMH,Chang-2019-AIIP,Nichols-2019-MLAA,Wang-2019-PIAU,Dimitriou-2019-DLWS,Kumar-2020-WSIP}). 
In the following part, the summary of survey papers related to the WSI analysis is presented.

The survey of~\cite{Pantanowitz-2011-RTCS} reviews the current status of WSI pathology, 
including supervision and verification, remote and routine pathological diagnosis, educational 
use, implementation issues, and cost-benefit analysis of WSI in routine clinical practice. However, 
this article only focuses on the application of CAD systems in WSI analysis. This review rarely 
mentions this, and only 12 references are about WSI.

The survey of~\cite{Gurcan-2009-HIAA} reviews the latest CAD techniques for 
digital histopathology. This article also briefly introduces new image analysis technologies 
developed and applied in the United States and Europe for some specific histopathology-related 
problems. More than 130 papers on CAD have been summarized and only three articles are about WSI.

The survey of~\cite{Kothari-2013-PIIQ} reviews the WSI informatics method of histopathology, 
related challenges and future research opportunities. However, this article reviews image quality 
control, feature extraction of image attributes captured at pixels, object, and semantic levels, 
image features for predictive modeling, and data and information visualization for diagnostic 
or predictive applications. It does not discuss the entire process of CAD and the viewing of WSI. 
More than 130 papers have been summarized. However, only three articles are about WSI.

The survey of~\cite{Veta-2014-BCHI} reviews the analysis methods of histopathological images 
of breast cancer, introduces the process of tissue preparation, staining, and slide digitization, 
and then discusses different image processing techniques and applications, from tissue staining 
analysis to CAD, and the prognosis of breast cancer patients. Although the histopathological 
images discussed in the article are WSIs, they are only about breast cancer and not comprehensive. 
More than 110 papers have been summarized. However, only four articles are about WSI.

The survey of~\cite{Sharma-2015-ARGB} provides a comprehensive overview of the graph-based 
methods explored so far in digital histopathology. More than 170 papers have been summarized. 
However, only four articles are about WSI.

The survey of~\cite{Li-2018-LSRM} reviews the latest methods of large-scale medical image 
analysis, which are mainly based on computer vision, machine learning, and information retrieval. 
Then, they comprehensively reviewed the algorithms and technologies related to the main 
processes in the pipeline, including feature representation, feature indexing, and search. 
However, WSI appears only in the sample dataset, and no actual analysis is performed. Of the more 
than 250 papers summarized in this paper, only three mention WSI.

The survey of~\cite{Komura-2018-MLMH} introduces the application of digital pathological image 
analysis using machine learning algorithms, solve some specific analysis problems, and propose 
possible solutions. However, there are only 11 articles related to WSI on the topics we are interested 
in. More than 120 papers have been summarized. But only 11 articles are about WSI.

The survey of~\cite{Chang-2019-AIIP} introduces the general situation of artificial intelligence, 
a brief history in the medical field and the latest developments in pathology, and the future 
prospects of pathology driven by it. This review only briefly mentions WSI in the part of the 
pathology application imaging and example datasets. Of the more than 70 papers summarized in 
this paper, only four mention WSI.

The survey of~\cite{Kumar-2020-WSIP}  introduces the technical aspects of WSI, its application 
in diagnostic pathology, training and research, and its prospects. It highlights the benefits, 
limitations, and challenges of delaying the use of this technology in daily practice. But this 
article only focuses on computer-aided pathologists to view WSI and its application in diagnosis, 
which are not discussed in this review. Of the 50 references, 20 are about WSI.

From the existing review papers mentioned above, we can find that many researchers are concerned 
about the current status and development trend of WSI technology itself, and hundreds of related 
works have been systematically summarized and discussed in those review papers. However, all 
these survey papers use WSI format datasets as examples only, and do not aim to introduce the 
detailed introduction of computer-aided pathologists to review WSI technology. Therefore, we present this review paper to analyze all related works using CAD combined with WSI in the past few decades. This survey summarizes more than 210 related 
works from 2004 to 2020. The audience for this review is related researchers in the field of medical 
imaging and medical professionals.

\subsection{Structure of This Review}
\label{ss:int:structure}

This structure of this paper is as follows: Sec. ~\ref{s:DEM} summarizes the related datasets and commonly used evaluation methods. Sec. ~\ref{s:FE} illustrates frequently used feature extraction methods. Sec. ~\ref{s:SM}, ~\ref{s:CM}, and ~\ref{ss:int:DM} present the related work of segmentation, classification, and detection using WSI and CAD technology. After the overview of different works, the most frequently used approaches are analyzed in Sec.~\ref{s:MA}. Finally, Sec.~\ref{s:CFW} concludes this review with prospective future direction.

\section{Datasets and Evaluation Methods}
\label{s:DEM} 
In this section, we have discussed some commonly used datasets and evaluation metrics for the 
classification, segmentation, and detection tasks.

\subsection{Publicly Available Datasets about WSI}
\label{ss:int:data}
To better analyze the CAD using WSI technology, we have summarized some frequently used publicly available datasets in our study. Tab.~\ref{DB} shows the necessary information of these datasets. Two of the most commonly used datasets are The Cancer Genome Atlas (TCGA)~\cite{TCGA} and the 
Camelyon datasets~\cite{Litjens-2018-1HSS}. Both datasets are often used for classification 
and detection. At the same time, we find two WSI datasets named TUPAC16~\cite{Veta-2019-PBTP} 
and Kimia Path24~\cite{Babaie-2017-CRDP}. TUPAC dataset is widely used, such as mitosis 
detection, prediction of breast tumor proliferation, automatic scoring (classification), 
and so on. Kimia Path24 is often used for classification and retrieval~\cite{Babaie-2017-CRDP}~\cite{Kumar-2018-DBFR}. The basic information 
of the common datasets are shown in Table.~\ref{DB}.
\begin{table*}[htbp!]
\renewcommand\arraystretch{2}
\setlength{\tabcolsep}{1 pt}
\centering
\scriptsize\caption{The basic information of the publicly available used datasets.}
\begin{tabular}{clcc}
\hline
\textbf{Databases}    & \multicolumn{1}{c}{\textbf{Year}} & \textbf{Field}   & \textbf{Number of images or or size} \\ \hline
TCGA                  & 2006                              & Cancer related   & Over 470 TB                          \\
NLST Pathology Images & 2009                              & Lung             & Around 1250 H\&E slides              \\
BreakHis              & 2015                              & Breast cancer    & 9,109 microscopic images             \\
TUPAC16               & 2016                              & Tumor mitosis    & Around 821 H\&E slides               \\
Camelyon              & 2017                              & Breast cancer    & Around 3TB                           \\
Kimia Path24          & 2017                              & Pathology Images & 24 WSIs                              \\ \hline
\end{tabular}
\label{DB}
\end{table*}

\subsubsection{TCGA database} 
\begin{figure}[htbp!]
\centerline{\includegraphics[width=0.98\linewidth]{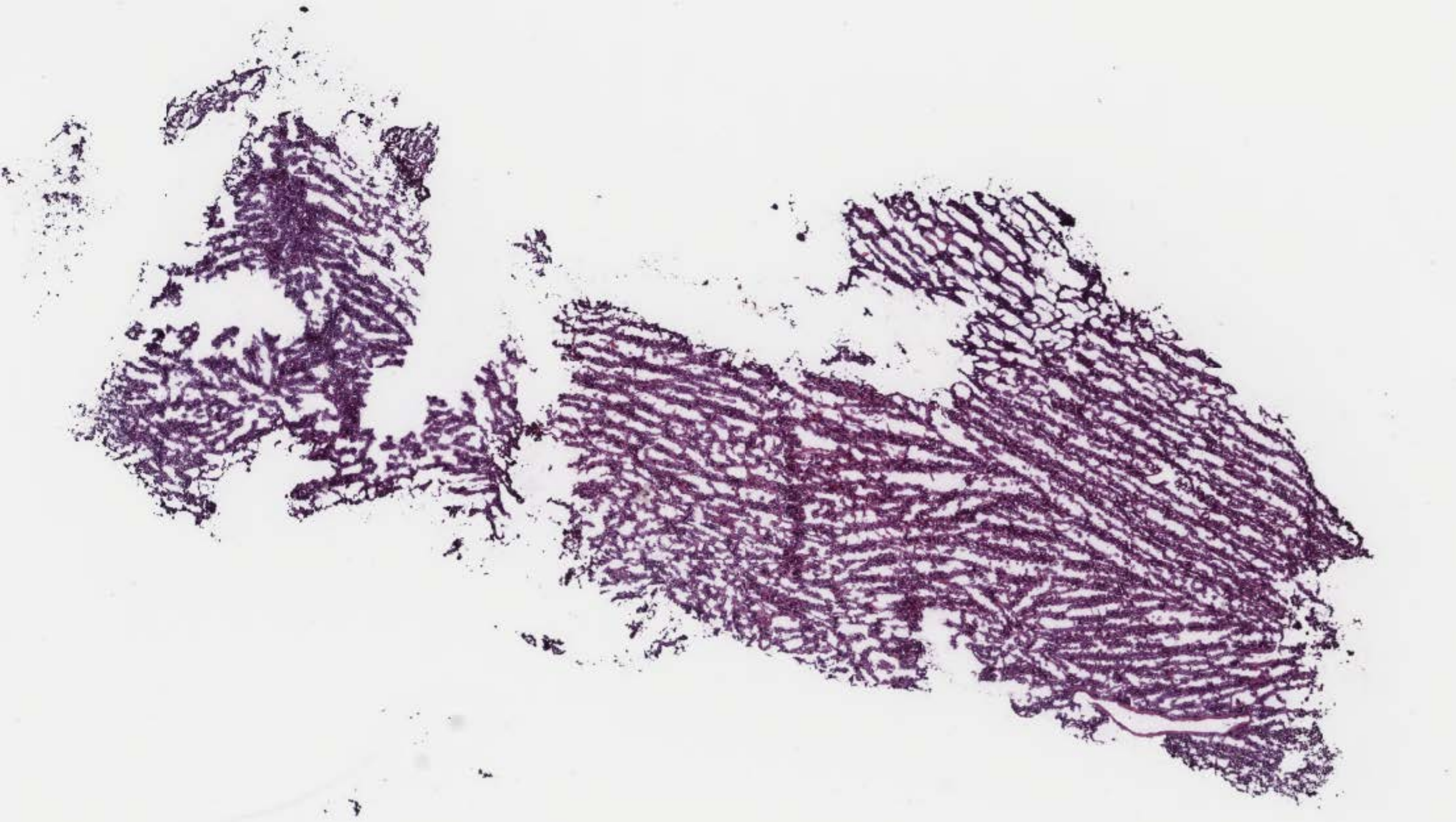}}
\caption{An example of WSI in an adrenal cortical carcinoma in the TCGA database~\cite{TCGA}.}
\label{fig:fig4}
\end{figure}

TCGA is a project jointly launched by The National Cancer Institute (NCI) and the 
National Human Genome Research Institute (NHGRI) in 2006~\cite{TCGA}. It contains 
clinical data, genome variation, mRNA expression, miRNA expression, methylation and 
other data of various human cancers (including subtypes of tumors). The database is designed to use high-throughput 
genomic analysis techniques to help people developing a better understanding of cancer 
and improve the ability to prevent, diagnose, and treatment~\cite{Tomczak-2015-CGAI}. 
While TCGA main work focuses on genomics and clinical data, it also accumulates a large 
number of WSIs in patient's tissue. Since WSI datasets are much larger than other datasets, to 
facilitate viewing, David et al.~\cite{Gutman-2013-CDSA} proposes an integrated network 
platform named Cancer Digital Slide Archive (CDSA) to accommodate all WSI in TCGA. 
Since the dataset contains many types of cancers, it has a wide range of uses. 
Fig.~\ref{fig:fig4} below is an example of WSIs in an adrenal cortical carcinoma 
in the TCGA database.

\subsubsection{Camelyon Database} 
\begin{figure}[htbp!]
\centerline{\includegraphics[width=0.98\linewidth]{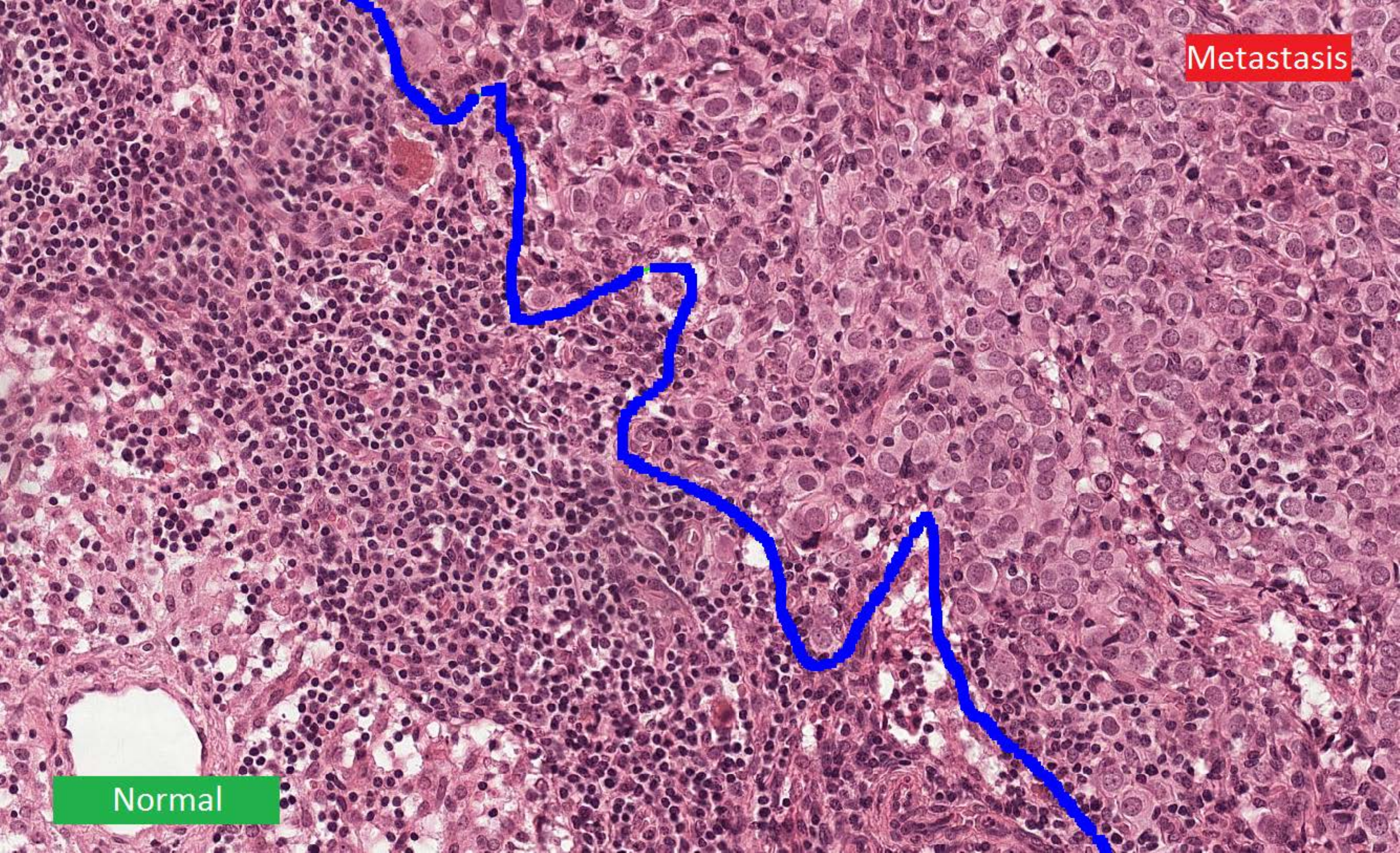}}
\caption{A pathological image of a lymph node in Camelyon~\cite{Litjens-2018-1HSS}. 
The left side belongs to normal cell tissue, 
and the right cell has been swallowed and occupied by cancer cells.}
\label{fig:fig5}
\end{figure}

The Camelyon Challenge is hosted by International Symposium on Biomedical Imaging 
(ISBI)~\cite{Litjens-2018-1HSS}. The whole competition dataset (Camelyon16, Camelyon17~) 
are derived from sentinel lymph nodes of breast cancer patients contains WSIs of Hematoxylin 
and Eosin (H\&E) stained node sections~\cite{Bejnordi-2017-DADL}\cite{Bandi-2018-DIMC}. 
Therefore, the Camelyon dataset is suitable for the automatic detection and classification 
of breast cancer in WSI. The data of Camelyon16 are from the Radboud University Medical 
Centre and the University of Utrecht Medical Centre. The Camelyon16 dataset is composed of 
170 phase I lymph node WSIs (100 normals and 70 metastatics) and 100 Phase II WSIs (60 normals 
and 40 metastatics), and the test dataset consisted of 130 WSIs from two universities. 
The Camelyon16 dataset is used as training values for the evaluation of Camelyon17. 
Fig.~\ref{fig:fig5} is a pathological picture of a lymph node in Camelyon. The left side 
belongs to normal cell tissue, and the right cell has been swallowed and occupied by cancer cells.

\subsubsection{TUPAC16 Database}
\begin{figure}[htbp!]
\centerline{\includegraphics[width=0.98\linewidth]{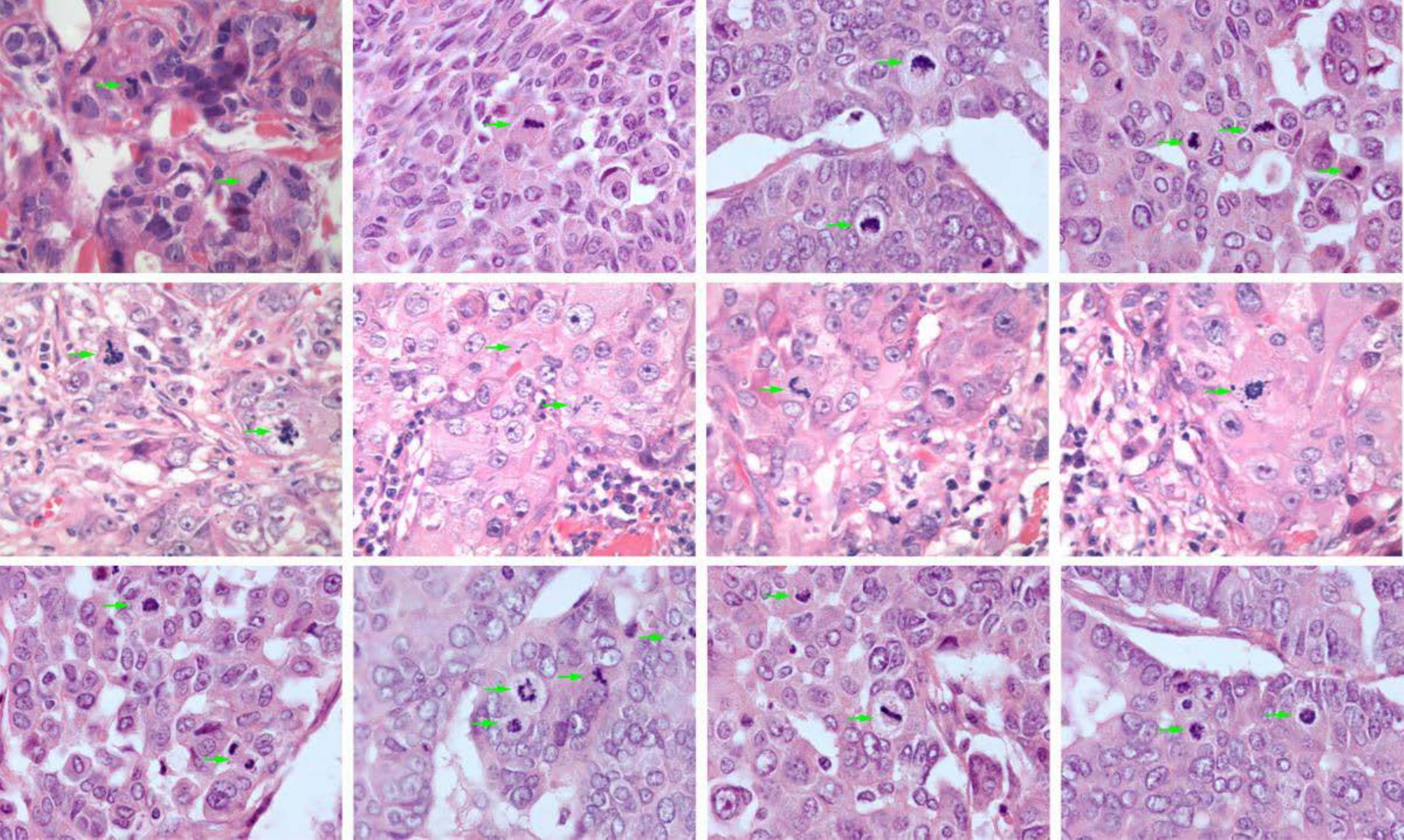}}
\caption{Some examples of mitosis diagrams in H\&E breast cancer slices in the 
TUPAC16~\cite{Veta-2019-PBTP}, with green arrows marking mitosis.}
\label{fig:fig6}
\end{figure}

The TUPAC16 challenge is held in the context of the MICCAI~\cite{Veta-2019-PBTP}. 
TUPAC16 main challenge dataset consists of 821 TCGA WSIs with two types of tumor 
proliferation data. 500 for training, 321 for testing. In addition to the main challenge 
dataset, there are two secondary datasets (area of interest and mitotic detection). 
The area of interest auxiliary dataset contains 148 cases that are randomly selected 
from the training dataset. The mitotic test dataset consisted of WSIs of 73 breast cancer 
cases from three pathological centers. Of the 73 cases, 23 are AMIDA13 
challenge~\cite{Veta-2015-AAMD}. The remaining 50 cases previously used to assess the 
interobserver agreement for mitosis counting are from two other pathology centers in the 
Netherlands. So the dataset is mainly used for automatic detection of tumor mitosis or 
other regions of interest(ROI). Fig.~\ref{fig:fig6} shows some examples of mitosis maps 
in H\&E breast cancer slices, with green arrows marking mitosis.

\subsubsection{Kimia Path24 Database}
\begin{figure}[htbp!]
\centerline{\includegraphics[width=0.65\linewidth]{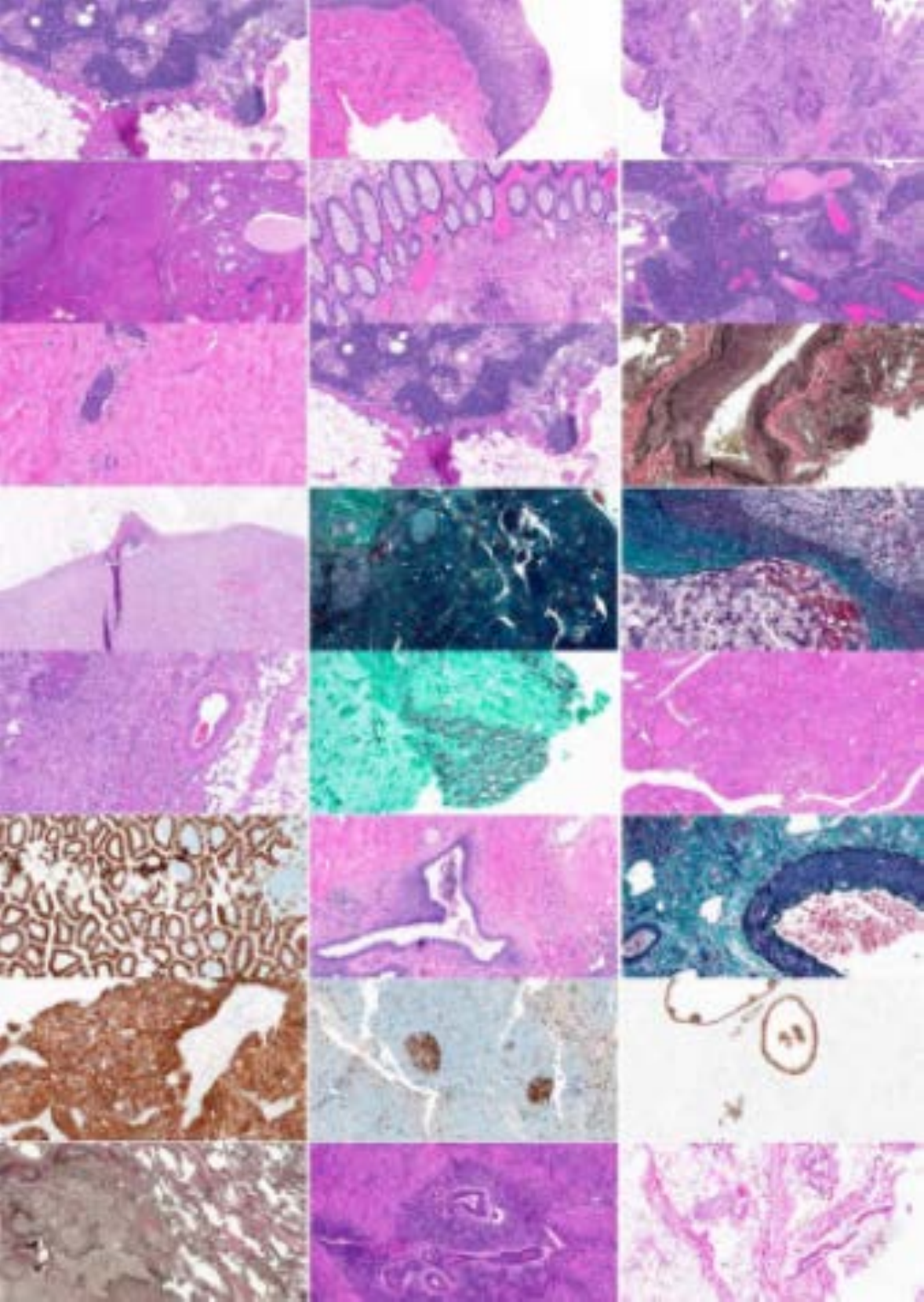}}
\caption{24 WSIs thumbnails in Kimia Path24 database~\cite{Babaie-2017-CRDP}.}
\label{fig:fig7}
\end{figure}

This dataset is consciously and manually selected from 350 WSIs from different body 
parts so that the 24 WSIs clearly represented different texture patterns. So this 
dataset is more like a computer vision dataset (as opposed to a pathology dataset) 
because visual attention is spent on the diversity of patterns rather than on anatomy 
and malignancy~\cite{Babaie-2017-CRDP}. Therefore, this dataset is mainly used for 
classification and retrieval of histopathological images. The 24 WSIs thumbnails in 
this dataset are shown in Fig.~\ref{fig:fig7}.

\subsection{Evaluation Method}
\label{ss:int:EM}
This subsection introduces the evaluation methods of classification, segmentation, 
and detection algorithms and related formulas.

\subsubsection{Basic Evaluation Indexs} 
\label{ss:int:BEI}
The confusion matrix is used to observe the performance of the model in each category, 
and the probability of each category can be calculated. The specific style of the 
confusion matrix is shown in Tab.~\ref{CM}. 
\begin{table}[htbp!]
\renewcommand\arraystretch{2}
\setlength{\tabcolsep}{1 pt}
\centering
\scriptsize\caption{Confusion matrix of basic evaluation indexs.}
\begin{tabular}{ccc}
\hline
\multicolumn{1}{l}{\textbf{Data Class}} & \multicolumn{1}{l}{\textbf{Classified as Pos}} & \multicolumn{1}{l}{\textbf{Classified as Neg}} \\ \hline
Pos                                     & True Positive(TP)                              & False Negative(FN)                             \\ \hline
Neg                                     & False positive(FP)                             & True Negative(TN)                              \\ \hline
\end{tabular}
\label{CM}
\end{table}

According to the confusion matrix, the True Positive Rate (TPR) can be defined as TP/(TP + FN), 
which represents the proportion of the actual positive instances in all positive instances of 
the positive class predicted by the classifier, the False Postive Rate (FPR) can be defined 
as FP/(FP+TN), which represents the proportion of the actual negative instances in all negative 
instances of the positive class predicted by the classifier. It can be seen that the mathematical 
expressions of the following evaluation metrics are shown in Table.~\ref{EM}~\cite{Sokolova-2009-SAPM}.
\begin{table}[htbp!]
\renewcommand\arraystretch{2}
\setlength{\tabcolsep}{1 pt}
\centering
\scriptsize\caption{Evaluation metrics. \emph{Acc}, \emph{P}, \emph{R}, \emph{Se}, \emph{Sp} and \emph{F1} denote accuracy, precision, recall, 
sensitivity, specificity and F1 score, respectively.}
\begin{tabular}{cc|cc}
\hline
Assessments & Formula   &  Assessments & Formula                          \\ \hline
\emph{Acc}    & $\frac{TP + TN}{TP + TN + FP + FN}$  &  \emph{Se}	& $\frac{TP}{TP + FN}$ \\ 
\emph{P}   & $\frac{TP}{TP + FP}$   &   \emph{Sp}	& $\frac{TN}{TN + FP}$           \\
\emph{R}      & $\frac{TP}{TP + FN}$   &   \emph{F1}	&$2×\frac{P×R}{P + R}$         \\ \hline
\end{tabular}
\label{EM}
\end{table}

\subsubsection{Evaluation of Segmentation Methods} 
Image segmentation~\cite{Haralick-1985-IST} is the segmentation of images with existing 
targets and precise boundaries. The commonly used indicators are accuracy, precision, 
recall, F-measure, sensitivity, and specificity. These metrics we have discussed in 
Sec.~\ref{ss:int:BEI} and their mathematical expressions are given Tab.~\ref{CM}. 
Dice co-efficient (D) and Jaccard index (J) are popular segmentation evaluation indexes in 
recent years. Dice co-efficient (D) represents the ratio of the area intersected by two 
individuals to the total area, that is, the similarity between ground truth and the 
segmentation result graph. If the segmentation is perfect, the value is 1. Then, if S 
stands for the segmentation result graph and G stands for ground truth, the expression 
of Dice co-efficient (D) is given in Eq.~(\ref{eq:Dice}).
\begin{equation}
D(S,G)=\frac{2|A \cap G|}{|A| + |G|}
\label{eq:Dice}
\end{equation}

Jaccard Index (J) represents the intersection ratio of two individuals, which is similar 
to Dice co-efficient. The formula is given in Eq.~(\ref{eq:Jaccard}).
\begin{equation}
J(S,G)=\frac{|A \cap G|}{|A \cup G|}
\label{eq:Jaccard}
\end{equation}

\subsubsection{Evaluation of Classification Methods} 
\label{ss:int:EMC}
Classification~\cite{Kamavisdar-2013SICA} is the operation of determining the properties 
of objects in the image. In the field of digital histopathology we studied, some are the 
classification of cancer~\cite{Petushi-2006-LSCH}, some are the operation of selecting 
ROI~\cite{Swiderska-2015-TMMH}, and some are the identification of cancer 
regions~\cite{Doyle-2010-BBMC}. The purpose of classification is achieved by the constructed 
classifier. The performance indicators used to evaluate these classifiers are critical to 
the final results. Accuracy is the most commonly used indicators to evaluate classifiers. 
Precision, recall, sensitivity, specificity, and F1 score are widely used to evaluate 
classifiers. Accuracy,precision, recall, F-measure, sensitivity, and specificity we have 
discussed in Sec.~\ref{ss:int:BEI} and their mathematical expressions are given in Tab.~\ref{CM}. 
With the continuous improvement of classification requirements in practical applications, 
ROC (Receiver Operating Characteristic), AUC (Area Under ROC Curve), a non-traditional 
measurement standard, have emerged. ROC is a curve drawn on a two-dimensional plane with 
FPR as the abscissa and TPR as the ordinate. It can reflect the sensitivity and specificity 
of the continuous variables as a comprehensive indicator. It can also solve the problem of 
class imbalance in the actual dataset. AUC quantifies the area under the ROC curve into a 
numerical value to make the results more intuitive.

\subsubsection{Evaluation of Detection Methods} 
Detection~\cite{Pal-1993-RIST} is another common task in analyzing histopathological WSIs. 
Detection is not only to determine the attributes of the region identified in WSI, but also 
to identify and obtain more detailed results. Because of the similarity between testing and 
classification, most of the evaluation indexes are the same as the classification, including 
accuracy, precision, recall, F-measure, sensitivity, and specificity that we have discussed in 
Sec.~\ref{ss:int:BEI}. However, in WSI detection, it is difficult to locate, determine and 
quantify multiple lesions. Therefore, FROC (Free Receiver Operating Characteristic 
Curve)~\cite{Egan-1961-OCSD} is proposed to evaluate the detection results. FROC curve is 
a small variation of the ROC curve. It is a curve drawn on a two-dimensional plane with FP 
as the horizontal coordinate and TPR as the vertical coordinate. This allows the detection 
of multiple lesion areas on a single WSI.

\subsection{Summary}
According to the review above, we can see that the commonly used public datasets are TCGA, 
TUPAC16, and Kimia Path24 for the classification, segmentation, and detection of histopathological 
images using the combination of WSI technology and CAD with the brief introduction. Also, the evaluation indicators of these three tasks. The 
basic commonly used evaluation indicators are accuracy, precision, recall, sensitivity, 
and specificity. In terms of classification, there are comprehensive indicators such as AUC. 
Dice co-efficient and Jaccard index in segmentation indicators have become popular in 
recent years, and FROC in detection indicators can be used for positioning, qualitative 
and quantitative analysis of multiple lesions.

\section{Feature Extraction}
\label{s:FE}
Traditional image feature extraction is generally divided into three steps: preprocessing, 
feature extraction, and feature processing. Then using machine learning methods to segment 
and classify the features. The purpose of preprocessing is to eliminate interference factors and 
highlight characteristic information. The main methods are: image 
standardization~\cite{Shen-2003-PSBM} (adjust the image size); image 
normalization~\cite{Khan-2014-NMAS} (adjust the image center of gravity to 0). The main 
purpose of feature processing is to eliminate features with a small amount of information 
and reduce the amount of calculation. The common feature processing method is principal 
components analysis~\cite{Jhajharia-2016-NNBB}.

\begin{figure}[htbp!]
\centerline{\includegraphics[width=0.65\linewidth]{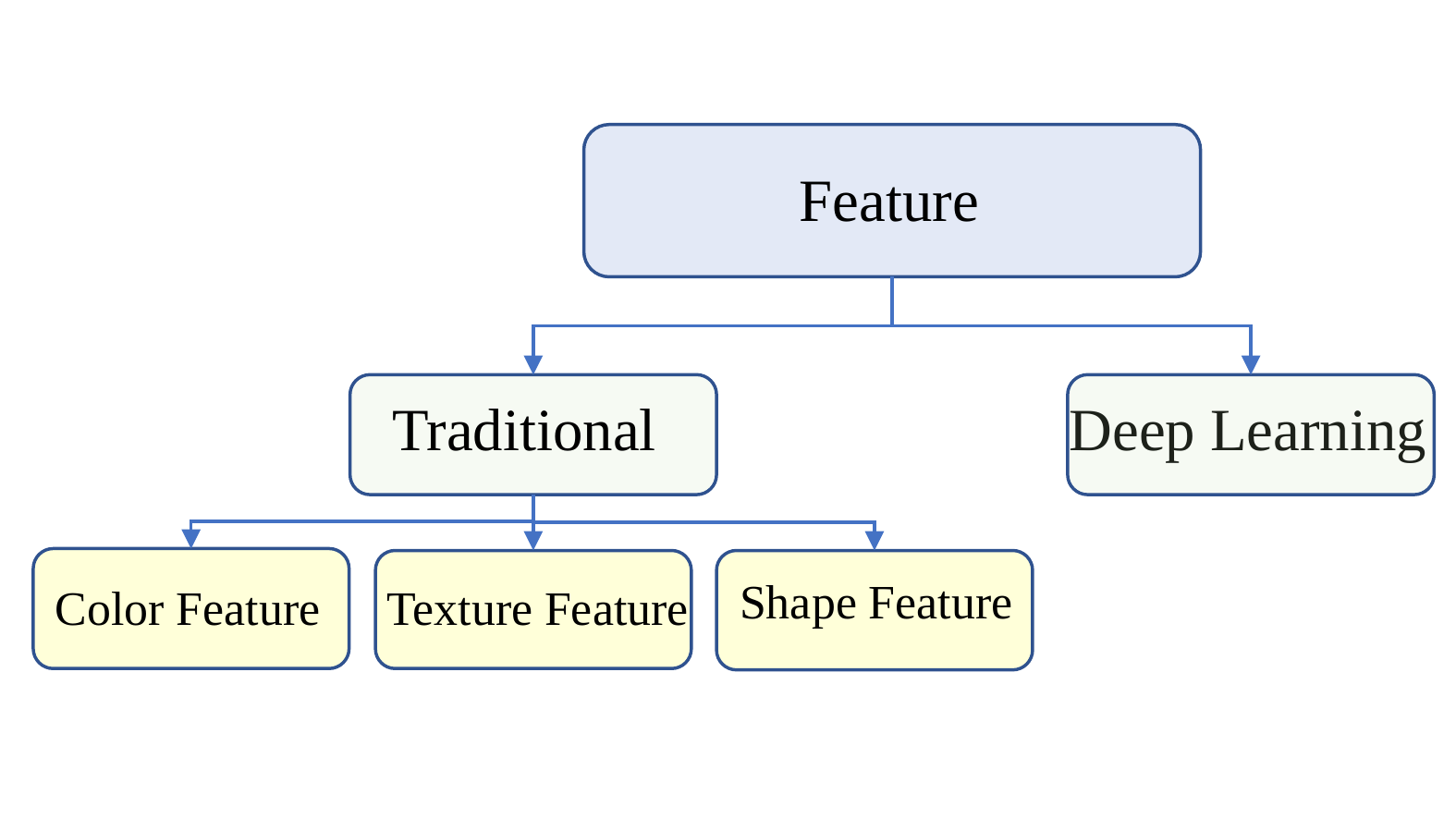}}
\caption{The types of feature extraction methods.}
\label{fig:FE}
\end{figure}

Among them, feature extraction is a crucial step. Converting input data into a set of 
features is called feature extraction~\cite{Ping-2013-RIFE}. The main goal of feature 
extraction is to obtain the most relevant information from the original data and represent 
the information in a lower-dimensional space~\cite{Kumar-2014-DRFE}. Therefore, in this 
section, we mainly summarize the features extracted in WSI for CAD. The types of extracted 
features are shown in Fig.~\ref{fig:FE}.

\subsection{Traditional Feature Extraction}
In the process of segmentation, classification, or detection combined with CAD and WSI 
technology, the commonly used extracted features include color features, texture features 
and shape features.

\subsubsection{Color Featur Extraction}
Color is an important feature, which is widely used for image representation~\cite{Kodituwakku-2004-CCFI}. 
The color of an image is invariant to rotation, translation, or scaling. Color characteristics are defined according to a specific color 
space or model~\cite{Ping-2013-RIFE}. Many color spaces are used in the literature such as 
RGB~\cite{Roullier-2010-GMRS}, HSV ((Hue Saturation Value))~\cite{Mercan-2016-LDRR}, and 
LAB~\cite{Mercan-2016-MIML}. Common color features include color histograms, color moments, 
and color coherence vector(CCV)~\cite{Pass-1997-CIUC}. Among the papers we summarized, 
24 papers used color features~\cite{Kong-2009-CAEN,Roullier-2010-GMRS,Samsi-2012-ECFA,Kothari-2012-BIMP,Akakin-2012-CBMI,Collins-2013-AMAU,Nayak-2013-CTHS,Veta-2013-DMFB,Kothari-2013-ETFA,Homeyer-2013-PQNH,Hiary-2013-SLWS,Bautista-2014-CSWS,Mercan-2014-LDRR,Yeh-2014-MSDP,Litjens-2015-ADPC,Weingant-2015-EPTC,Li-2015-FRID,Mercan-2016-LDRR,Barker-2016-ACBT,Mercan-2016-MIML,Brieu-2016-SSMS,Mercan-2017-MIML,Cruz-2018-HTAS,Morkunas-2018-MLBC}.

\paragraph{RGB-based Color Features}~{}\\

We have found 12 studies that utilized RGB feature extraction technique~\cite{Kong-2009-CAEN, Roullier-2010-GMRS, Akakin-2012-CBMI, Collins-2013-AMAU, Veta-2013-DMFB, Homeyer-2013-PQNH, Hiary-2013-SLWS, Bautista-2014-CSWS, Yeh-2014-MSDP, Litjens-2015-ADPC, Cruz-2018-HTAS, Morkunas-2018-MLBC}.

\begin{figure}[htbp!]
\centerline{\includegraphics[width=0.8\linewidth]{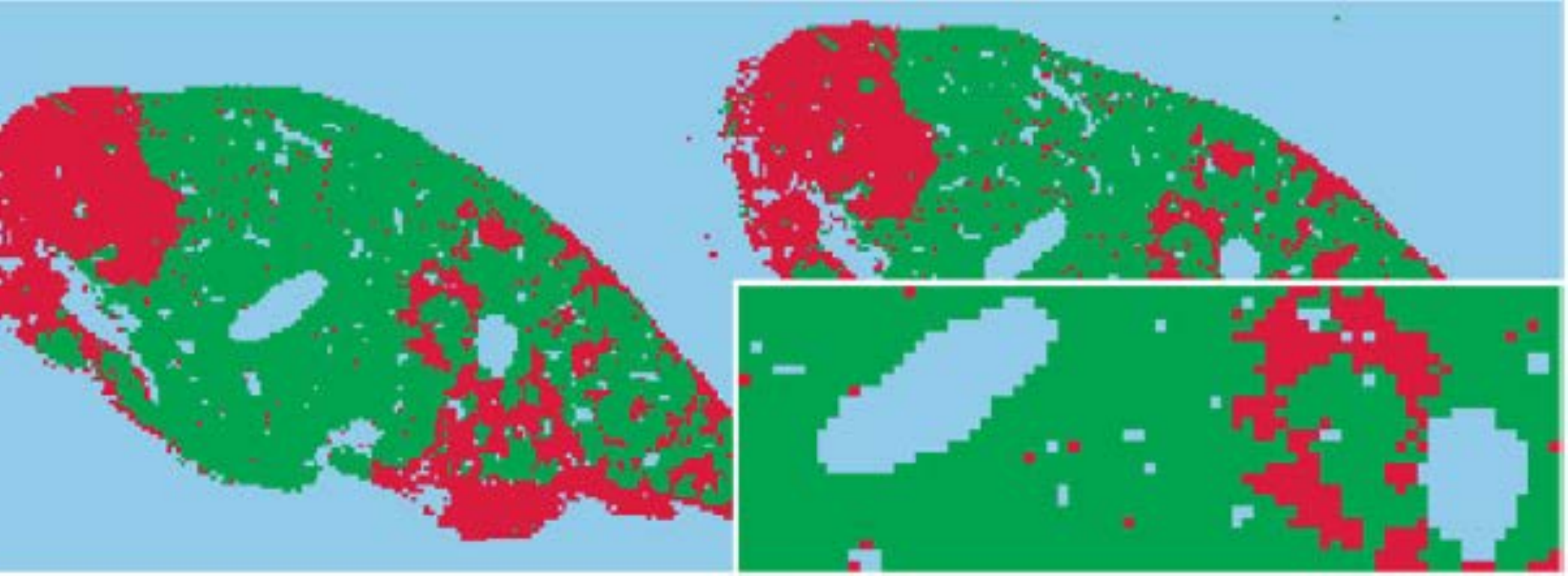}}
\caption{An image after RGB feature extraction and classification in~\cite{Homeyer-2013-PQNH}. 
This figure corresponds to Fig.2 (2) in the original paper.}
\label{fig:PQNH}
\end{figure}

\begin{figure}[htbp!]
\centerline{\includegraphics[width=0.6\linewidth]{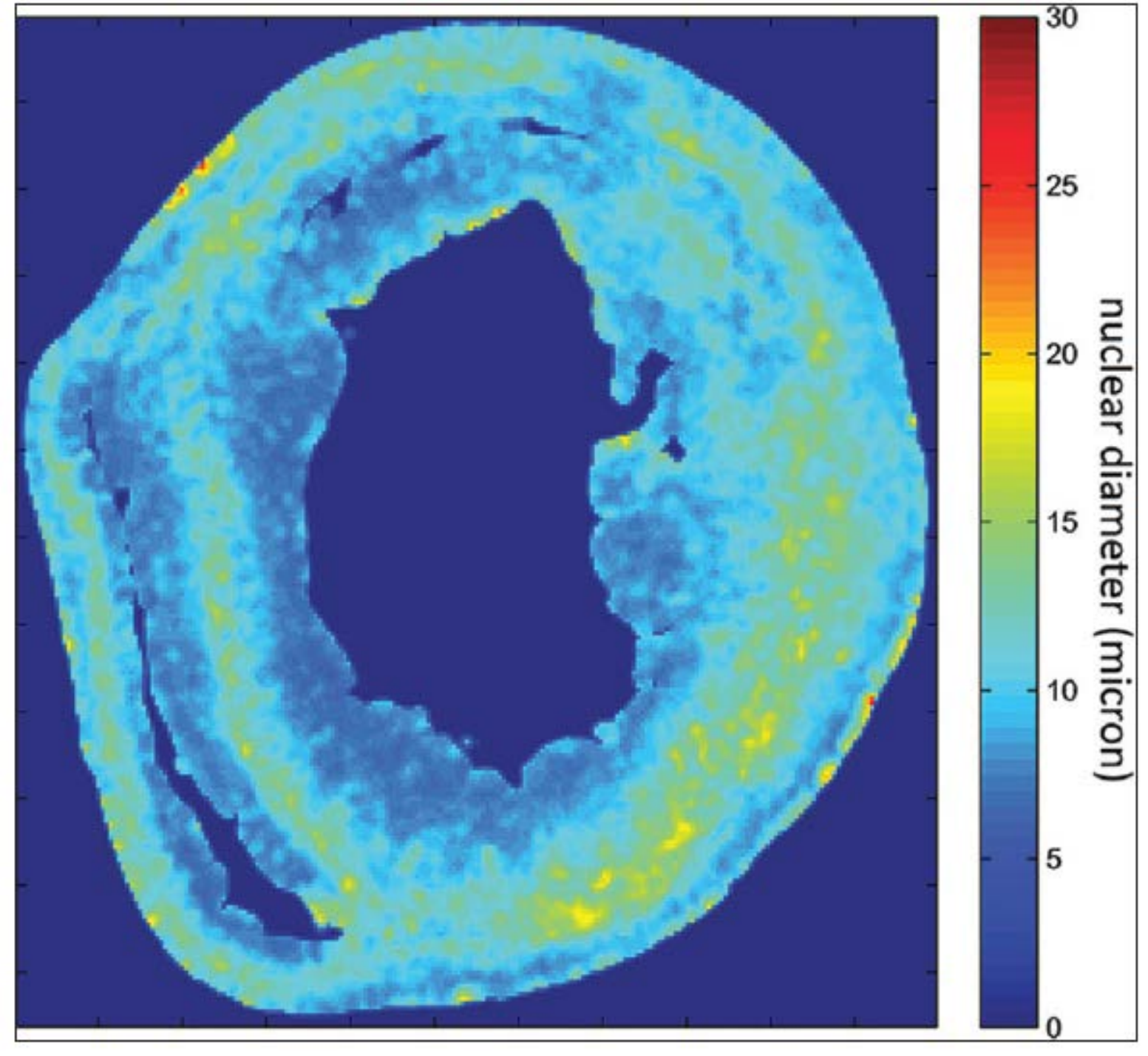}}
\caption{An example of feature distribution image in~\cite{Yeh-2014-MSDP}. 
This figure corresponds to Fig.10 in the original paper.}
\label{fig:RGB}
\end{figure}

The color features extracted by ~\cite{Kong-2009-CAEN} are combined with the color and 
entropy information extracted from the RGB image channel. In~\cite{Roullier-2010-GMRS}, 
the vector extracted with the RGB feature indicates that the feature vector of each pixel 
is the local entropy of the red-green difference calculated in the square neighborhood 
around the pixel. In~\cite{Akakin-2012-CBMI}, the mean and standard deviations are 
calculated as first-order and second-order statistical features from the three RGB channels, and there
are six features. The author in ~\cite{Collins-2013-AMAU} extracts core RGB 
features. The color features of~\cite{Veta-2013-DMFB} are described by the average, 
standard deviation, minimum and maximum values of the three color channels in the RGB 
color space in the candidate area and two other areas. In Fig.~\ref{fig:RGB} an example of 
features distribution image showing the spatial distribution of the cell nuclear diameter 
in~\cite{Yeh-2014-MSDP}. The work of~\cite {Homeyer-2013-PQNH} extracts the PVS function 
of each R, G, and B channel (8 pixel value statistics). PVS is composed of the minimum, 
maximum, sum, average, and standard deviation of the constituent pixels, and the lower 
quartile, median and upper quartile are composed of values in a specific color channel. 
Fig.~\ref{fig:PQNH} shows the RGB feature extraction and classification 
in~\cite{Homeyer-2013-PQNH}.

The author in ~\cite{Hiary-2013-SLWS} extracts the variance in each color channel of RGB: s2R, s2G, s2B, 
the variance (maximum value) between the peaks of each color channel s2. The work 
of~\cite{Bautista-2014-CSWS} uses color saturation and RGB color. \cite{Yeh-2014-MSDP} extracts color features from the RGB channels. Fig.~\ref{fig:RGB} shows an 
example of feature distribution of the cell nuclear 
in diameter~\cite{Yeh-2014-MSDP}.

The author in ~\cite{Litjens-2015-ADPC} extracts RGB histogram, and overlay features. 
\cite{Cruz-2018-HTAS} extracts the first-order statistics of 14 color channels, and the 
color histogram of each RGB channel is 8×38 bin histogram. In~\cite{Morkunas-2018-MLBC}, 
for each 2D superpixel (for example, grayscale superpixel), two statistics are calculated 
of mean and standard deviation of the pixel value. For 3D superpixels (such as RGB 
superpixels), eight statistics are calculated the mean and standard deviation of the 
pixel value for each color channel and each RGB superpixel, then this as a color function.

\paragraph{HSV-based Color Features}~{}
\\

There are four papers on color feature extraction based on 
HSV~\cite{Samsi-2012-ECFA, Akakin-2012-CBMI, Homeyer-2013-PQNH, Bautista-2014-CSWS}.

In~\cite{Samsi-2012-ECFA}, the extracted color feature is the hue channel converted 
from the HSV color space of the original image. In~\cite{Akakin-2012-CBMI}, from the 
three HSV channels, the average and standard deviation are calculated as first-order 
and second-order statistical features, and a total of six features are extracted. 
\cite{Homeyer-2013-PQNH} extracts the PVS function of each H, S and V channel. 
\cite{Bautista-2014-CSWS} uses color saturation and value and RGB color as a function 
in HSV color space.

\paragraph{LAB-based Color Features}~{}
\\ 

There are five papers on color feature extraction based on 
LAB~\cite{Kong-2009-CAEN, Akakin-2012-CBMI, Mercan-2014-LDRR, Mercan-2016-LDRR, Mercan-2017-MIML}.

\begin{figure}[htbp!]
\centerline{\includegraphics[width=0.6\linewidth]{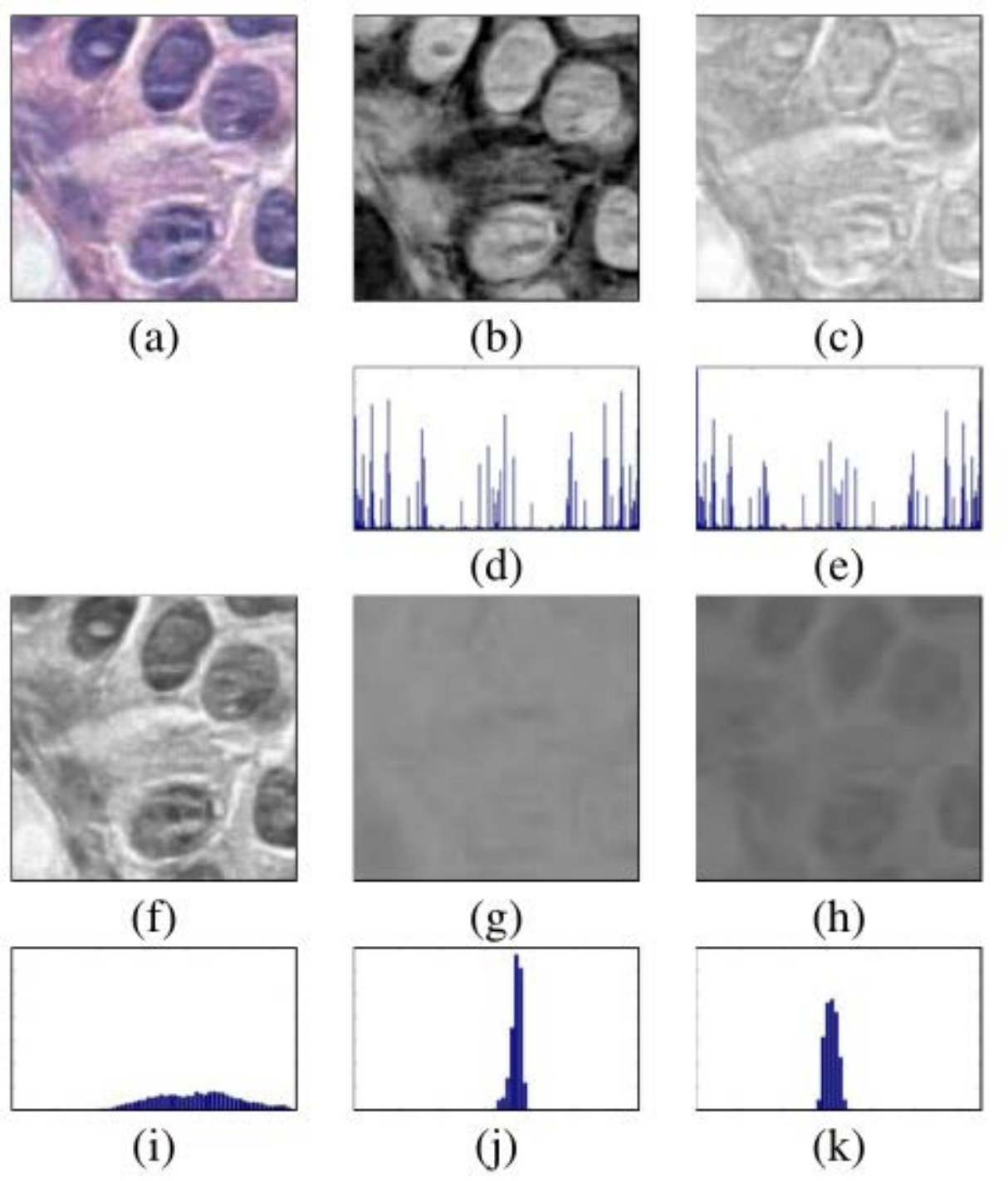}}
\caption{(a) Original $120 \times 120$ pixel patch, 
(b) deconvolved color channel that shows good contrast for nuclei dyed with haematoxylin, 
(c) deconvolved color channel that shows good contrast for eosin dye, 
(d) LBP histogram calculated on haematoxylin channel, 
(e) LBP histogram calculated on eosin channel,(f, g, h) L, a, and b channels of the image 
in LAB color space,(i, j, k) color histograms of L, A and B channels. At the end, LBP 
histograms on two channels are concatenated to produce the first set of features, color 
histograms on LAB channels are concatenated to produce the second set of features. 
These figure are from~\cite{Mercan-2014-LDRR}. This figure corresponds to Fig.3 in the 
original paper.}
\label{fig:LDRRLAB}
\end{figure}

In~\cite{Kong-2009-CAEN}, the extracted color features are composed of color and 
entropy information extracted from LAB image channels. In~\cite{Akakin-2012-CBMI}, 
from the three channels of CIELAB, the average and standard deviation are calculated 
as first-order and second-order statistical features, and a total of six features are 
extracted. \cite{Mercan-2014-LDRR} extracted the color histogram calculated in LAB 
space. Fig.~\ref{fig:LDRRLAB} shows the color histogram mentioned in the paper. 
Cutting the WSI into a patch is a normal operation in the image processing process. 
(a,b,c,f,g,h) in Fig.~\ref{fig:LDRRLAB} are the image blocks after the WSI slice.

The color feature of ~\cite{Mercan-2016-LDRR} is the LAB histogram of the color. 
\cite{Mercan-2017-MIML} uses the color histogram calculated for each channel in the 
CIE-Lab space as the color feature.

\paragraph{Others Color Features}~{}
\\ 

Other papers related to color feature extraction total of 
six~\cite{Nayak-2013-CTHS, Kothari-2013-ETFA, Hiary-2013-SLWS, Weingant-2015-EPTC, Li-2015-FRID, Barker-2016-ACBT, Brieu-2016-SSMS}.

\begin{figure}[htbp!]
\centerline{\includegraphics[width=0.6\linewidth]{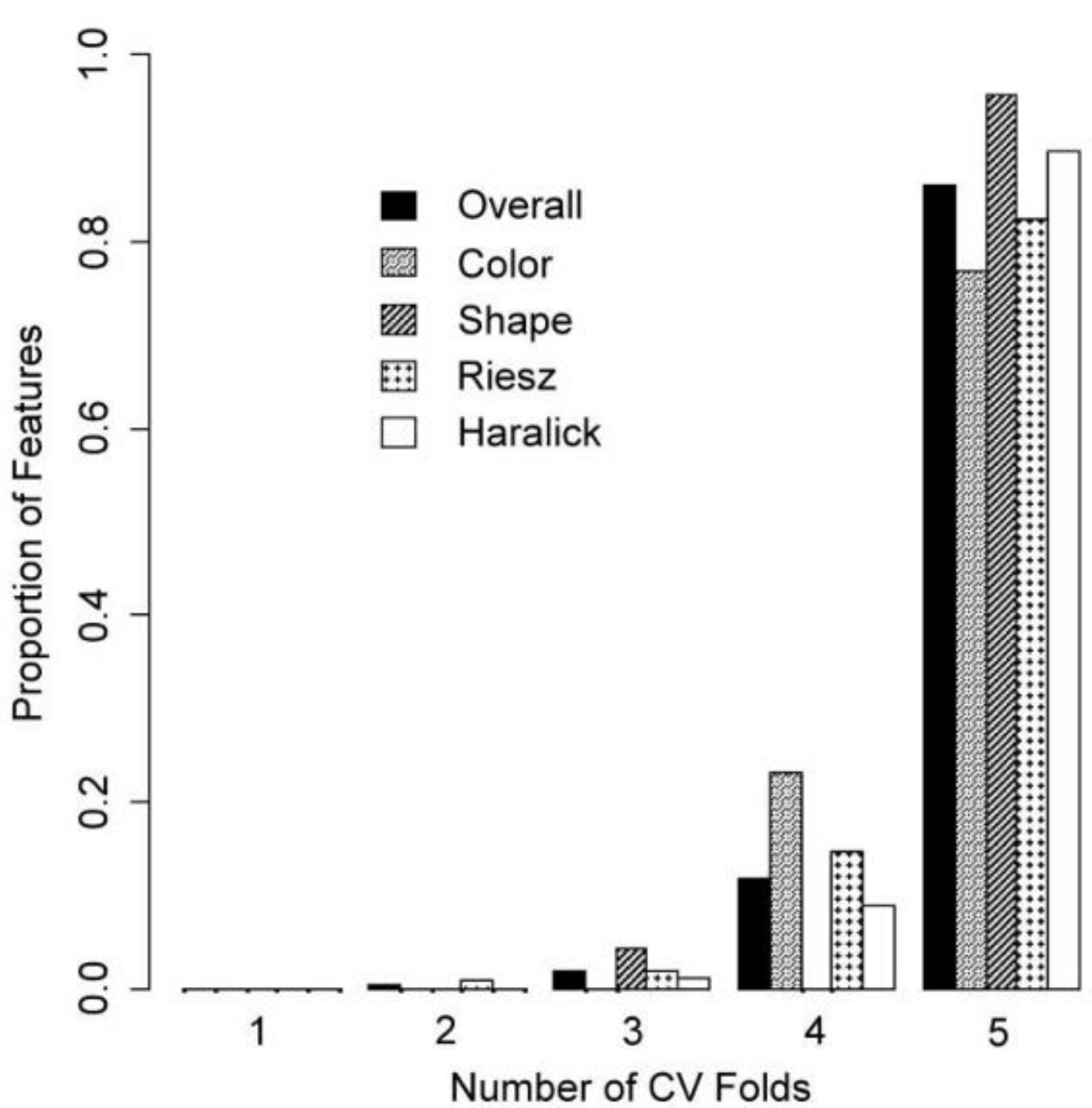}}
\caption{Histogram distribution of features used in Elastic Net models, showing the 
number of models in which features of a given class appear.  The number of features 
is normalized based on the total number of features represented for each class. 
Bar patterns represent the feature class.  Most features for each class appear in all, or nearly all models, as expected if they have diagnostic value. This 
figure is from ~\cite{Barker-2016-ACBT}. This figure corresponds to Fig.7 in the 
original paper. }
\label{fig:ACBTO}
\end{figure}

\cite{Nayak-2013-CTHS} and~\cite{Kothari-2013-ETFA} extract the color information in 
WSIs as features. \cite{Hiary-2013-SLWS} extracts the average value of the variance, 
saturation, brightness of HSI and the hue value of the color model $\mu$. 
\cite{Weingant-2015-EPTC} extracts the color channel histogram as the color feature. 
\cite{Li-2015-FRID} extracts the histogram of the three-channel HSD color model as color 
features. \cite{Brieu-2016-SSMS} extracts the color information in WSI as features. 
\cite{Barker-2016-ACBT} extracts rough color features. The rough feature is 
the use of feature analysis of the diversity of rough areas in WSI to roughly characterize 
their shape, color, and texture. The fine feature refers to the more comprehensive image 
features extracted from the slice to express deeper features. The feature histogram extracts 
in~\cite{Barker-2016-ACBT} is shown in Fig.~\ref{fig:ACBTO}.

From the above review, we can see that in terms of color feature extraction, the RGB 
color features are used most frequently, focusing on the period from 2009 to 2018. 
Then HSV color feature and LAB (CIE-LAB) color feature. The dates are 2012 to 2014 and 
2009 to 2017. In other papers, HSI and HSD color models are used as color characteristics.

\subsubsection{Texture Featur Extraction}
The texture feature describes the surface properties of the object corresponding 
to the image or image area. Unlike the color feature, the texture feature is 
not based on the feature of pixels. It needs to be calculated in the area containing 
multiple pixels. Texture feature is an effective method when judging images with large differences in thickness and density. However, when the thickness, 
density, feature are easy to distinguish between the information. However, 
difficult for the usual texture features to accurately reflect the differences between 
the textures with different human visual perception.

Commonly used texture information description methods are: statistical methods ( such as 
gray-level co-occurrence matrix (GLCM)~\cite{Mohanaiah-2013-ITFE}), geometric methods 
(such as voronoi checkerboard feature method~\cite{Tuceryan-1990-TSUV}), model methods 
(such as random fields~\cite{Chellappa-1985-CTUG}), and signal processing methods (such 
as wavelet transform~\cite{Pichler-1996-CTFE}). Among the papers we summarized, 57 papers 
used texture features~\cite{Diamond-2004-UMCT,Petushi-2006-LSCH,Sertel-2008-CAPN,Sertel-2009-CCCS,Kong-2009-CAEN,Sertel-2009-CPNW,Roullier-2010-MEBC,Doyle-2010-BBMC,Difranco-2011-ESWS,Kong-2011-IMAG,Roullier-2011-MRGA,Grunkin-2011-PCIA,Nguyen-2011-PCDF,Akakin-2012-CBMI,Sharma-2012-DSHI,Nayak-2013-CTHS,Jiao-2013-CCDU,Veta-2013-DMFB,Kothari-2013-ETFA,Kong-2013-MBMA,Homeyer-2013-PQNH,Hiary-2013-SLWS,Mercan-2014-LDRR,Apou-2014-FSTC,Bejnordi-2015-MSSC,Sharma-2015-ABND,Swiderska-2015-CMSA,Weingant-2015-EPTC,Li-2015-FRID,Zhang-2015-FGHI,Cooper-2015-NGPA,Apou-2015-SWSI,Peikari-2015-TDRR,Mercan-2016-LDRR,Barker-2016-ACBT,Bejnordi-2016-ADDW,Zhao-2016-AGEW,Harder-2016-CFCG,Shirinifard-2016-DPAU,Gadermayr-2016-DACC,Leo-2016-ESHF,Mercan-2016-MIML,Brieu-2016-SSMS,Saltz-2017-CSSG,Mercan-2017-MIML,Hu-2017-DLBC,Bejnordi-2017-DADL,Valkonen-2017-MDWS,Mirschl-2018-DLCI,Xu-2018-AACM,Yoshida-2018-AHCWG,Han-2018-ACDL,Morkunas-2018-MLBC,Simon-2018-MRLF,Klimov-2019-WSIB}.

\paragraph{Local Binary Pattern-based Texture Features}~{}
\\ 

\begin{figure}[htbp!]
\centerline{\includegraphics[width=0.98\linewidth]{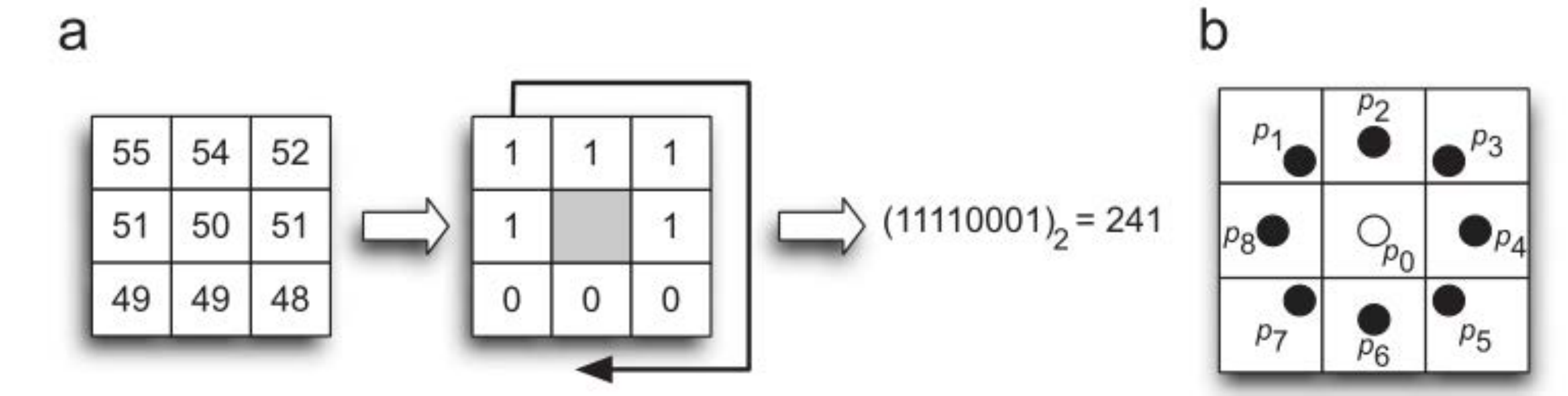}}
\caption{ (a) The conventional LBP operator; 
(b) Circular pattern used to compute rotation invariant uniform patterns. 
This figure corresponds to Fig.5 in ~\cite{Sertel-2009-CPNW}.}
\label{fig:CAPNLBP}
\end{figure}

There are 14 papers on texture feature extraction based on local binary pattern 
(LBP)~\cite{Sertel-2008-CAPN, Sertel-2009-CPNW, Roullier-2010-MEBC, Roullier-2011-MRGA, Homeyer-2013-PQNH, Mercan-2014-LDRR, Bejnordi-2015-MSSC, Mercan-2016-LDRR, Mercan-2016-MIML, Gadermayr-2016-DACC, Babaie-2017-CRDP, Mercan-2017-MIML, Bejnordi-2017-DADL, Simon-2018-MRLF}. 
In~\cite{Sertel-2008-CAPN, Sertel-2009-CPNW, Roullier-2010-MEBC, Roullier-2011-MRGA, Homeyer-2013-PQNH , Bejnordi-2015-MSSC, Mercan-2016-LDRR, Mercan-2016-MIML , Babaie-2017-CRDP, Mercan-2017-MIML}, 
texture histograms of LBP features are extracted as texture features. Fig.~\ref{fig:CAPNLBP} 
shows the calculation of the LBP feature of a given pixel in~\cite{Sertel-2009-CPNW}.

Texture features compute using LBP for small image patches are extracted from~\cite{Mercan-2014-LDRR}. 
In~\cite{Gadermayr-2016-DACC}, multi-resolution LBP is extracted as the texture feature. 
In~\cite{Bejnordi-2017-DADL}, different structural texture features are extracted, including 
LBP features. In ~\cite{Simon-2018-MRLF}, LBP, MRC LBP feature are extracted as texture features.
\paragraph{Haralick-based Texture Features}~{}
\\ 

\begin{figure}[htbp!]
\centerline{\includegraphics[width=0.5\linewidth]{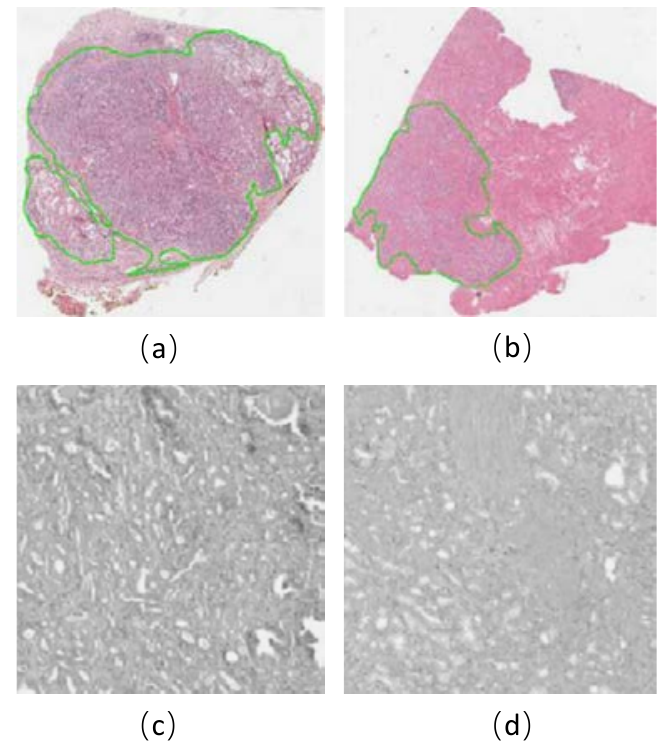}}
\caption{ (a, b) is the experimental image, 
(c, d) is the Haralick intensity texture. 
This figure corresponds to Fig.2 in~\cite{Leo-2016-ESHF}.}
\label{fig:ESHF}
\end{figure}

Five papers involve texture feature extraction based on 
Haralick~\cite{Diamond-2004-UMCT, Sertel-2008-CAPN, Kong-2009-CAEN, Leo-2016-ESHF, Hu-2017-DLBC}.
In~\cite{Diamond-2004-UMCT}, 4 Haralick features are the most suitable for discriminating stroma 
from PCA. Haralick features are extracted as texture features 
in~\cite{Sertel-2008-CAPN, Kong-2009-CAEN, Leo-2016-ESHF}. 
In~\cite{Hu-2017-DLBC}, extracts 57 subcellular location features, including Haralick texture 
features and DNA overlapping the features (experiments). 
The experimental images in~\cite{Leo-2016-ESHF} and the extracted haralick features are 
shown in Fig.~\ref{fig:ESHF}.

\paragraph{GLCM-based Texture Features}~{}
\\ 
Eight papers involve texture feature extraction based on 
GLCM~\cite{Sertel-2009-CCCS, Doyle-2010-BBMC, Difranco-2011-ESWS, Jiao-2013-CCDU, Hiary-2013-SLWS, Weingant-2015-EPTC, Bejnordi-2016-ADDW, Morkunas-2018-MLBC}.

In~\cite{Sertel-2009-CCCS}, the extraction of the mean and variance of the range of 
values within the local neighborhoods and entropy and homogeneity of co-occurrence 
histograms as texture features. Co-occurrence features are extracted as texture features 
from~\cite{Doyle-2010-BBMC} and~\cite{Difranco-2011-ESWS}. 16 
features such as mean and variance are combined to form a group of 18 features~\cite{Jiao-2013-CCDU}. 
In~\cite{Hiary-2013-SLWS}, co-occurrence matrix, correlation, and energy are extracted 
as texture features. In~\cite{Weingant-2015-EPTC}, GLCM features are extracted. 
In~\cite{Bejnordi-2016-ADDW}, co-occurrence matrix statistics are extracted for each 
hyperpixel. In~\cite{Morkunas-2018-MLBC}, 2D hyper-pixel texture is obtained by using 
the spatial grayscale symbiosis matrix and 1px displacement vector of 3D hyper-pixel. 
From the co-occurrence matrix, the second moment of angle, contrast, correlation, sum 
of squares, deficit moment, average, sum variance, sum entropy, entropy, 
difference variance, difference entropy, and correlation information measures 1 and 2 
are calculated. The average value of the 13 parameters obtained is the characteristic descriptor.

\paragraph{Filter and Scale-invariant Feature Transform(SIFT)-based Texture Features}~{}
\\

There are five related papers on filter and SIFT based texture feature 
extraction~\cite{Nguyen-2011-PCDF, Peikari-2015-TDRR, Bejnordi-2016-ADDW, Bejnordi-2017-DADL, Valkonen-2017-MDWS}.

\begin{figure}[htbp!]
\centerline{\includegraphics[width=0.98\linewidth]{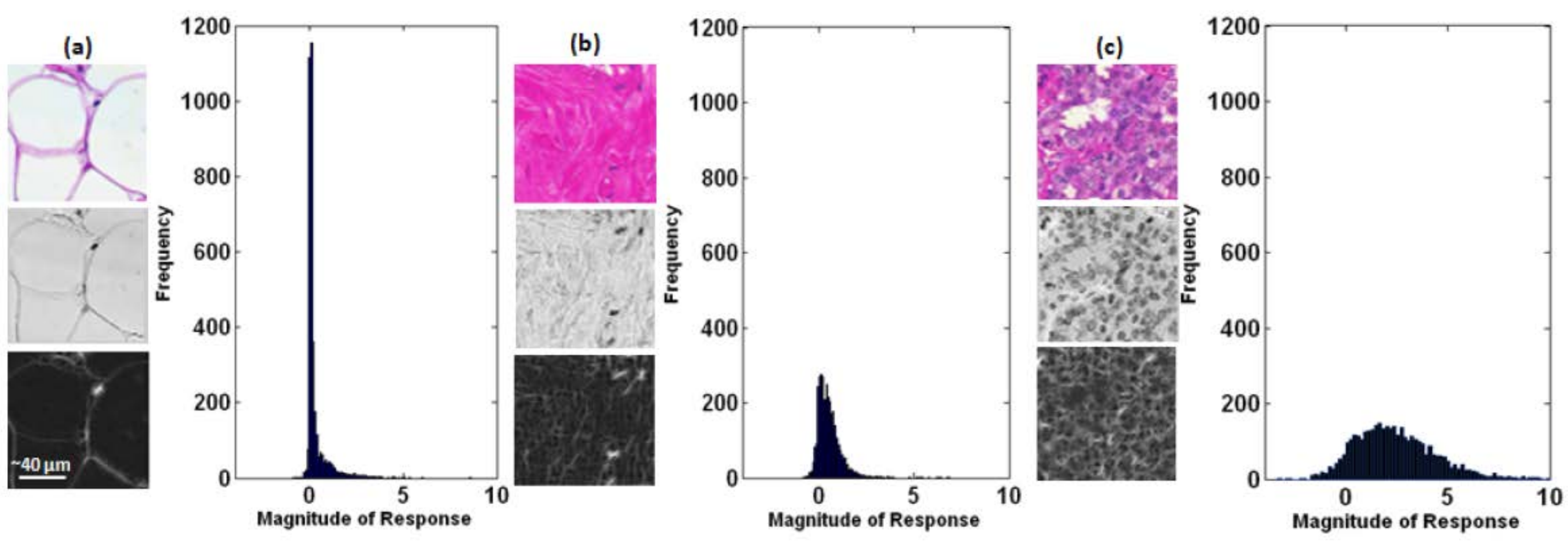}}
\caption{ Patches of three different tissue types (in RGB) corresponding to (a) fat, 
(b) stroma, and (c) epithelial cells -top- along with their normalized luminance channel 
images (after converting them from RGB to LAB) -middle-, and maximal filter responses 
after convolving Gaussian-like filters at all directions of onescale (($\sigma$x, $\sigma$y) = (1,3)) 
-bottom-, and plot showing their histogram of filter response magnitudes. This figure 
corresponds to Fig.3 in~\cite{Peikari-2015-TDRR}.}
\label{fig:TDRRF}
\end{figure}

In~\cite{Nguyen-2011-PCDF}, first-order statistics, second-order statistics, and 
gabor filter features are used as texture features. In~\cite{Peikari-2015-TDRR}, a Gauss-like 
texture filter is applied to extract texture features. Fig.~\ref{fig:TDRRF} shows the 
uniform distribution of histogram filter response in~\cite{Peikari-2015-TDRR}.

In~\cite{Bejnordi-2016-ADDW}, the gray histogram statistics extracted from the 
filter bank response for each hyperpixel. Different structural texture features, 
such as SIFT features, are extracted from~\cite{Bejnordi-2017-DADL}. The vlfeat 
implementation of MSER (Maximally Stable Extremal Regions) and SIFT is used by 
extracting from~\cite{Valkonen-2017-MDWS}.

\paragraph{Others Texture Features}~{}
\\ 

\begin{figure}[htbp!]
\centerline{\includegraphics[width=0.7\linewidth]{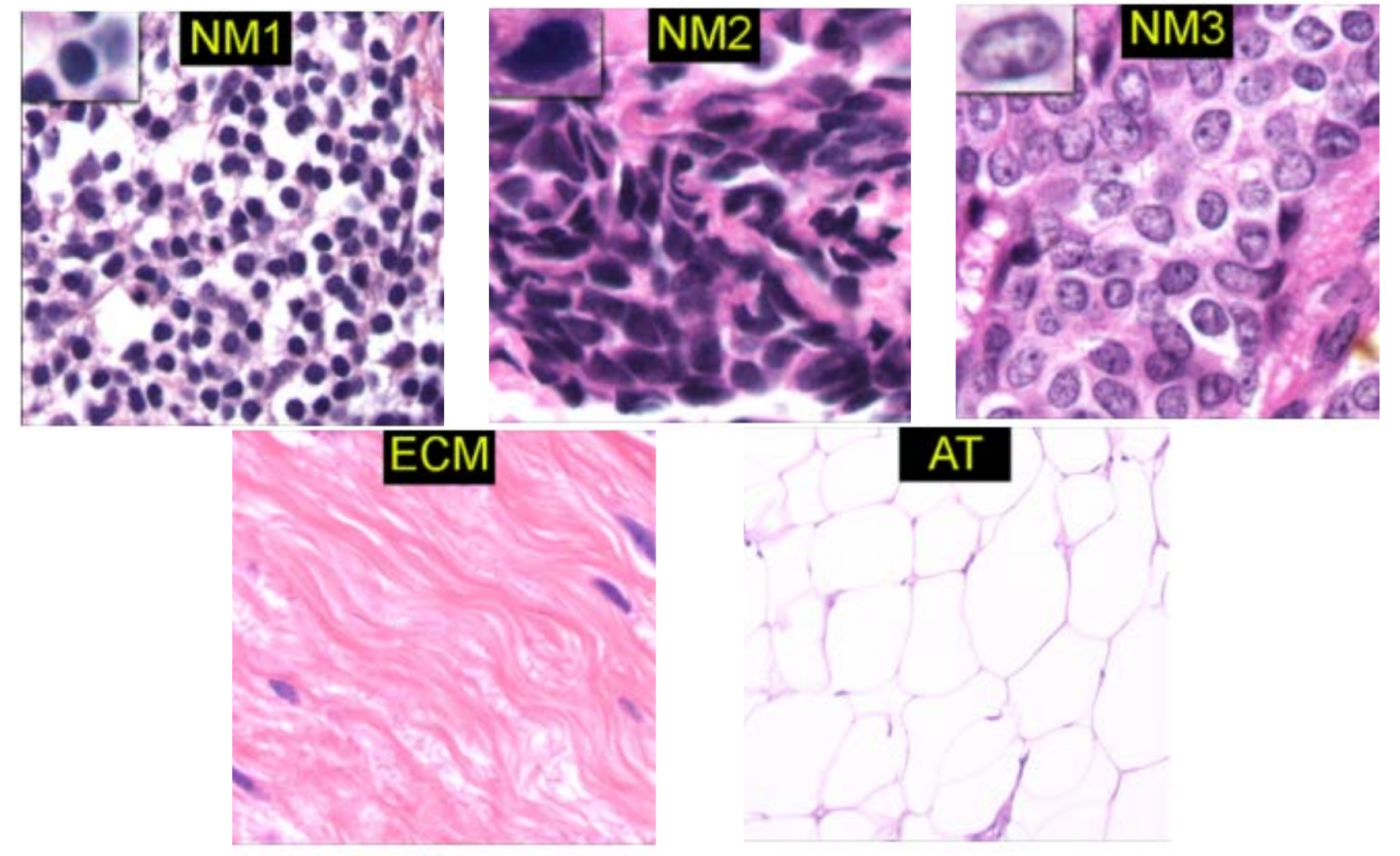}}
\caption{ Tissue microtextures identified using image processing. 
The first three columns show examples of cell nuclei belonging to nucleus morphology 
categories NM1, NM2, and NM3, respectively. The textures identified as ECM and AT represent, 
respectively the collagen-rich stroma, and fat and tissue-devoid regions of the slide. 
This figure corresponds to Fig.2 in~\cite{Petushi-2006-LSCH}.}
\label{fig:LSCH}
\end{figure}

There are a total of 24 related papers based on the extraction of other texture 
features~\cite{Petushi-2006-LSCH, Kong-2011-IMAG, Grunkin-2011-PCIA, Akakin-2012-CBMI, Sharma-2012-DSHI, Nayak-2013-CTHS, Kothari-2013-ETFA, Kong-2013-MBMA, Apou-2014-FSTC, Swiderska-2015-CMSA, Li-2015-FRID, Zhang-2015-FGHI, Cooper-2015-NGPA, Apou-2015-SWSI, Barker-2016-ACBT, Shirinifard-2016-DPAU, Gadermayr-2016-DACC, Brieu-2016-SSMS, Mirschl-2018-DLCI, Xu-2018-AACM, Yoshida-2018-AHCWG, Han-2018-ACDL, Klimov-2019-WSIB}. 
Texture Parameters: DNM1, DNM2,DNM3, DNM1-2, DNM1-3, DNM2-3, DNM2-2-3, DT,DN are extracted 
from~\cite{Petushi-2006-LSCH}. The feature vector extracted in~\cite{Petushi-2006-LSCH} is shown 
in Fig.~\ref{fig:LSCH}.

\begin{figure}[htbp!]
\centerline{\includegraphics[width=0.98\linewidth]{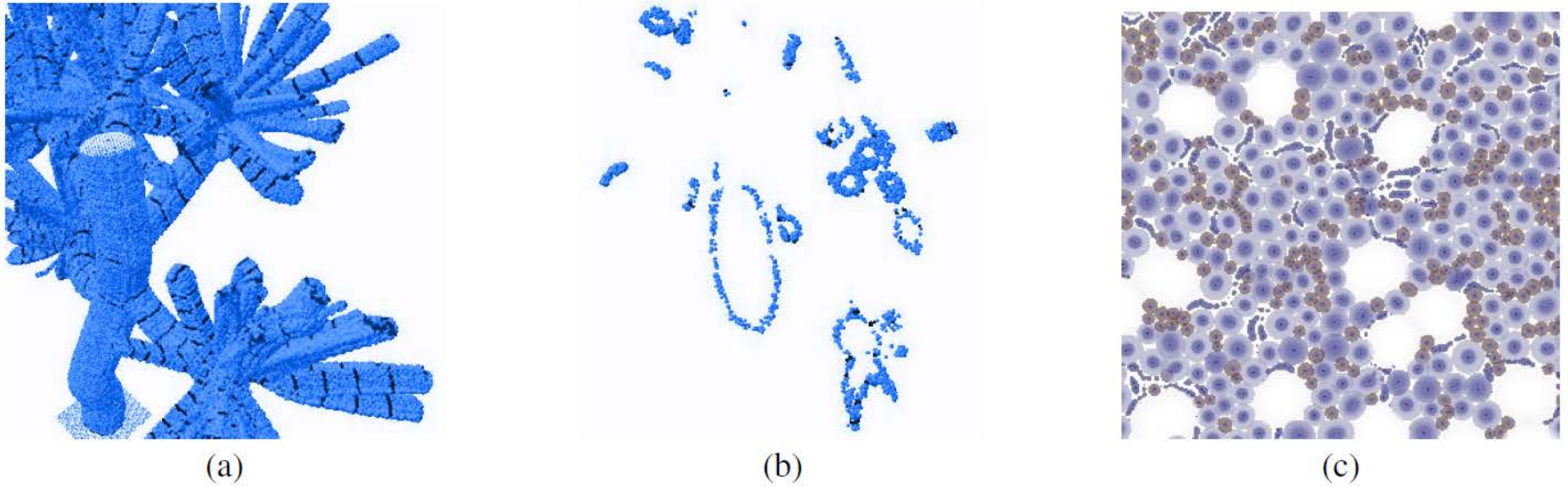}}
\caption{ Procedural structure and texture: 
(a) Procedural 3D model of a lobular epithelial layer. 
(b) Slice of (a) showing 2D lobular patterns. 
(c) Procedural cell rendering (random layout, arbitrary colors). 
This figure corresponds to Fig.5 in~\cite{Apou-2015-SWSI}. }
\label{fig:SWSI}
\end{figure}

The texture is applied to the cytoplasmic region around the nucleus in~\cite{Kong-2011-IMAG}. 
In~\cite{Grunkin-2011-PCIA}, texture features are used to identify areas with high or low 
intensity variability in the image. Average, standard deviation, contrast, correlation, 
energy, entropy, and uniformity are extracted from~\cite{Akakin-2012-CBMI} as texture 
features. Texton-based texture is extracted from~\cite{Sharma-2012-DSHI}. In~\cite{Nayak-2013-CTHS}, texture information is used as feature. In ~\cite{Kothari-2013-ETFA}, quantitative 
image features are extracted to capture its texture. Nuclear texture features are extracted from the 
chromatin content and distribution in~\cite{Kong-2013-MBMA}. Each area is tagged according 
to its texture description in~\cite{Apou-2014-FSTC}. Intensity on the basis of histograms 
of the sum and difference images are extracted as texture features in~\cite{Swiderska-2015-CMSA}. 
In~\cite{Li-2015-FRID}, texture features are extracted for classification. 
In~\cite{Zhang-2015-FGHI}, the texture feature is extracted from the cell image and compressed 
into a binary code. These compressed features are stored in a hash table that allows constant 
time access across many images. In~\cite{Cooper-2015-NGPA}, the texture features of each 
nucleus are extracted. In~\cite{Apou-2015-SWSI}, the area is rendered using a manually 
positioned texture unit. The Fig.~\ref{fig:SWSI} shows the procedural structure and texture 
rendered in~\cite{Apou-2015-SWSI}.

\cite{Barker-2016-ACBT}, use the riesz texture features. The texture features of slides 
stained with Ki67 are extracted from~\cite{Shirinifard-2016-DPAU}. 
The author of ~\cite{Gadermayr-2016-DACC} extracts Histograms of Oriented Gradients (HOG), and Fisher 
Vectors (FV) as texture features. The kernel texture features are extracted 
in~\cite{Brieu-2016-SSMS} and~\cite{Saltz-2017-CSSG}. In~\cite{Mirschl-2018-DLCI}, 
texture features are extracted from the ROI. In~\cite{Xu-2018-AACM}, the texture of the 
subregion is extracted for statistical analysis and classification. In~\cite{Yoshida-2018-AHCWG}, 
the standard deviation (variance) within the area defined by the contour is used as the 
texture feature. \cite{Han-2018-ACDL}, extract the first-and second-order texture features. 
A total of 166 texture features are extracted from the convolved hematoxylin (nuclear staining) 
channel in~\cite{Klimov-2019-WSIB}.

\subsubsection{Shape Featur Extraction}
The shape feature is just what the name suggests. Under normal circumstances, there are 
two ways to represent shape features. One is contour features, and the other is regional 
features. The contour feature of the image is mainly for the outer boundary of the image, 
and the regional feature of the image is related to the entire shape area~\cite{Zhang-2004-RSRD}.
Commonly used shape feature extraction methods include boundary feature method (such as hough 
transform method~\cite{Philip-1994-FHSF}), geometric parameter method (such as moment, 
area, circumference~\cite{Huang-1997-OLSV}), fourier descriptors~\cite{Zhang-2012-MFDS}, 
and other methods. There are 15 papers that use shape features among the papers we have summarized~\cite{Diamond-2004-UMCT,Kong-2011-IMAG,Grunkin-2011-PCIA,Lu-2012-ASAE,Kothari-2012-BIMP,Veta-2012-PVAE,Lopez-2013-ABDM,Collins-2013-AMAU,Veta-2013-DMFB,Kothari-2013-ETFA,Kong-2013-MBMA,Cooper-2015-NGPA,Barker-2016-ACBT,Saltz-2017-CSSG,Xu-2018-AACM}.

\paragraph{Basic Geometric Parameter-based Shape Feature}~{}
\\ 

\begin{figure}[htbp!]
\centerline{\includegraphics[width=0.98\linewidth]{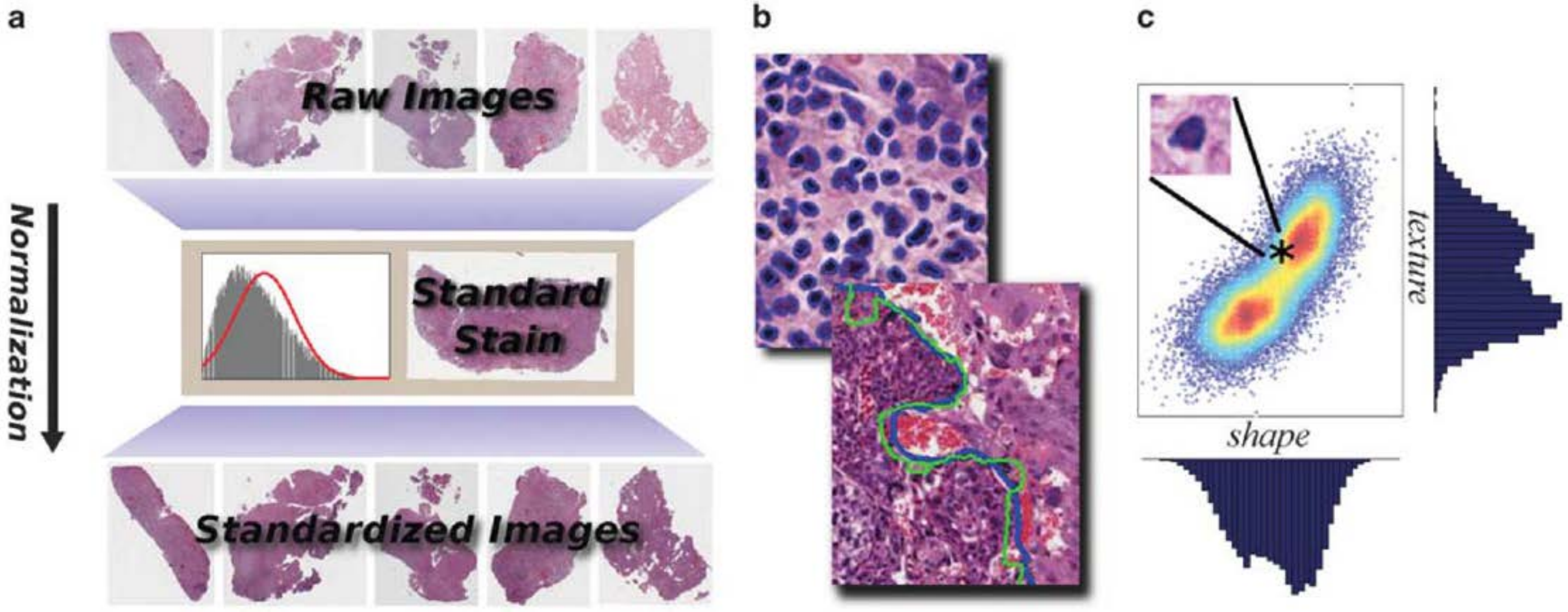}}
\caption{ Image analysis enables the reliable and objective characterization of tissues through 
a process of image normalization, image segmentation, and feature extraction. }
\label{fig:MBMAS}
\end{figure}

Among the papers that used shape feature extraction, seven papers extracted basic geometric shape 
features~\cite{Lu-2012-ASAE, Kothari-2012-BIMP, Veta-2012-PVAE, Collins-2013-AMAU, Veta-2013-DMFB, Kong-2013-MBMA, Grunkin-2011-PCIA}. 
The feature extraction steps in six of the papers are all used to classify, segment, or detect 
the task before it is used to better represent the image. In~\cite{Lu-2012-ASAE}, the major axis length to minor axis length ratio of a best-fit ellipse is extracted as the shape feature to eliminate false regions. In~\cite{Kothari-2012-BIMP}, eosinophilic-object shape features (pixel area, 
elliptical area, major-minor axes lengths, eccentricity, boundary fractal, bending energy, 
convex hull area, solidity, perimeter, and count) are extracted. The author in ~\cite{Veta-2012-PVAE} extracts 
two morphometric features, the mean nuclear area and standard deviation of the nuclear area, 
using a fully automatic segmentation method on WSIs. The author in ~\cite{Collins-2013-AMAU} extracts basic 
morphologic features and calculates its odds ratio for malignant tumors. The author in ~\cite{Veta-2013-DMFB} 
extracts compactness, eccentricity, firmness, and sphericity as shape features. The author in ~\cite{Kong-2013-MBMA} extractes perimeter, eccentricity, circularity, major axis length, 
minor axis length as geometric shape feature. Fig.~\ref{fig:MBMAS} shows the morphological 
characteristic spectrum of the image in~\cite{Kong-2013-MBMA}. The seventh paper~\cite{Grunkin-2011-PCIA} is 
the morphological feature extraction for post-processing.

\paragraph{Other Shape Features}~{}
\\ 

In the early days, some existing library functions and third-party existing functions were 
usually used to directly extract features. In~\cite{Diamond-2004-UMCT, Kong-2011-IMAG, Saltz-2017-CSSG, Xu-2018-AACM}, the nuclear 
morphological features of the nuclei in WSI are extracted as the shape features of the images.

Over time, many other advanced extraction methods have emerged. There are also four papers 
on other shape feature 
extraction~\cite{Lopez-2013-ABDM, Kothari-2013-ETFA, Cooper-2015-NGPA, Barker-2016-ACBT}. 
\cite{Lopez-2013-ABDM} extractes multiple sharpness features. In~\cite{Kothari-2013-ETFA}, the author extractes 461 quantitative image features capturing the texture, color, shape, and topological 
properties of a histopathological image. \cite{Cooper-2015-NGPA} extracts precise quantitative 
morphometric features. There are shape features in the core feature group extracted 
by~\cite{Barker-2016-ACBT}.

\subsection{Deep Learning Feature Extraction}
Convolutional Neural Network (CNN) is widely used to extract the deep learning features in various WSI analysis tasks. In the papers, a total of 53 papers used CNN for deep learning feature 
extraction~\cite{Puerto-2016-DPAW, Sharma-2016-DCNN, Wang-2016-DLIM, Geccer-2016-DCBC, Sirinukunwattana-2016-LSDL, Hou-2016-PCNN, Sheikhzadeh-2016-ALMB, Cruz-2017-ARIB, Wollmann-2017-AGBC, Araujo-2017-CBCH, Bejnordi-2017-CASC, Sharma-2017-DCNN, Korbar-2017-DLCC, Jimenez-2017-DMCR, Xu-2017-LSTH,Korbar-2017-LUHD, Ghosh-2017-SLCA, Das-2017-CHWS, Cui-2018-DLAO, Zanjani-2018-CDHW, Courtiol-2018-CDLH, Kumar-2018-DBFR, Bychkov-2018-DLTA, Gecer-2018-DCCW, Ren-2018-DPCP, Tellez-2018-GWSI, Sirinukunwattana-2018-IWSS, Kwok-2018-MCBC, Das-2018-MILD, Lin-2018-SFDS, Campanella-2018-TSDM, Jiang-2018-TMDD, Wang-2018-WSLW, Shou-2018-WSIC, Tellez-2018-WSMD, Bertram-2019-LSDM, Li-2019-AMRM, Liu-2019-AIBC, Campanella-2019-CGCP, Yue-2019-CCOP, Maksoud-2019-CCOR, Bilaloglu-2019-EPCW, Lin-2019-FSFD, Yu-2019-LSGC, Sanghvi-2019-PAIA, Wang-2019-RRMI, Kohlberger-2019-WSIF, Xu-2019-MTPW, Chen-2020-ITWS, Sornapudi-2020-CWSH, Pantanowitz-2020-AIAP}.

The basic configuration of CNN is the convolutional layer, pooling layer and fully connected layer ~\cite{Rahaman-2020-ICSC}.  
These three layers can be stacked. Take the input of the previous layer as the output of the 
next layer, and finally get $N$ feature maps ~\cite{Rawat-2017-DCNN} with very low dimensions. 
Because it is an end-to-end learning model, it can learn more fully and extract features 
better~\cite{Zhiqiang-2017-ROBC}. The convolutional layer acts as a feature extractor, and the 
neurons in the convolutional layer are arranged into feature maps. Since different feature maps 
in the same convolution have different weights, $N$ features can be extracted at each 
position~\cite{Lecun-1998-GLAD}~\cite{Lecun-2015-DL}.

\paragraph{Deep Learning Features of the VGG Series}~{}
\\ 

\begin{figure}[htbp!]
\centerline{\includegraphics[width=0.98\linewidth]{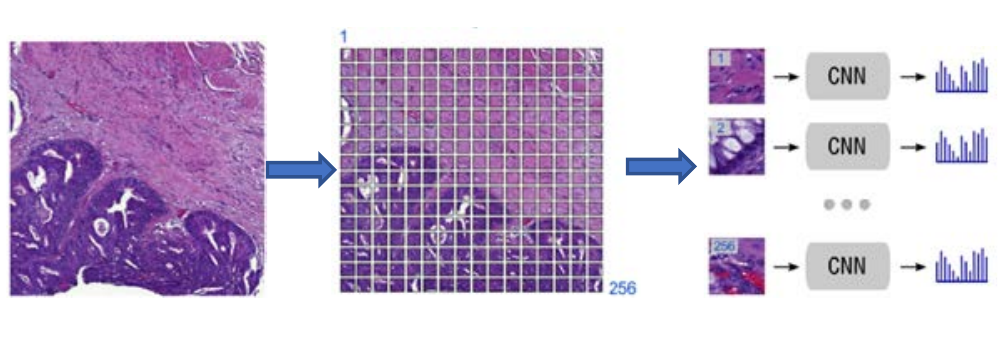}}
\caption{ The VGG-16 network produces a high-dimensional feature vector for each individual 
tile from an input image. This figure corresponds to Fig.1 in~\cite{Bychkov-2018-DLTA}. }
\label{fig:VGGF}
\end{figure}

In CNN, several classical improved network structures are often applied to extract deep 
features on WSI. VGGNet is an improvement based on the original framework 
of~\cite{Krizhevsky-2017-ICDC}. The full name of VGG is Visual Geometry Group, which belongs 
to the Department of Science and Engineering of Oxford University. It can be applied to 
face recognition, image classification, etc. VGGNet increases the network depth by adding 
more convolutional layers and fixing other parameters of the network 
framework~\cite{Simonyan-2014-VDCN}. All layers use $3 \times 3$ convolution filters that 
there are fewer parameters and lower cost. Among the papers we have summarized, the papers 
that use VGGNet to extract deep learning features are~\cite{Kumar-2018-DBFR, Bychkov-2018-DLTA, Campanella-2018-TSDM, Wang-2018-WSLW, Li-2019-AMRM, Yue-2019-CCOP, Lin-2019-FSFD}. The 
process of VGG extracting features in ~\cite{Bychkov-2018-DLTA} is shown in Fig.~\ref{fig:VGGF}.                                     

\paragraph{Deep Learning Features of the ResNet Series}~{}
\\ 

ResNet is another widely used CNN structure. The full name of ResNet is Residual Network 
and the proponent is Balduzzi D. ResNet has pushed deep learning to a new level, reducing the 
error rate to a level lower than that of humans for the first time. The residual module in 
ResNet makes the network deeper, but with lower complexity. It also makes the network easier 
to optimize and solves the problem of vanishing gradient~\cite{He-2016-DRLI}. The bottlen 
neck layer in ResNet uses 1$×$1 networks, which expands the dimension of the featuremap and 
greatly reduces the amount of calculation~\cite{He-2016-IMDR}. Among the papers we reviewed, 
the papers that use ResNet to extract deep learning features 
include~\cite{Korbar-2017-DLCC, Korbar-2017-LUHD, Kwok-2018-MCBC, Campanella-2018-TSDM, Jiang-2018-TMDD, Shou-2018-WSIC, Bertram-2019-LSDM, Xu-2019-MTPW}.

\paragraph{Deep Learning Features of the U-net Series}~{}
\\ 
 
The full name of U-net is Unity Networking. it is a network architecture established to 
solve the problem of medical image segmentation. This structure is based on FCN (Fully 
Convolutional Neural Network). It adds an upsampling stage and adds many feature channels, 
allowing more original image texture information in high-resolution layers, using valid 
for convolution throughout, ensuring that the results obtained are based on no missing 
context features~\cite{Ronneberger-2015-UCNB}. In the paper we summarized, the use of U-net for deep learning feature extraction have~\cite{Bandi-2017-CDMT, Seth-2019-ALBD, Seth-2019-ASDW, Feng-2020-DLMA}.

\paragraph{Other Deep Learning Feature}~{}
\\ 

There are other improved structures based on CNN, such as GoogLeNet~\cite{Wang-2016-DLIM}. 
The Google Academic team carefully prepares GoogLeNet to participate in the ILSVRC 2014 competition. The main idea is to approximate the optimal sparse structure by building a dense block structure to improve performance without increasing the amount of calculation. The initial version of GoogLeNet appeared in~\cite{Szegedy-2015-GDC}.

There are also other improved structures based on Recurrent Neural Networks 
(RNN)~\cite{Maksoud-2019-CCOR} and LSTM ~\cite{Ren-2018-DPCP}, which are used to extract 
deep learning features. RNN appeared in the 1980s, and its prototype has seen in the 
Hopfield neural network model proposed by American physicists in 1982~\cite{Hopfield-1982-NNPS}. 
RNN has a strong processing ability for variable length sequence data. Therefore, it is very 
effective for data with time-series characteristics and can mine time series information and 
language information in the data. The Long Short-Term Memory (LSTM) model appeared because of 
the drawbacks of RNN, and LSTM can solve the problem of gradient disappearance in 
RNN~\cite{Hochreiter-1997-LSTM}.

\subsection{Summary}
It can be seen from the content we reviewed above, in the traditional feature extraction, 
color feature, texture feature, and shape feature are the three most commonly used features. 
The texture feature is the most used. In the papers we summarized, from 2004 to 2019, a total 
of 51 papers used texture features. The second is color features, which are generally based 
on the three color spaces of RGB, HSV, and LAB. Among them, the RGB color space is the most 
commonly used. The least applied is the shape feature. For more details and analysis in this 
regard, see the detailed introduction in the following chapters.

Over times, the level of science and technology has also continuously improved. As can be 
seen from the papers we summarized, since 2016, deep learning features have been gradually 
applied to this day. The specific deep learning network architecture will be introduced in a 
separate method analysis later. Table.~\ref{CDFE} is a summary of the CAD methods used for 
feature extraction in WSI. 
\onecolumn
\begin{center}\tiny
\renewcommand\arraystretch{1.85}
\setlength{\tabcolsep}{0.001pt}
\newcommand{\tabincell}[2]
{\begin{tabular}{@{}#1@{}}#2\end{tabular}}
\centering
\begin{longtable}{cccccc}
\caption{Summary of the CAD methods used for feature extraction in WSI (Traditional
(T)).} \\
\endfirsthead
\caption[l]{Continue: Summary of the CAD methods used for feature extraction in WSI.}\\
\hline
\endhead
 \hline
\endfoot
\hline
Feature Type&Method  & Reference                  & Year & Team                                                   & \multicolumn{1}{c}{Details}                                                                             \\ \hline
T&Color        & ~\cite{Kong-2009-CAEN}             & 2009 & J Kong, O Sertel                                       & \tabincell{c}{Combined Color and entropy\\ information extracted from RGB\\ and LAB image channels                     } \\
T&Color        & ~\cite{Roullier-2010-GMRS}         & 2010 & V. Roullier                                            & RGB feature vector                                                                                 \\
T&Color        & ~\cite{Samsi-2012-ECFA}            & 2012 & Siddharth Samsi                                        & Hue channel for HSV color space conversion                                                              \\
T&Color        & ~\cite{Kothari-2012-BIMP}          & 2012 & Sonal Kothari                                          & \tabincell{c}{RGB, CIELAB, HSV, mean and standard \\ deviation in each channel in three spaces                       }  \\
T&Color        & ~\cite{Akakin-2012-CBMI}           & 2012 & Hatice Cinar Akakin                        & Nuclear RGB features                                                                                    \\
T&Color        & ~\cite{Collins-2013-AMAU}          & 2013 & Brian T. Collins                                       & color information                                                                                       \\
T&Color        & ~\cite{Nayak-2013-CTHS}            & 2013 & Nandita Nayak                                          & \tabincell{c}{The average, standard deviation,\\ minimum and maximum values of\\ each color channel \\in the RGB space}    \\
\end{longtable}
\end{center}

\section{Segmentation Methods}
\label{s:SM} 
In recent years, with the increasing size and quantity of medical images, computers must facilitate processing and analysis. In particular, computer algorithms for 
delineating anatomical structures and other areas of interest are becoming increasingly 
important in assisting and automating specific histopathological tasks. These algorithms 
are called image segmentation algorithms~\cite{Pham-2000-CMMI}.

Image segmentation refers to the process of dividing a digital image into multiple segments, 
namely a set of pixels. The pixels in a region are similar according to some homogeneity 
criteria (such as color, intensity, or texture), to locate and identify objects and 
boundaries in the image ~\cite{Gonzalez-2007-IP}. The practical applications of image 
segmentation include: filtering noise images, medical applications (locating tumors and 
other pathologies, measuring tissue volume, computer-guided surgery, diagnosis, treatment 
planning, anatomical structure research) ~\cite{Patil-2013-MISR}, locating objects in 
satellite images (roads, forests, etc.), facial recognition, fingerprint recognition, etc. 
The selection of segmentation techniques and the level of segmentation depends on the 
specific type of image and the characteristics of the problem being considered~\cite{Dass-2012-IST1}.

In the process of medical image segmentation, the details required in the segmentation 
process largely depend on the clinical application of the 
problems~\cite{Masood-2015-SMIS}~\cite{Zuva-2011-ISAT}. The purpose of segmentation is to improve 
the visualization process to deal with the detection process more effectively. Medical image 
segmentation is faced with many problems because the quality of the segmentation process is 
affected~\cite{Shrimali-2009-CTSM}. When there is noise in the image, there will be uncertainty, 
which makes it difficult to classify the image~\cite{Birkfellner-2016-AMIP}. The reason is that 
the intensity value of the pixel has been modified due to noise in the image. Such a change in 
pixel intensity value will disturb the uniformity of the image intensity range~\cite{Al-2010-CSRN}. 
Therefore, to deal with this uncertainty, image segmentation plays a crucial role in medical 
diagnostic systems~\cite{He-2013-MIS}.

\begin{figure}[htbp!]
\centerline{\includegraphics[width=0.65\linewidth]{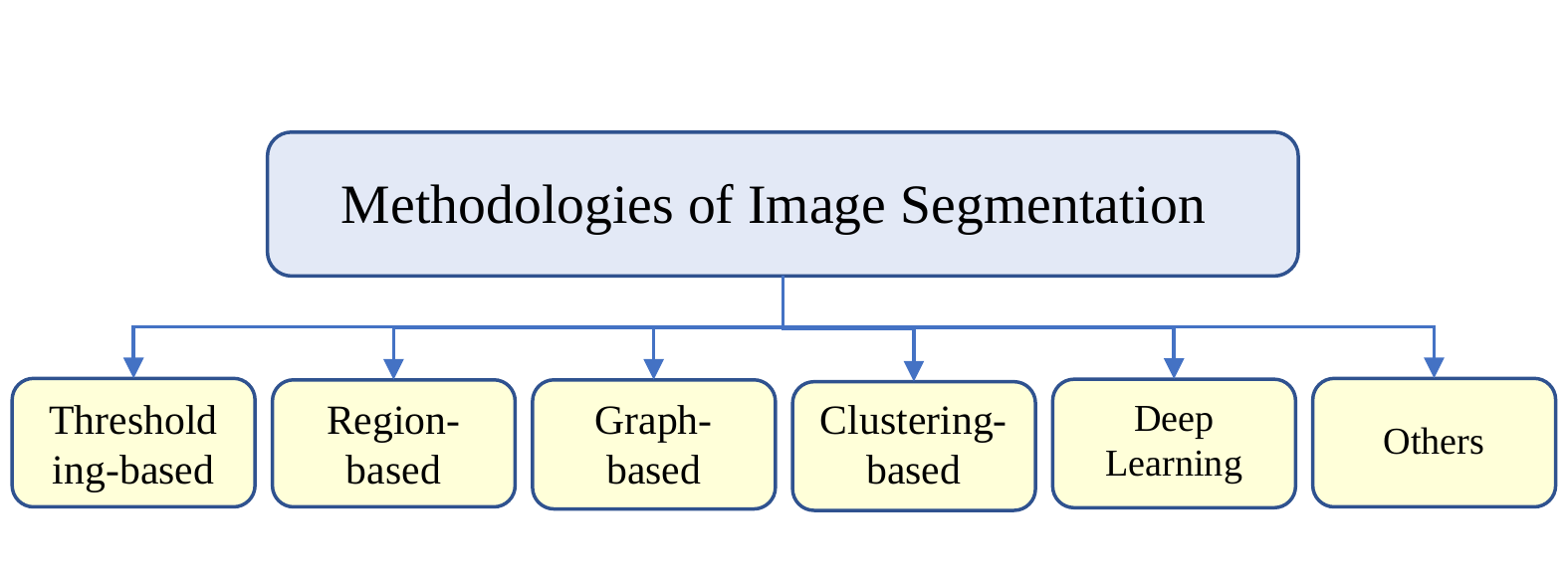}}
\caption{The composition of the segmentation method in WSI.}
\label{fig:structureseg}
\end{figure}

As a crucial step in CAD pathologists, segmentation techniques have flourished in recent years. 
As shown in Fig.~\ref{fig:trend}, from 2010 to 2020, the number of papers using segmentation WSI 
technology to assist doctors in diagnosis has increased from 2 to 28. According to the papers we have reviewed, 
segmentation is divided into five different techniques including thresholding-based, region-based, 
graph-based, clustering-based, deep learning, and other image segmentation methods. Its composition 
and structure diagram are shown in Figure.~\ref{fig:structureseg}.

\subsection{Thresholding based Segmentation Method}
Threshold segmentation is a classic method in image segmentation. It uses the difference 
in grayscale between the target and the background to be extracted in the image, and divides 
the pixel level into several categories by setting a threshold to achieve the separation of 
the target and the background~\cite{Fu-1981-SIS, Haralick-1985-IST}. The threshold segmentation 
method is simple to calculate, and can always use closed and connected boundaries to define 
non-overlapping regions. Images with a strong contrast between the target and the background 
can provide better segmentation effect~\cite{Sahoo-1988-STT}.

Among them, the selection of the optimal threshold is a significant issue. Commonly used 
threshold selection methods are: manual experience selection method, histogram 
method~\cite{Rosenfeld-1983-HCAA}, maximum between-class variance method 
(OTSU)~\cite{Ohtsu-1978-DLST}, adaptive threshold method. Because the threshold segmentation 
method is simple to implement, the amount of calculation is small, and the performance is 
relatively stable, it has become the most in image segmentation. As the basic and most widely 
used segmentation technology, it has been used in many fields. Among the reviewed papers, five 
are based on threshold based segmentation~\cite{Kong-2011-IMAG, Lu-2012-ASAE, Shu-2013-SOCN, Vo-2016-CWSI, Arunachalam-2017-CAIS}.

In~\cite{Kong-2011-IMAG}, an image analysis tool for segmentation and characterization of 
cell nuclei is developed. The microscopic images of glioblastoma from the TCGA project are used. 
To reliably identify cell nuclei, a fast hybrid gray-scale reconstruction algorithm is applied 
to the image to normalize the background area degraded by artifacts produced by tissue preparation 
and scanning~\cite{Vincent-1993-MGRI}. This operation separates the foreground from the 
normalized background and allows simple threshold processing to identify the nucleus.

In~\cite{Lu-2012-ASAE}, a computer-aided technique is proposed for segmentation and analysis 
of the whole slide skin histopathological images. Before using the segmentation technique, 
determine the single-color channel that provides good discrimination information between the 
epidermis and dermis regions. Then multi-resolution image analysis is used in the proposed 
segmentation technique. First, a low-resolution image of the WSI is generated. Then, the 
global threshold method and shape analysis to segment low-resolution images are used. Based 
on the segmented skin area, the layout of the skin is determined, and a high-resolution image 
block of the skin is generated for further manual or automatic analysis. Experiments on 16 
different whole slide skin images show that the technology has high performance, $92\%$ 
sensitivity, $93\%$ accuracy, and $97\%$ specificity are achieved.

In~\cite{Shu-2013-SOCN}, a new method is proposed to segment severely aggregated overlapping 
cores. The proposed method first involves applying a combination of global and local thresholds 
to extract foreground regions.

In~\cite{Vo-2016-CWSI} and~\cite{Vo-2019-MCMH}, a highly scalable and cost-effective image 
analysis framework based on MapReduce is proposed, and a cloud-based implementation is provided. 
The framework adopts a grid-based overlap segmentation scheme and provides parallelization of 
image segmentation based on MapReduce. In the segmentation step, a threshold method is applied 
to segment the nucleus.

\begin{figure}[htbp!]
\centerline{\includegraphics[width=0.9\linewidth]{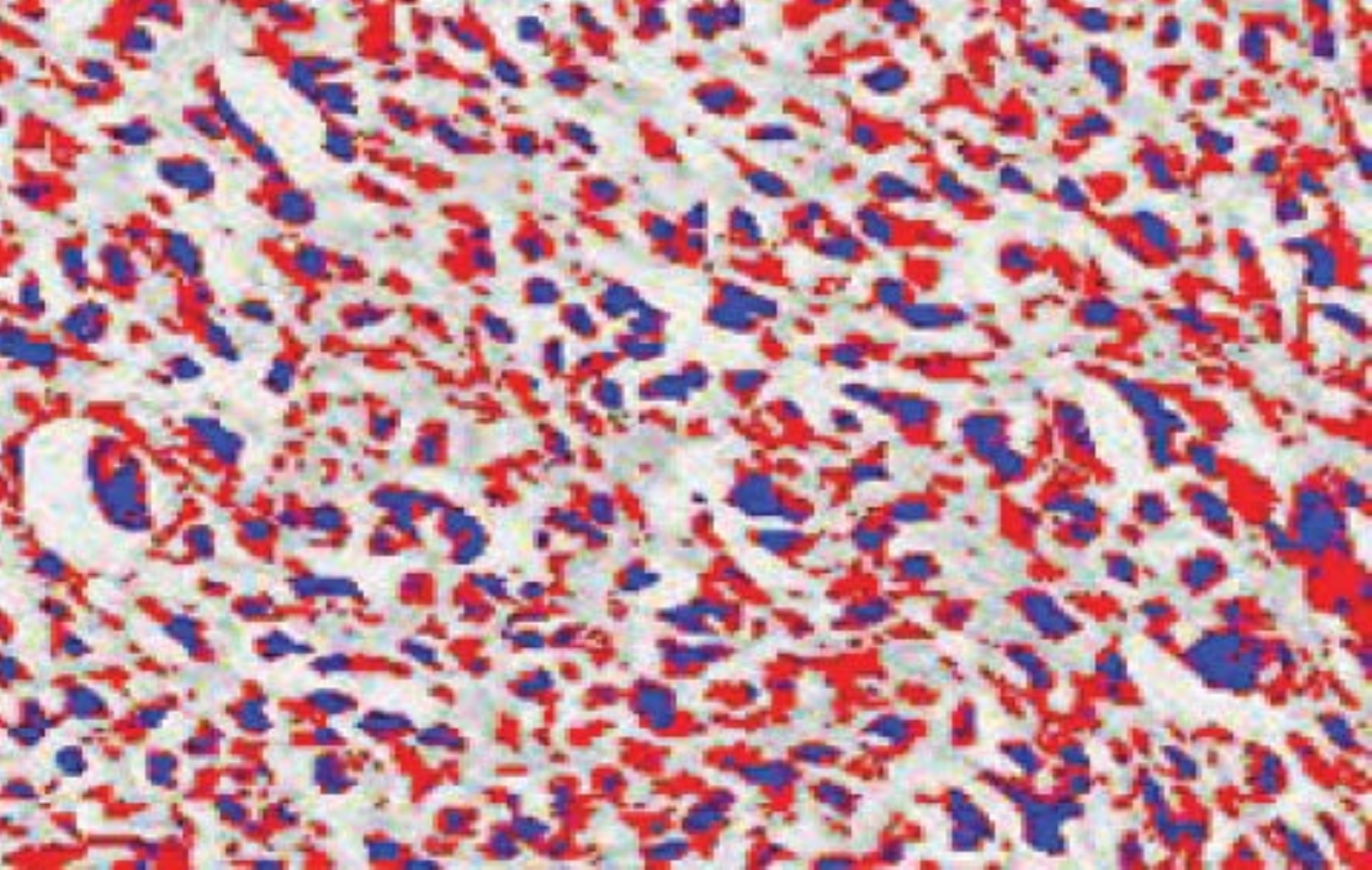}}
\caption{ Otsu output showing more blue color. 
This figure corresponds to Fig.3 in~\cite{Arunachalam-2017-CAIS}.}
\label{fig:CAIS}
\end{figure}

In~\cite{Arunachalam-2017-CAIS}, the segmentation of tumor and non-tumor areas on the WSIs 
datasets of osteosarcoma histopathology. The method in this article combines pixel-based and 
object-based methods, using tumor attributes, such as nucleus clusters, density, and circularity, 
and using multi-threshold Otsu segmentation technology to further classify tumor regions as live 
and inactive. The pan-fill algorithm clusters similar pixels into cell objects and calculates the cluster data to analyze the studied area further. The final experimental results 
show that for all the sampled datasets used, the accuracy of the method in question in identifying live tumors and coagulative necrosis is $100\%$, while the accuracy of fibrosis 
and acellular/low cell tumors is about $90\%$. The WSI effect after multi-threshold Otsu segmentation 
is shown in Figure.~\ref{fig:CAIS}.

\subsection{Region-based Segmentation Method}
Region-based segmentation is a kind of segmentation techniques based on directly finding 
the region. In fact, similar to the boundary-based image segmentation technology, it uses the similarity of the object's gray distribution and background. Generally, 
region-based image segmentation methods contain two categories: watershed segmentation 
and region growing.

\paragraph{Watershed Segmentation}~{}
\\ 

The watershed algorithm draws on the theory of morphology and is a region-based image 
segmentation algorithm. In this method, an image is regarded as a topographic map, and the 
gray value corresponds to the height of the terrain. High gray values correspond to mountains, 
and low gray values correspond to valleys. If rain falls on the surface, the low-lying area 
is a basin, and the ridge between the basins is called a watershed. Watershed is equivalent 
to an adaptive multi-threshold segmentation algorithm~\cite{Vincent-1991-WDSE}.

\begin{figure}[htbp!]
\centerline{\includegraphics[width=0.5\linewidth]{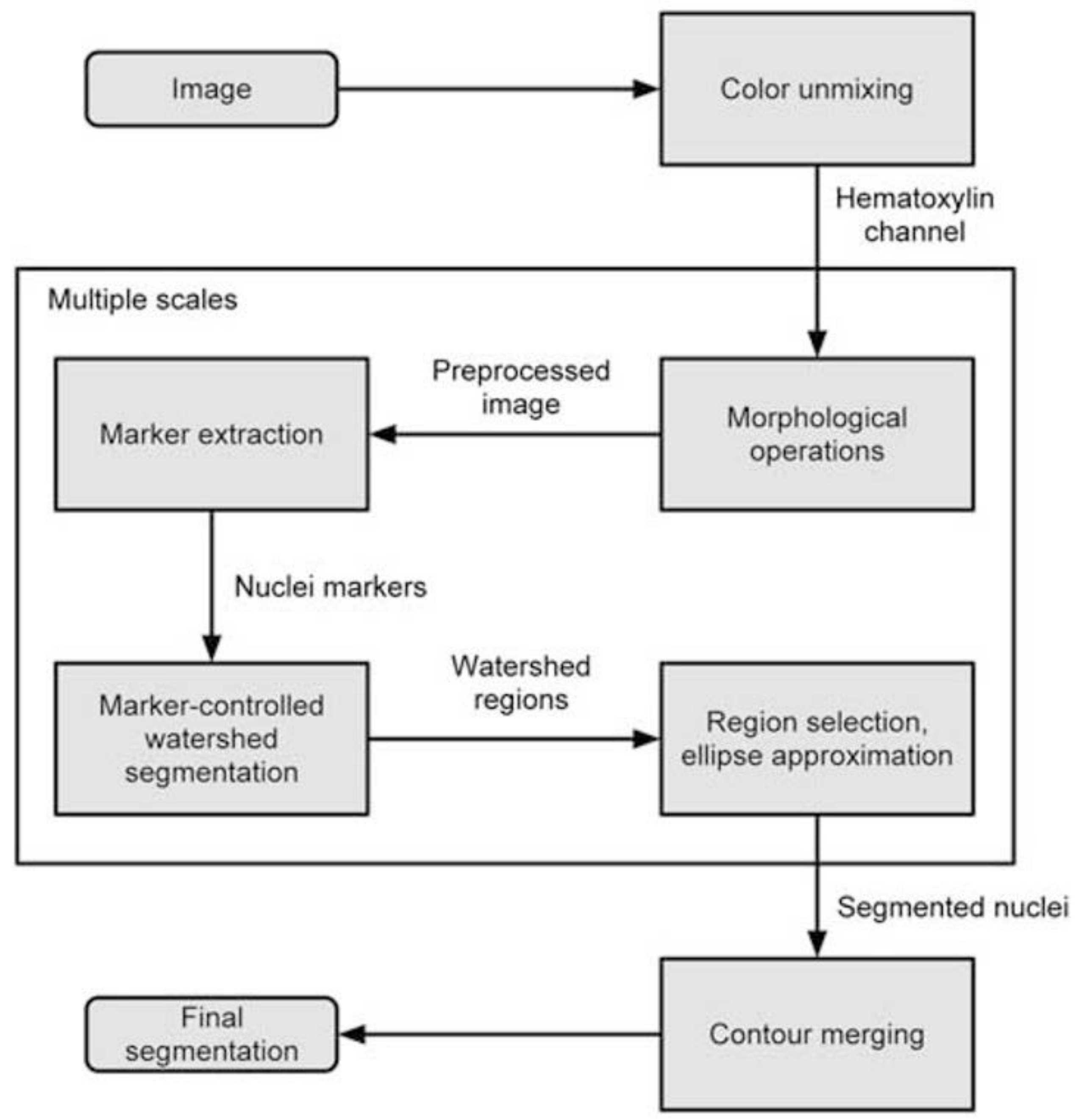}}
\caption{Overview of the automatic nuclei segmentation method. 
This figure corresponds to Fig.1 in~\cite{Veta-2012-PVAE}.}
\label{fig:PVAE}
\end{figure}

In~\cite{Kong-2011-IMAG}, overlapping nuclei are separated using the watershed method. 
In~\cite{Veta-2012-PVAE}, an automatic cell nucleus segmentation algorithm is used to extract 
size-related morphometric features of cell nuclei and analyze their prognostic value in male 
breast cancer. The segmentation process consists of four main steps: preprocessing, watershed 
segmentation controlled by multi-scale markers, post-processing, and merging of multi-scale 
results. The overall process of this automatic segmentation method is shown in the 
Figure.~\ref{fig:PVAE}. In~\cite{Veta-2013-ANSH}, the same automatic segmentation method 
as in ~\cite{Veta-2012-PVAE} is used in H\&E stained breast cancer histopathology images.

\begin{figure}[htbp!]
\centerline{\includegraphics[width=0.5\linewidth]{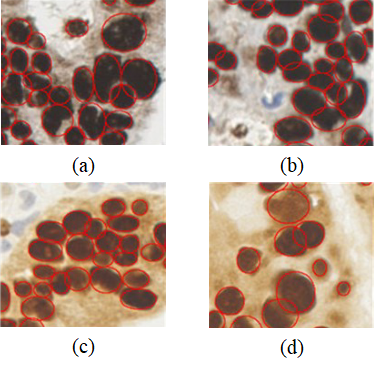}}
\caption{Some segmentation results. 
This figure corresponds to Fig.7 in~\cite{Shu-2013-SOCN}.}
\label{fig:SOCN}
\end{figure}

In ~\cite{Shu-2013-SOCN}, to segment the overlapping nuclei gathered in the foreground region, 
seed markers are obtained using morphological filtering and intensity-based region growth. 
Then the seed watershed and separate the aggregated nuclei are applied. Finally, a post-processing 
step of identifying positive nuclear pixels is added to eliminate false pixels. Some segmentation 
results are shown in Fig.~\ref{fig:SOCN}. In~\cite{Vo-2016-CWSI} and~\cite{Vo-2019-MCMH}, 
watershed technology is used to separate overlapping nuclei in objects is used.

\paragraph{Region Growing}~{}
\\ 

Region growing is an image segmentation method of serial region segmentation. Region growth 
refers to starting from a certain pixel and gradually adding neighboring pixels according 
to specific criteria. When certain conditions are met, the regional growth is terminated, that is, 
The region's growth depends on the selection of the initial point (seed point), growth criteria, and termination conditions~\cite{Hojjatoleslami-1998-RGNA}.

The region growing is relatively a common method. It can achieve the best performance 
when there is no prior knowledge available, and it can be used to segment more complex images. 
However, the regional growth method is iterative, and space and time costs are relatively high ~\cite{Adams-1994-SRG}. Among the WSI-based segmentation tasks we have 
summarized, there is one paper related to region growth~\cite{Shu-2013-SOCN}.

\subsection{Graph-based Segmentation Method}
Graph-based segmentation is a classic image segmentation algorithm. The algorithm is a greedy 
clustering algorithm based on the graph. Its advantages include simple implementation and 
faster speed~\cite{Felzenszwalb-2004-EGIS}. Many popular algorithms are based on this method~\cite{Van-2011-SSSO}.

Graph-based segmentation first expresses image as a graph in graph theory, so that, each point 
in a pixel is regarded as a vertex $v_{i} \in V$, and each pixel and 8 adjacent pixels (eight 
neighborhoods) form a graph edge $e_{i} \in E$, so a graph $G = (V,E)$ is constructed. The 
weight of each side of the graph is the relationship between the pixel and the neighboring 
pixels, which expresses the similarity between the neighboring pixels. Treat each node (pixel) 
as a single area, and then merge it according to the parameters of the area and the internal 
difference to get the final segmentation~\cite{West-1996-IGH}.

Because WSIs are usually very large, they are stored as pyramids of tiled images, so that they 
can be processed in a hierarchical manner, that is, low-resolution to determine the area of 
interest, high-resolution image classification, top-down segmentation. Among the papers we reviewed, 
there are three papers that combine graph-based segmentation methods with 
multi-resolution~\cite{Roullier-2011-MRGA, Roullier-2010-GMRS, Roullier-2010-MEBC}.

\begin{figure}[htbp!]
\centerline{\includegraphics[width=0.98\linewidth]{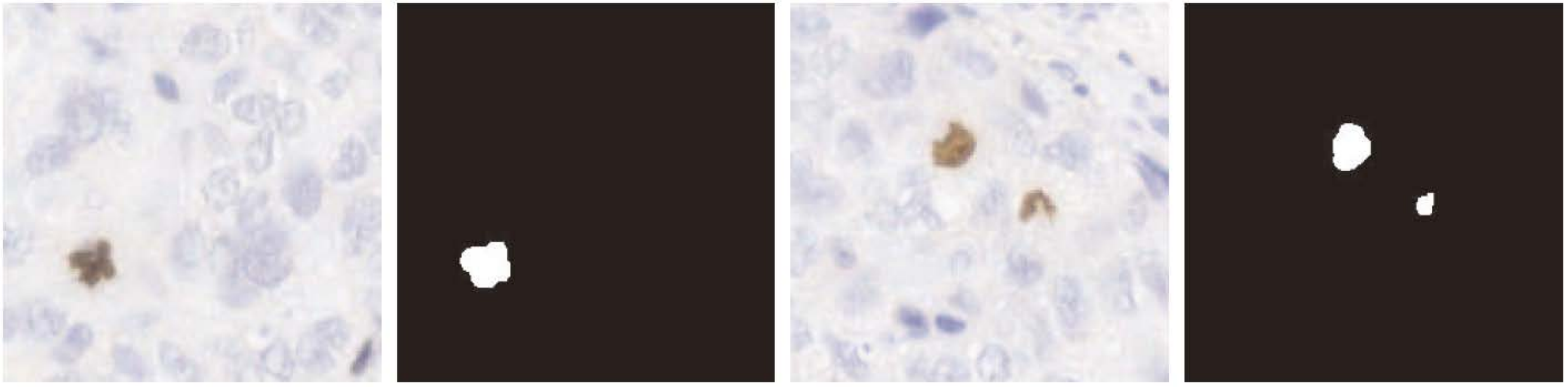}}
\caption{The composition of the segmentation method in WSI. 
This figure corresponds to Fig.3 in~\cite{Roullier-2010-GMRS}}
\label{fig:GMRS}
\end{figure}

In ~\cite{Roullier-2010-GMRS}, \cite{Roullier-2010-MEBC} and~\cite{Roullier-2011-MRGA}, the 
mitosis in WSI of breast tissue is extracted. The image is simplified by discrete regularization, 
and clustering is performed by unsupervised 2-mean clustering. Clustering is performed in 
a specific area divided at the previous resolution. The obtained clusters are expanded with 
finer resolution levels through pixel duplication and refined in specific areas. At the last 
resolution level, the mitotic figure is extracted. The segmentation result of mitosis 
in~\cite{Roullier-2010-GMRS} is shown in the Figure.~\ref{fig:GMRS}.

\subsection{Clustering-based Segmentation Method}
Clustering is based on the relationship between each pixel and neighboring pixels. 
If a pixel is similar in color, texture, or gray to its neighboring pixels, 
then they will be merged into the same class.

$k$-means algorithm is more commonly used in clustering algorithms. The cluster center 
point is obtained by calculating the average of all pixels in each cluster in the sample. 
The basic working principle of the $k$-means algorithm is to receive the parameter $k$ 
input by the user, divide the given $n$ data sample points into $k$ groups on average, 
take the input $k$ points as the cluster centers to be converged, and calculate the other 
clusters. The Euclidean distance from the sampling point to the $k$ convergence centers. 
And compare the distance between all sampling points and the convergence center point. 
The classification is made by comparing the minimum form of Euclidean distance. Then after 
repeated iterations, the mean value of $k$ clusters is successively obtained. Until the performance criterion function, clustering is the best, the overall error is the smallest, and the best clustering effect is obtained~\cite{Macqueen-1967-SMCA}.

Among the papers we reviewed, the papers on $k$-means clustering and segmentation 
are~\cite{Hiary-2013-SLWS} and~\cite{Arunachalam-2017-CAIS}. In~\cite{Hiary-2013-SLWS}, 
the author uses $k$-means unsupervised learning for WSI segmentation, which produces a highly 
robust correctness result equivalent to supervised learning, which is $95.5\%$ accuracy. 
In~\cite{Arunachalam-2017-CAIS}, the $k$-means clustering technique with color normalization 
is used for tumor separation.

\subsection{Deep Learning based Segmentation}
Among the papers we reviewed, 10 papers used deep learning methods for WSI 
segmentation~\cite{Bandi-2017-CDMT, Xu-2017-LSTH, Dong-2018-RANT, Cui-2018-DLAO, Sirinukunwattana-2018-IWSS, Mehta-2018-LSBB, Vo-2019-MCMH, Seth-2019-ALBD, Seth-2019-ASDW, Feng-2020-DLMA}.

\begin{figure}[htbp!]
\centerline{\includegraphics[width=0.98\linewidth]{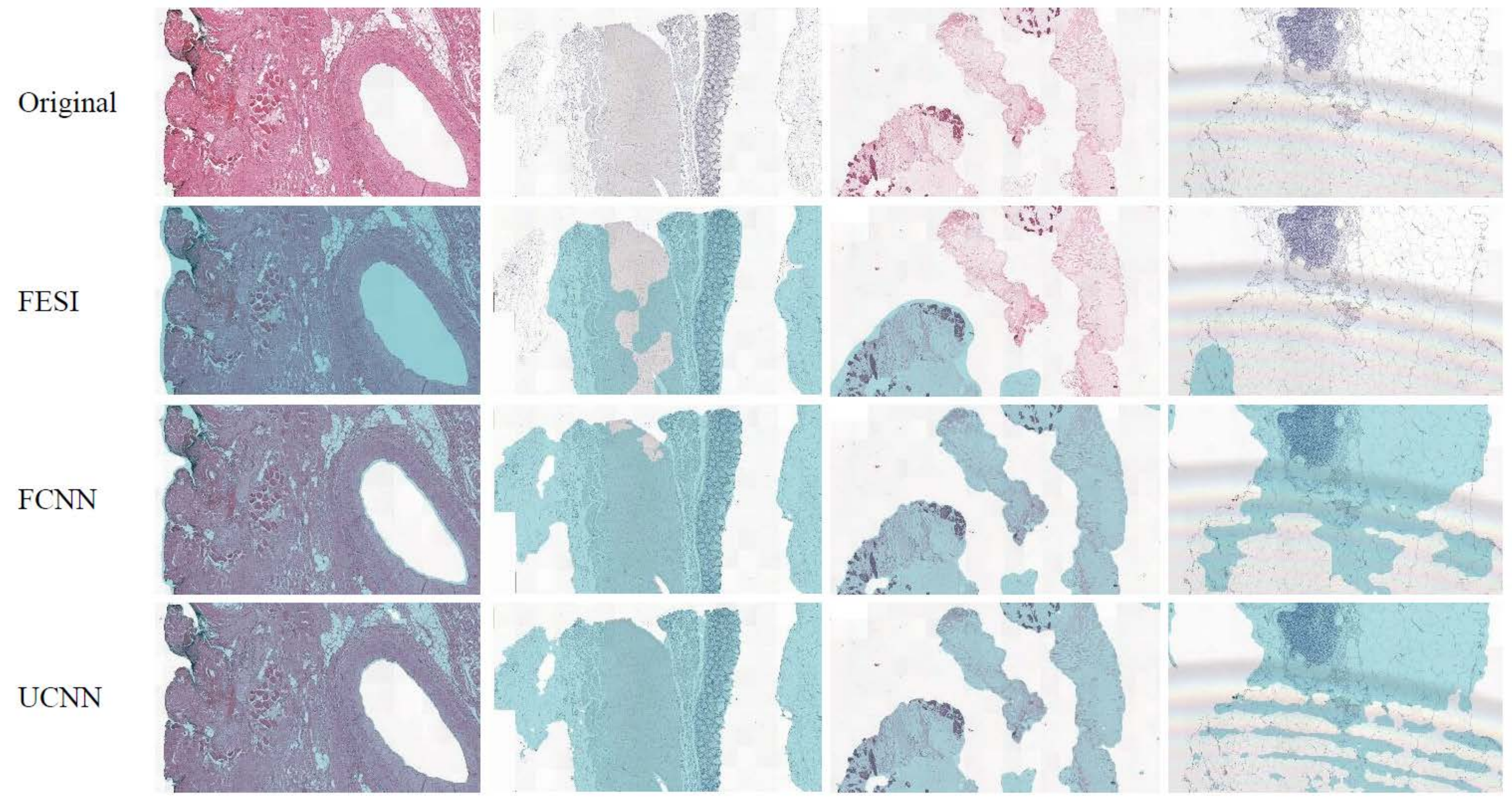}}
\caption{Qualitative results for the different algorithms. 
This figure corresponds to Fig.2 in~\cite{Bandi-2017-CDMT}}
\label{fig:CDMT}
\end{figure}

In~\cite{Bandi-2017-CDMT}, two different CNN structures, FCN and U-net, are used to segment 
and accurately identify tissue slices. Here, the two methods are compared with the traditional 
foreground extraction (FESI) algorithm based on structural information. These three methods 
are applied to 54 WSIs, and the average value of the Yakoka index and the standard deviation 
of the Yakoka index are used for evaluation. The final U-net result is the best (Jaccard index 
is 0.937). The qualitative effects of different algorithms are shown in Figure.~\ref{fig:CDMT}.

In~\cite{Xu-2017-LSTH}, CNN-based ImageNet is used to extract features and convert them 
into histopathological images. And Support Vector Machine (SVM) is used to define segmentation 
as a classification problem. This method is applied to the digital pathology and colon cancer 
dataset of the MICCAI 2014 Challenge, and finally won the first place with $84\%$ test data accuracy.

In~\cite{Dong-2018-RANT}, a simple and effective framework called Reinforced Auto-Zoom Net 
(RAZN) is proposed, which considers the accurate and fast prediction of breast 
cancer segmentation. RAZN learns a strategy network to decide whether to zoom on a given 
area of interest. Because the amplification action is selective, RAZN is robust to unbalanced 
and noisy ground truth labels and effectively reduces overfitting. Finally, the method is 
evaluated on the public breast cancer dataset. It can be seen from the experimental results 
that RAZN is superior to single-scale and multi-scale baseline methods, and obtains better 
accuracy with lower inference cost.

In~\cite{Cui-2018-DLAO}, an automatic end-to-end deep neural network algorithm is proposed to 
segment the single core. The kernel-boundary model is introduced to predict the kernel and its 
boundary simultaneously using FCN. Given the color normalized image, the model directly outputs 
the estimated kernel map and boundary map. After post-processing, the final segmented core is 
produced. A method for extracting and assembling overlapping blocks is designed to seamlessly 
predict the cores in a large WSI. The final result proves the effectiveness of the data expansion 
method for cell nucleus segmentation tasks. The experiment shows that this method is superior to 
the prior art method and it is possible to accurately segment WSI within an acceptable time.

\begin{figure}[htbp!]
\centerline{\includegraphics[width=0.98\linewidth]{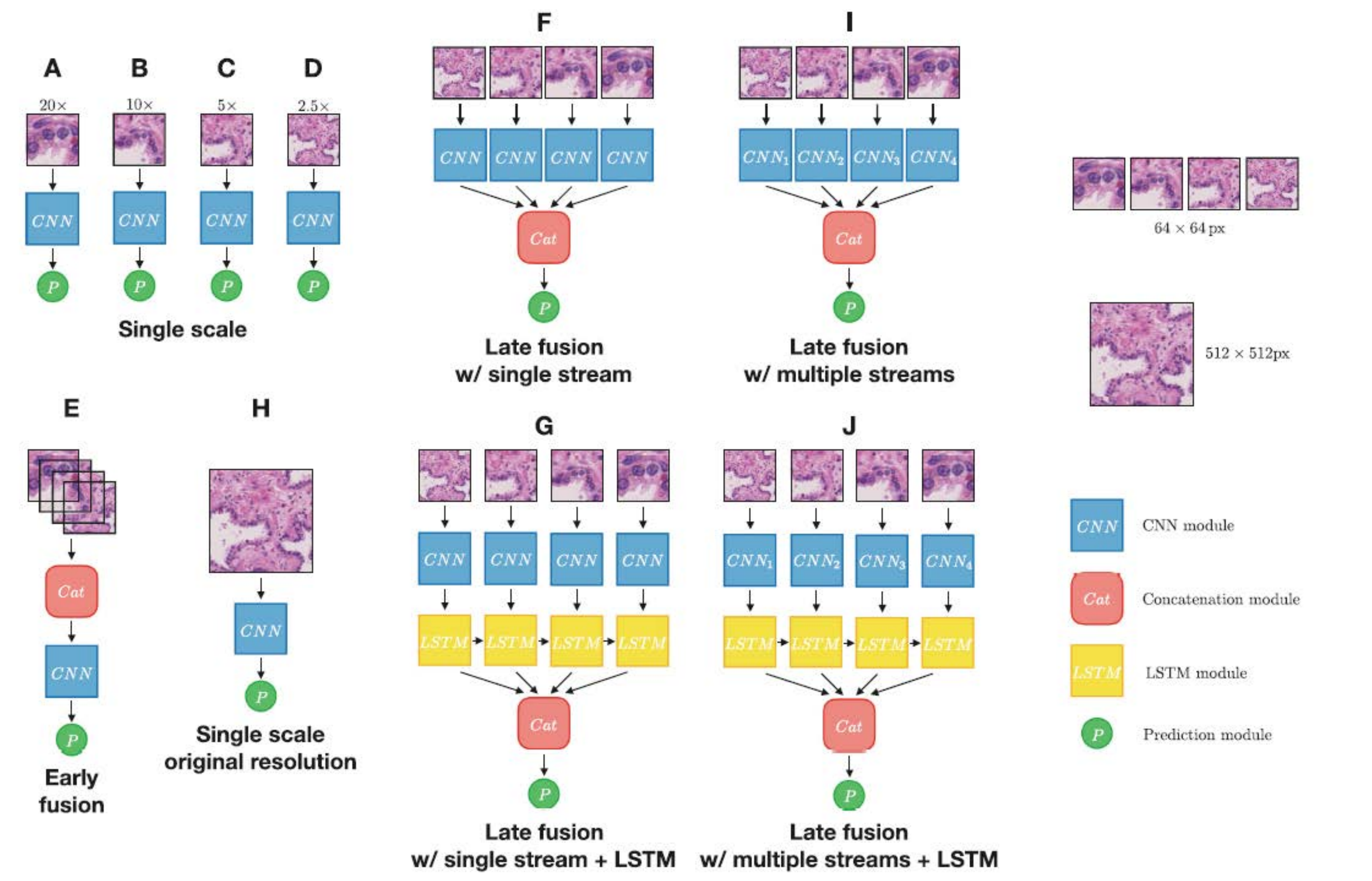}}
\caption{Used architectures. 
This figure corresponds to Fig.2 in~\cite{Sirinukunwattana-2018-IWSS}}
\label{fig:IWSS}
\end{figure}

In~\cite{Sirinukunwattana-2018-IWSS}, different architectures are systematically compared to 
evaluate how the inclusion of multi-scale information affects segmentation performance. The 
architectures are shown in Figure.~\ref{fig:IWSS}. It uses a public breast cancer dataset and 
a locally collected prostate cancer dataset. The result shows that the visual environment and 
scale play a vital role in the classification of histological images.

\begin{figure}[htbp!]
\centerline{\includegraphics[width=0.98\linewidth]{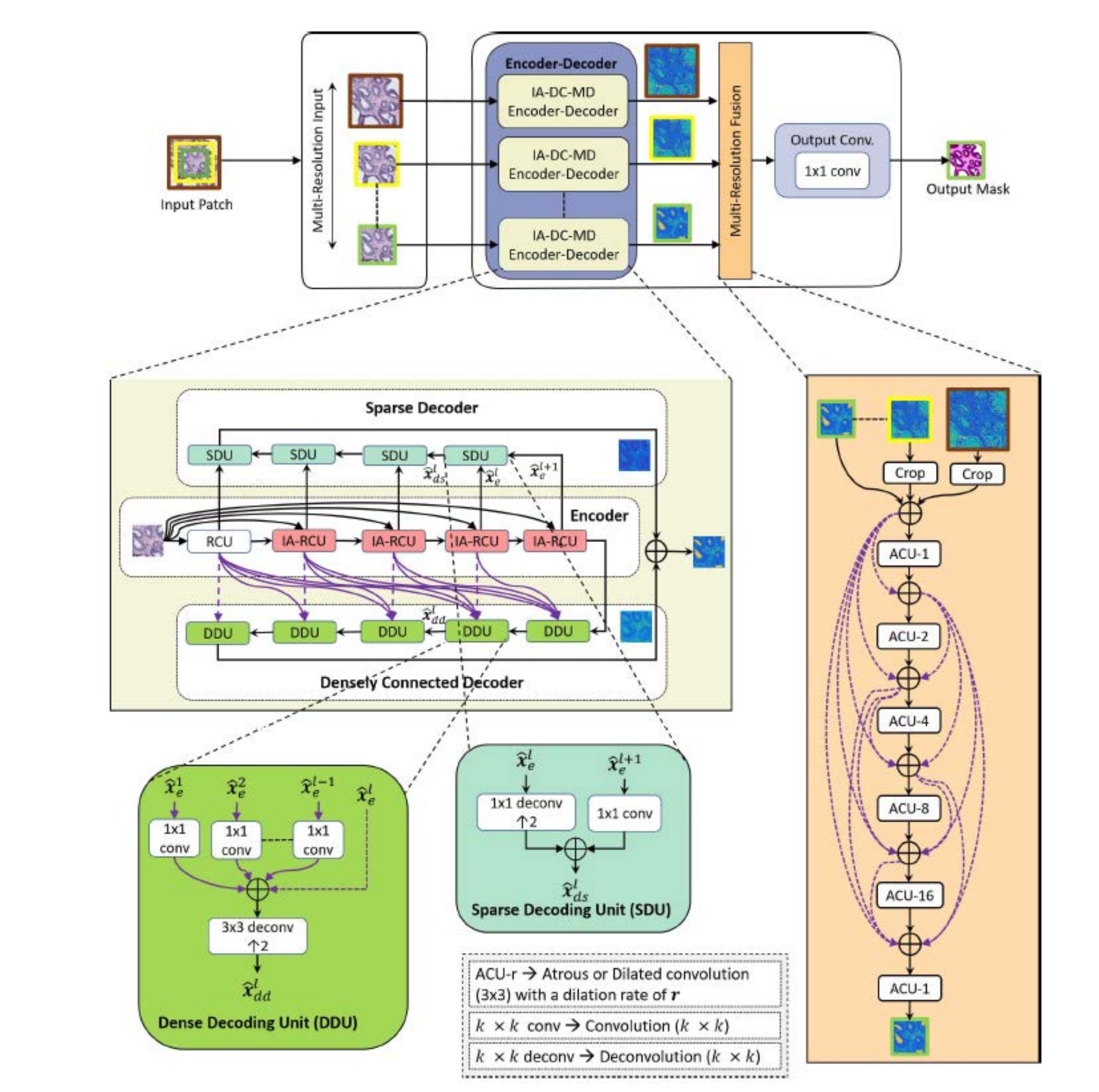}}
\caption{Multi-resolution encoder-decoder network structure diagram. 
This figure corresponds to Fig.4 in~\cite{Mehta-2018-LSBB}.}
\label{fig:LSBB}
\end{figure}

\begin{figure}[htbp!]
\centerline{\includegraphics[width=0.98\linewidth]{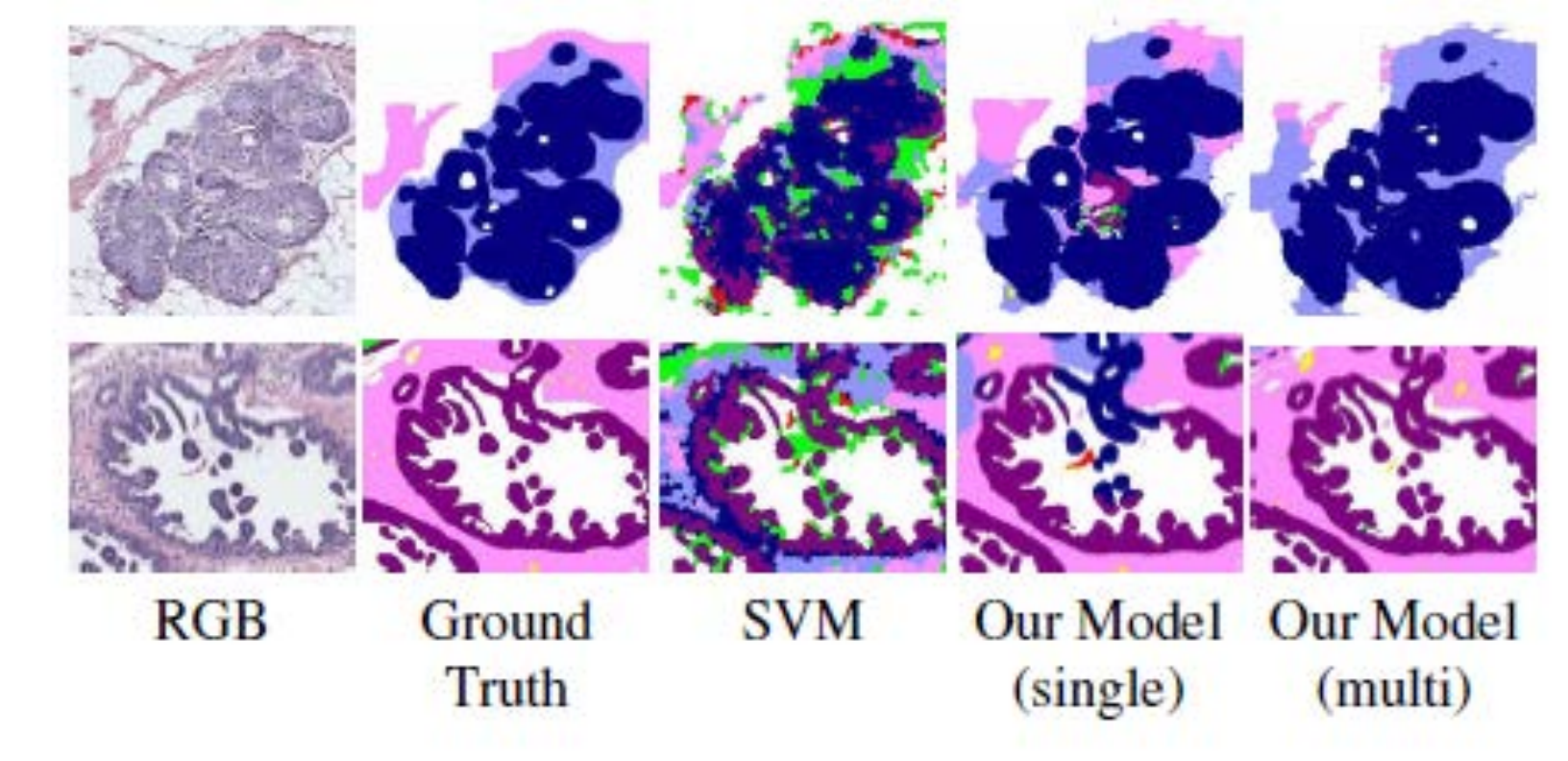}}
\caption{Segmentation result. 
The first line describes aggressive cases, 
while the second line describes benign cases. 
This figure corresponds to Fig.9 in~\cite{Mehta-2018-LSBB}.}
\label{fig:LSBBR}
\end{figure}

In~\cite{Mehta-2018-LSBB}, a new encoder-decoder architecture is proposed to solve the 
semantic segmentation problem of breast biopsy WSI. The designed new architecture contains 
four new functions: (1) Input Perceptual Coding Block (IA-RCU), which enhances the input 
inside the encoder to compensate for the loss of information due to down-sampling operations, 
(2) densely connected decoding network and (3) additional sparsely connected decoding network 
to effectively combine the multi-level features aggregated by the encoder, and (4) a 
multi-resolution network for context-aware learning, which uses densely connected fusion 
blocks to combine different resolutions rate output. The architecture is shown in 
Figure.~\ref{fig:LSBB}. The result after segmentation is shown in Figure.~\ref{fig:LSBBR}.

In~\cite{Seth-2019-ALBD} and~\cite{Seth-2019-ASDW}, several U-net architectures-DCNN designed 
to output probability maps are trained to segment ductal carcinoma in-situ (DCIS) WSI in 
wireless sensor networks and verified the minimum required to achieve excellent accuracy at 
the slide level good patch vision.  U-net is trained five times to achieve the best test results 
(DSC = 0.771, F1 = 0.601), which means that U-net benefits from seeing wider contextual 
information.

In~\cite{Feng-2020-DLMA}, a multi-scale image processing method is proposed to solve the 
segmentation problem of liver cancer in histopathological images. These eight networks are 
compared, and then the network most suitable for liver cancer segmentation is selected. 
Through a comprehensive comparison of performance, U-net is selected. The local color normalization method of pathological images is used to solve the influence of the background, 
and then a seven-layer Gaussian pyramid representation is established for each WSI to obtain 
a multi-scale image set. The trained U-net is fine-tuned at each level to obtain an independent 
model. Then, shift cropping and weighted overlap are used in the prediction process to solve block continuity. Finally, the predicted image is mapped back to the original 
size, and a voting mechanism is proposed to combine the multi-scale predicted images. The 
experimental data are the verification images of the 2019 MICCAI PAIP Challenge. The evaluation 
results show that this algorithm is better than other algorithms.

\subsection{Other Segmentation Methods}
In~\cite{Bergmeir-2012-SCCN}, a new algorithm for segmenting cell nuclei is used. Before 
determining the precise shape of the cell nucleus by the elastic segmentation algorithm, 
the proposed algorithm uses a voting scheme and prior knowledge to locate the cell nucleus. 
After removing noise through the mean shift and median filtering, the Canny edge detection 
algorithm is used to extract edges. Since the nucleus is observed to be surrounded by cytoplasm, 
its shape is roughly elliptical, and the edges adjacent to the background are removed. The 
random hough transform of the ellipse finds the candidate kernel, and then processes it through 
the level set algorithm. The algorithm is tested and compared with other algorithms in a database 
containing 207 images obtained from two different microscope slides, and the results passed the 
positive predictive value (PPV) and TPR. The high value of, is displayed, resulting in a high 
measurement value of $96.15\%$.

\begin{figure}[htbp!]
\centerline{\includegraphics[width=0.4\linewidth]{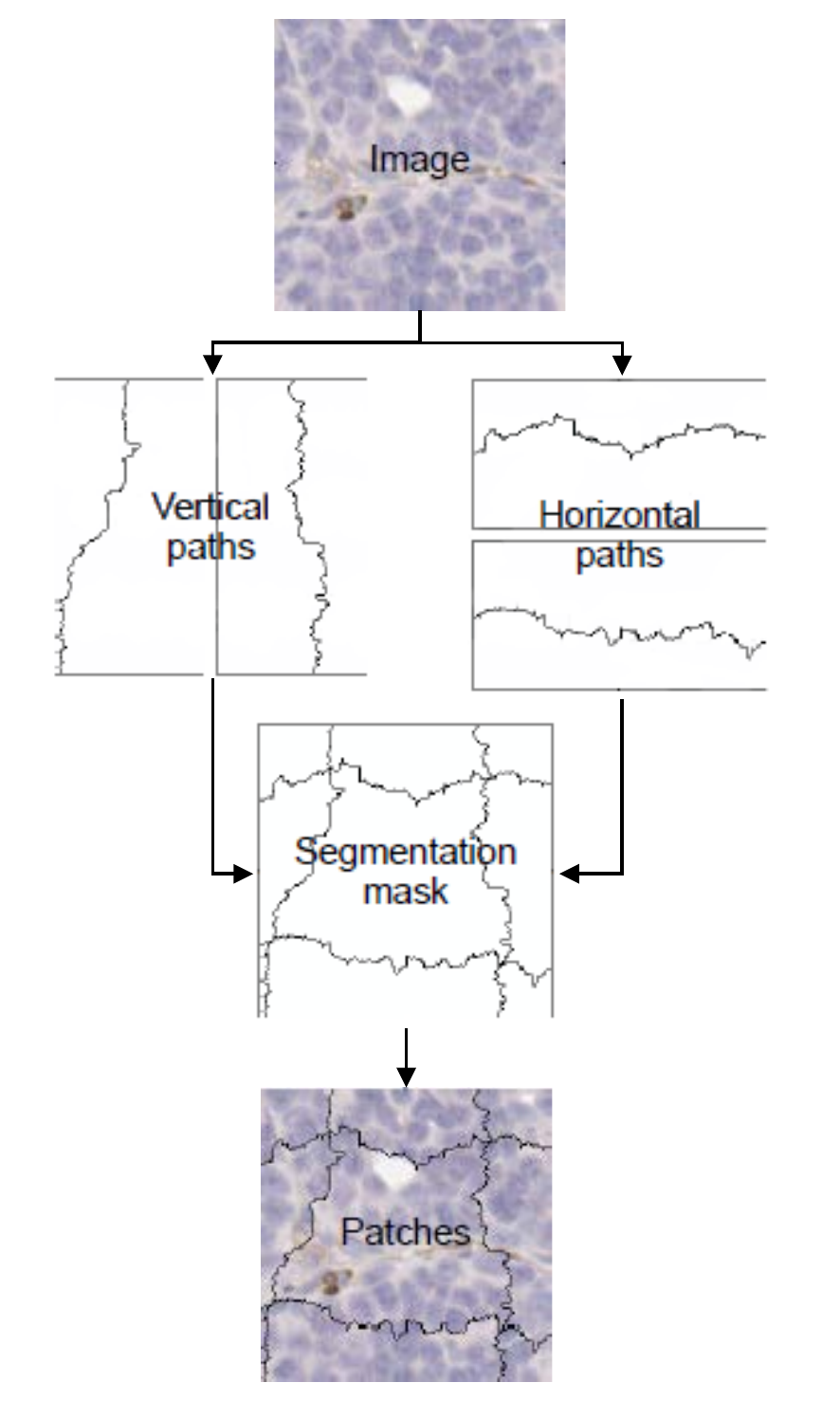}}
\caption{Path computation in horizontal and vertical strips, leading to an image partition. 
This figure corresponds to Fig.4 in~\cite{Apou-2014-FSTC}}
\label{fig:FSTC}
\end{figure}

In~\cite{Apou-2014-FSTC}, a fast segmentation method based on WSI is proposed. Due to the 
large size of the WSI, a set of horizontal and vertical optimal paths that follow the high 
gradient of the image are used to segment the image so that the relevant segmentation of the 
image is provided in an effective manner. Then other subsequent steps are executed. The schematic 
diagram of the segmentation process is shown in Figure.~\ref{fig:FSTC}.

\begin{figure}[htbp!]
\centerline{\includegraphics[width=0.98\linewidth]{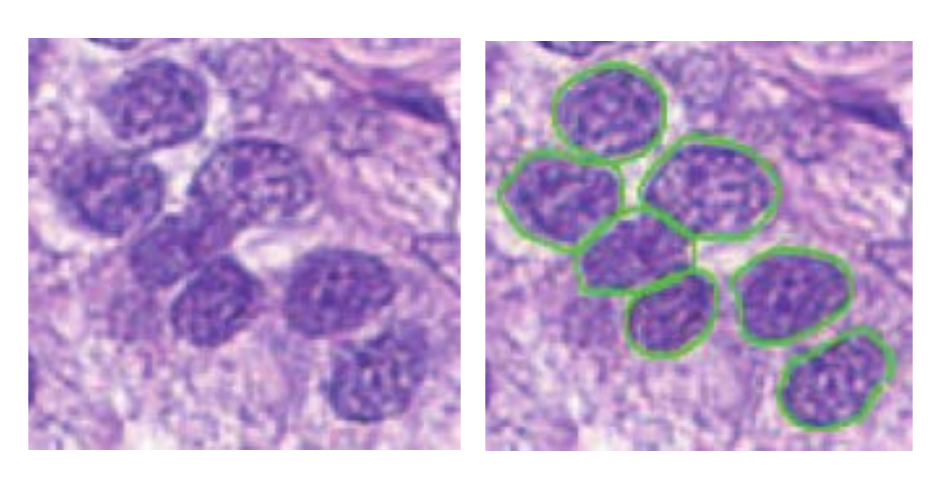}}
\caption{Segmentation results of different methods on a randomly picked patch. 
From left to right: original image and ours. This figure corresponds to Fig.3 in~\cite{Zhang-2015-FGHI}}.
\label{fig:FGHI}
\end{figure}

In~\cite{Zhang-2015-FGHI}, a robust segmentation method is developed to accurately delineate 
ROI (eg, cells) using hierarchical voting and repelling active contours. Its segmentation is 
based on active contours with repelling terms~\cite{Cohen-1991-OACM}. The exclusion term is 
used to prevent the contour lines from intersecting and merging. Based on the detection result, 
the circle is associated with each detected cell as the initial contour. The final segmentation 
result is shown in Figure.~\ref{fig:FGHI}.

In~\cite{Li-2015-FRID}, a new ROI search and segmentation algorithm based on superpixels 
is proposed. First, the initial recognition of the ROI is obtained by gathering superpixels 
at low magnification. Then, by marking the corresponding pixels, the superpixels are mapped 
to the higher magnification image. This process is repeated several times until the segmentation 
is stable. This method is different from the previous classic segmentation methods based on 
superpixels~\cite{Achanta-2012-SSCS, Yamaguchi-2014-EJSO}. This algorithm provides image 
segmentation with topology preservation.

\begin{figure}[htbp!]
\centerline{\includegraphics[width=0.98\linewidth]{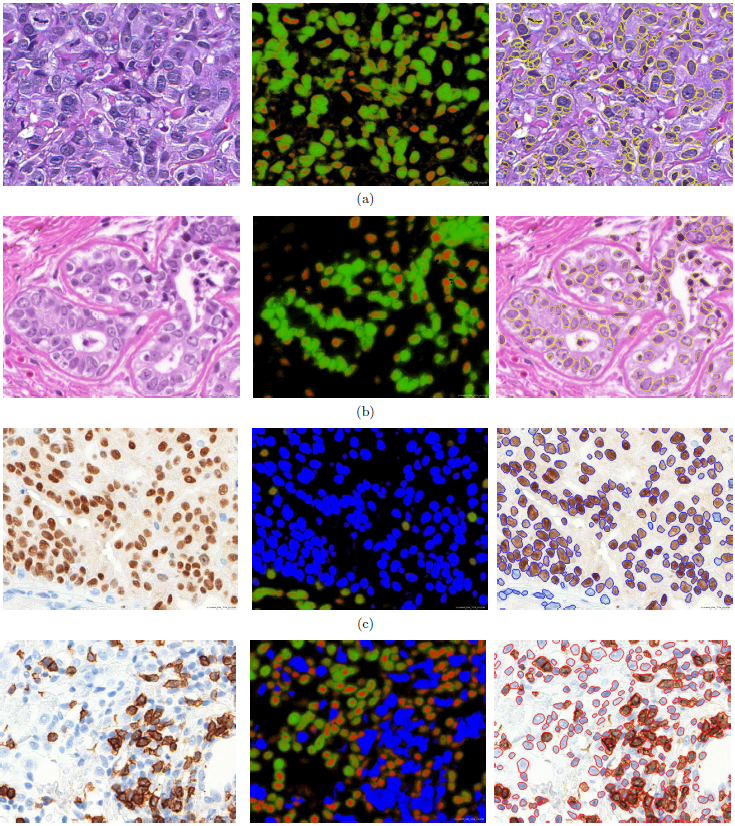}}
\caption{The schematic diagram after segmentation is shown in the figure. 
Left: H\&E (a)-(b), IHC-nuclei (c), and CD8 (d) regions; Center: posterior maps for marked 
nuclei and cells (blue channel if applicable), for homogeneous nuclei (red channel) and for 
textured nuclei (green channel); Right: outline of the segmentation result. 
This figure corresponds to Fig.3 in~\cite{Brieu-2016-SSMS}. }
\label{fig:SSMS}
\end{figure}

In~\cite{Brieu-2016-SSMS}, threshold processing is performed on the foreground posterior image 
to detect the foreground area, and spot detection algorithms such as  MSER and Gaussian difference are used further to identify the brightness of the image. According to the classification 
function combining suitable candidate 
parameters are selected. The remaining area is further divided by calculating the minimum path 
of the posterior mapping between the concave points and checking the goodness of fit of 
the candidate area in turn. The schematic diagram after segmentation is shown in 
Figure.~\ref{fig:SSMS}.

\subsection{Summary}
As can be seen from the content we reviewed above, machine learning is used in 
combination with WSI technology to assist diagnosis. In the field of image segmentation, 
used methods include thresholding-based segmentation, region-based segmentation, graph-based 
segmentation, clustering-based segmentation, deep learning-based segmentation, and other 
segmentation methods.

We can know that the graph-based segmentation method is a more classic image algorithm, so it 
became popular earlier, and many algorithms are based on the graph-based segmentation method. 
Thresholding-based segmentation methods are often used in WSI in combination with the watershed 
algorithm in region-based segmentation methods. In the papers we reviewed, three of them used 
both methods at the same time. Then clustering-based segmentation methods and some other segmentation 
methods appeared to be used. Until 2017, deep learning methods are widely used, and the application 
of deep learning to WSI segmentation began to get good results. Among them, the U-net architecture 
based on multi-resolution has been used many times. Table. ~\ref {SCMS} is a summary of the CAD 
method for segmentation technology in WSI. 
\onecolumn
\begin{center}\tiny
\renewcommand\arraystretch{1.85}
\setlength{\tabcolsep}{0.001 pt}
\newcommand{\tabincell}[2]{\begin{tabular}{@{}#1@{}}#2\end{tabular}}
\centering
\begin{longtable}{ccccc}
\caption{Summary of the CAD methods used for segmentation in WSI.} \\
\endfirsthead
\caption[l]{Continue: Summary of the CAD methods used for segmentation in WSI.}\\
\hline
\endhead
 \hline
\endfoot
\hline
Method                       & Reference                  & Year & Team                       & Details                                                                                                                                                                           \\ \hline
Thresholding                 & ~\cite{Kong-2011-IMAG}             & 2011 & Jun Kong                   & Simple threshold watershed method                                                                                                                                                          \\
Thresholding                 & ~\cite{Lu-2012-ASAE}               & 2012 &Cheng Lu                   & Otsu threshold                                                                                                                                                                             \\
Thresholding                 & ~\cite{Shu-2013-SOCN}              & 2013 & Jie Shu                    & \tabincell{c}{Automatic threshold method\\ (foreground and background \\classification), region \\growth, watershed \\segmentation (seriously\\ clustered overlapping \\cores for segmentation)}         \\
Thresholding                 & ~\cite{Vo-2016-CWSI}               & 2016 & Vo H                       & \tabincell{c}{Parallelization of \\image segmentation based\\ on MapReduce} \\
Thresholding                 & ~\cite{Arunachalam-2017-CAIS}      & 2017 & Arunachalam H B            & \tabincell{c}{The combination of pixel-based and\\ object-based methods (k-means and\\ non-tumor images, and multi-level\\ Otsu threshold segmentation)}                                          \\
Region-based(Watered)        & ~\cite{Kong-2011-IMAG}             & 2011 & Jun Kong                   & \tabincell{c}{Simple threshold \\watershed method                                                                                                                                                         } \\
Region-based(Watered)        & ~\cite{Veta-2012-PVAE}             & 2012 & Mitko Veta                 & \tabincell{c}{Watershed segmentation \\controlled by\\ multi-scale markers}                                                                                                                          \\
Region-based(Watered)        & ~\cite{Veta-2013-ANSH}             & 2013 & Veta M                     & \tabincell{c}{Marker-controlled \\watershed segmentation, \\with multiple scales and\\ different markers\\ (automatic nuclear segmentation)}                                                           \\
Region-based(Watered)        & ~\cite{Shu-2013-SOCN}              & 2013 & Jie Shu                    & \tabincell{c}{Automatic threshold method\\ (foreground and background\\ classification), region growth,\\ watershed segmentation\\ (seriously clustered overlapping \\cores for segmentation)}                  \\
Region-based(Watered)        & ~\cite{Vo-2016-CWSI}               & 2016 & Vo H                       & \tabincell{c}{Parallelization of Image\\ Segmentation Based\\ on MapReduce}                                                                                                                          \\
Region-based(Watered)        & ~\cite{Vo-2019-MCMH}               & 2019 & Hoang Vo                   & \tabincell{c}{Core segmentation\\ (MapReduce architecture)} \\
Region-based(Region growing) & ~\cite{Shu-2013-SOCN}              & 2013 & Jie Shu                             & \tabincell{c}{Automatic threshold method\\ (foreground and background\\ classification), region growth, \\watershed segmentation (seriously \\clustered overlapping cores\\ for segmentation)}         \\
Graph-based                  & ~\cite{Roullier-2010-GMRS}         & 2010 & V. Roullier                & \tabincell{c}{Integrates graph-based \\segmentation (discrete semi-\\supervised clustering)\\ and multi-resolution\\ segmentation (cluster\\ space refinement)}                                          \\
Graph-based                  & ~\cite{Roullier-2010-MEBC}         & 2010 & Vincent Roullier           & \tabincell{c}{Integrates graph-based \\segmentation (discrete semi-\\supervised clustering)\\ and multi-resolution\\ segmentation (cluster\\ space refinement)}                                          \\
Graph-based                  & ~\cite{Roullier-2011-MRGA}         & 2011 & Vincent Roullier           & \tabincell{c}{Multi-resolution segmentation\\ method (regularization framework,\\ histogram construction, \\histogram 3 mean clustering,\\ partition and spatial refinement)}                        \\
Clustering                   & ~\cite{Hiary-2013-SLWS}            & 2013 & Hazem Hiary                & $k$-means clustering                                                                                                                                                                \\
Clustering                   & ~\cite{Arunachalam-2017-CAIS}      & 2017 & Arunachalam H B            & \tabincell{c}{The combination of pixel\\-based and object-based methods \\(k-means and multi-level\\ Otsu threshold)}                                                                   \\
Deep Learning                & ~\cite{Bandi-2017-CDMT}            & 2017 & Bándi P                    & \tabincell{c}{Using FCN and U-net \\to organize background\\ segmentation} \\
Deep Learning                & ~\cite{Xu-2017-LSTH}               & 2017 & Xu Y                       & SVM-CNN                                                                                                                                                                                    \\Deep Learning                & ~\cite{Dong-2018-RANT}             & 2018 & Nanqing Dong               & \tabincell{c}{High-resolution image \\semantic segmentation\\ framework-Reinforced \\Auto-Zoom Net (RAZN) (FCN)}                                                            \\Deep Learning                & ~\cite{Cui-2018-DLAO}              & 2018 & Cui Y                      & \tabincell{c}{Supervised FCN method \\for nuclear segmentation\\ in histopathological images}                                                                                                      \\Deep Learning                & ~\cite{Sirinukunwattana-2018-IWSS} & 2018 & K Sirinukunwattana         & \tabincell{c}{Combining multiple \\CNNs of different\\ scales with LSTM}                                                                                                                             \\Deep Learning                & ~\cite{Mehta-2018-LSBB}            & 2018 & Sachin Mehta               & \tabincell{c}{Multi-resolution \\encoder-decoder network\\ semantic segmentation}                                                                                                                    \\Deep Learning                & ~\cite{Vo-2019-MCMH}               & 2019 & Hoang Vo                   & \tabincell{c}{Core segmentation \\(MapReduce architecture)                                                                                                                                       } \\Deep Learning                & ~\cite{Seth-2019-ALBD}             & 2019 & Nikhil Seth                & \tabincell{c}{Multi-resolution \\U-net architecture                                                                                                                                              } \\Deep Learning                & ~\cite{Seth-2019-ASDW}             & 2019 & Seth N, Akbar S            & \tabincell{c}{Multi-resolution \\U-net architecture                                                                                                                                              } \\Deep Learning                & ~\cite{Feng-2020-DLMA}             & 2020 & Feng Y, Hafiane A          & \tabincell{c}{Multi-resolution \\seven-layer pyramid U-net} \\Other                        & ~\cite{Bergmeir-2012-SCCN}         & 2012 & Christoph Bergmeir         & \tabincell{c}{Random Hough transform\\ ellipse fitting cell\\ nucleus to segment the nucleus\\ (level set algorithm)}                                                                                \\Other                        & ~\cite{Apou-2014-FSTC}             & 2014 & Apou G, Naegel B           & \tabincell{c}{The path calculation in \\the horizontal and \\vertical strips}                                                                                                                      \\Other                        & ~\cite{Zhang-2015-FGHI}            & 2015 & Xiaofan Zhang              & \tabincell{c}{(Robust segmentation method)\\ Use Euclidean distance. \\Based on active contours\\ with repelling terms}                                                 \\Other                        & ~\cite{Li-2015-FRID}               & 2015 & Ruoyu Li and Junzhou Huang & \tabincell{c}{The superpixels are clustered\\ at low magnification to \\obtain ROI. Then, the \\superpixels are mapped\\ to the higher magnification\\ image. This process is\\ repeated several times.} \\Other                        & ~\cite{Brieu-2016-SSMS}            & 2016 & Brieu N, Pauly O           & \tabincell{c}{It is further divided\\ by calculating the \\minimum path of the \\posterior mapping between \\the concave points and \\checking the goodness of\\ fit of the candidate\\ area in turn}    \\
\label{SCMS}\\
\end{longtable}
\end{center}

\section{Classification Methods}
\label{s:CM} 
Image classification, as the name suggests, is to have a fixed set of classification labels, 
and then for the input image, find a classification label from the classification label set, 
and finally assign the classification label to the input image. It is at the heart of computer 
vision and is the most fundamental issue that forms the basis for other computer vision tasks 
such as positioning, detection, and segmentation~\cite{Rawat-2017-DCNN}~\cite{Karpathy-2015-DVSA}. 
It is widely used in practice. While a simple task for humans, it can be challenging for computer 
systems. Many seemingly different problems in computer vision (such as object detection and 
segmentation) can be reduced to image classification.

In the analysis of pathological images, the most studied task is CAD. It also helps the pathologist 
make a diagnosis. The diagnostic process is the task of mapping one or more WSIs to a disease 
category. Since errors produced by machine learning systems are different from those produced by 
human pathologists~\cite{Wang-2016-DLIM}, the use of computer-aided design systems can improve 
classification accuracy~\cite{Komura-2018-MLMH}.

In recent years, due to the progress of computer technology, histopathological image classification 
has gradually become a research hotspot in the field of medical image processing. To the human 
anatomy area and the pathological changes area to carry on the accurate classification, may the 
maximum degree doctor accurate, the rapid diagnosis condition. This is of great significance to 
the further diagnosis of doctors and the further treatment of patients.

In the papers we summarized, there are around 54 articles from 2004 to 2020 on image classification 
using WSI techniques to assist pathologists in diagnosis. We can see from the development trend 
in Fig.~\ref{fig:trend} that the application of classification has increased. This 
reflects the wide application of classification technology. From these papers, we can briefly 
summarize the classification methods they applied including traditional machine learning algorithms, 
deep learning algorithm, and other methods.

\subsection{Traditional Machine Learning based Classification Method}
Among the papers we reviewed, there are 18 papers involve the classification of WSI by using 
traditional machine learning algorithms for CAD.

\paragraph{SVM-based Classification Method}~{}
\\

SVM is a type of supervised machine learning technique. It was first published in 1963 by 
Vladimir N. Vapnik and Alexander Y. Lerner~\cite{Vapnik-1963-RPHG}. It uses the hypothesis 
space of linear functions in hyperspace~\cite{Roweis-2000-NDRL} and trains with the learning 
algorithm of optimization theory, which realizes the learning bias derived from statistical 
learning theory. The purpose of classification by SVM is to find an effective computational 
method to learn good separation hyperplanes in hyperspace~\cite{Kumar-2012-RCIC}. SVM is 
designed for binary classification. when an SVM is applied to a multi-class classification 
problem, it internally splits the task into multiple binary classification problems and uses 
several SVMs to solve them~\cite{Kamath-2018-CSTM}~\cite{Madzarov-2009-MCSC}. A total of ten 
papers in our review involve the use of SVMs for WSI classification.

\begin{figure}[htbp!]
\centerline{\includegraphics[width=0.98\linewidth]{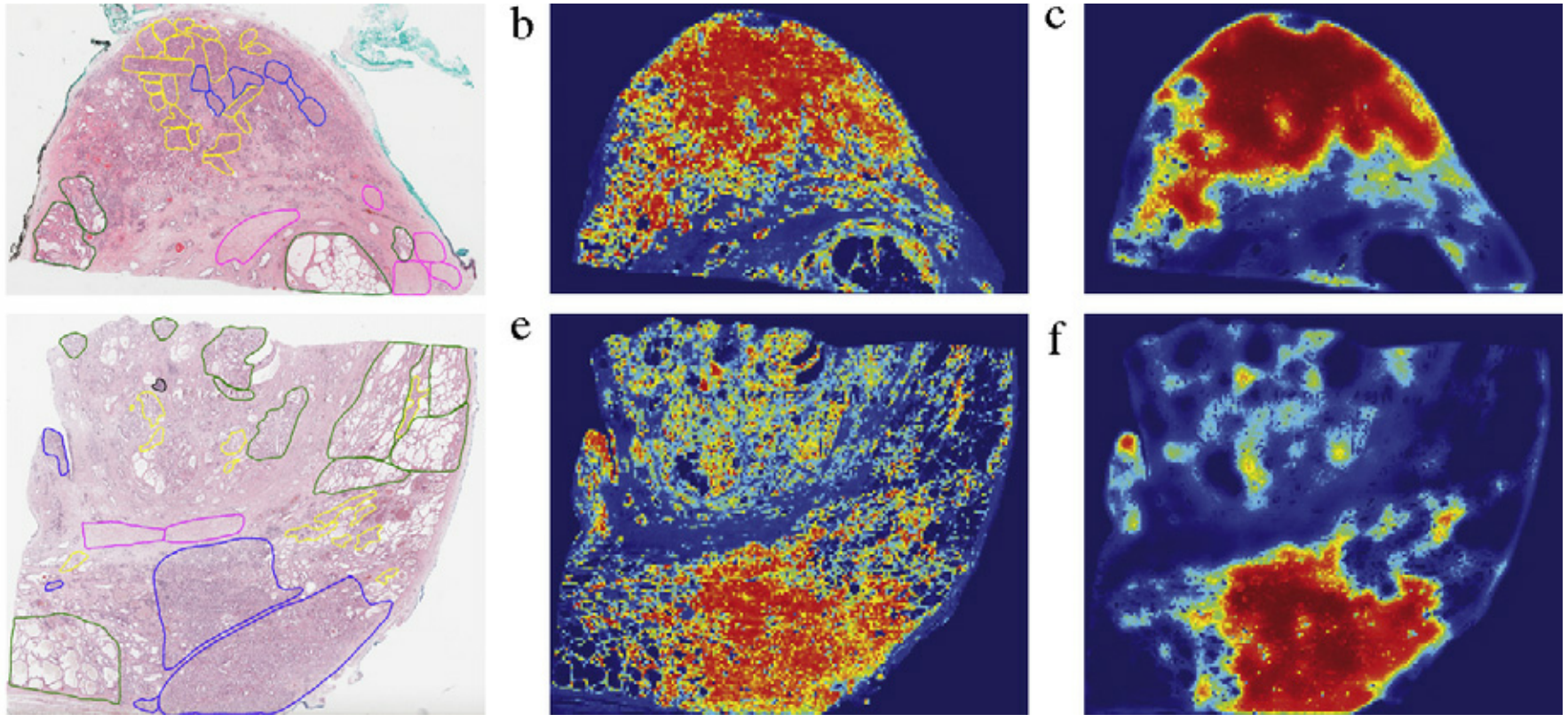}}
\caption{This figure shows the utility of using classification probabilities for heat map 
visualization at the tile level. There is a good difference between strong positive and 
negative results (red versus blue). This figure corresponds to Fig.16 in~\cite{Difranco-2011-ESWS}.}
\label{fig:ESWS}
\end{figure}

In~\cite{Difranco-2011-ESWS}, a tile-based approach is proposed to generate clinically relevant 
probability maps of prostate cancer in prostate WSIs. The probability of cancer 
existence is calculated from the response of each classifier in the ensemble. Before classification, 
texture feature extraction and spatial filtering are performed. The classification is then 
performed using either an RF or an SVM (linear and radial kernels). Different feature subsets and 
different subsampled training data strategies are used for performance comparison. The final best 
classification result is obtained by Radial Basis Function (RBF) kernel SVM, which reports an AUC value of $95.50\%$. The final heat map is shown in Figure.~\ref{fig:ESWS}.

\begin{figure}[htbp!]
\centerline{\includegraphics[width=0.98\linewidth]{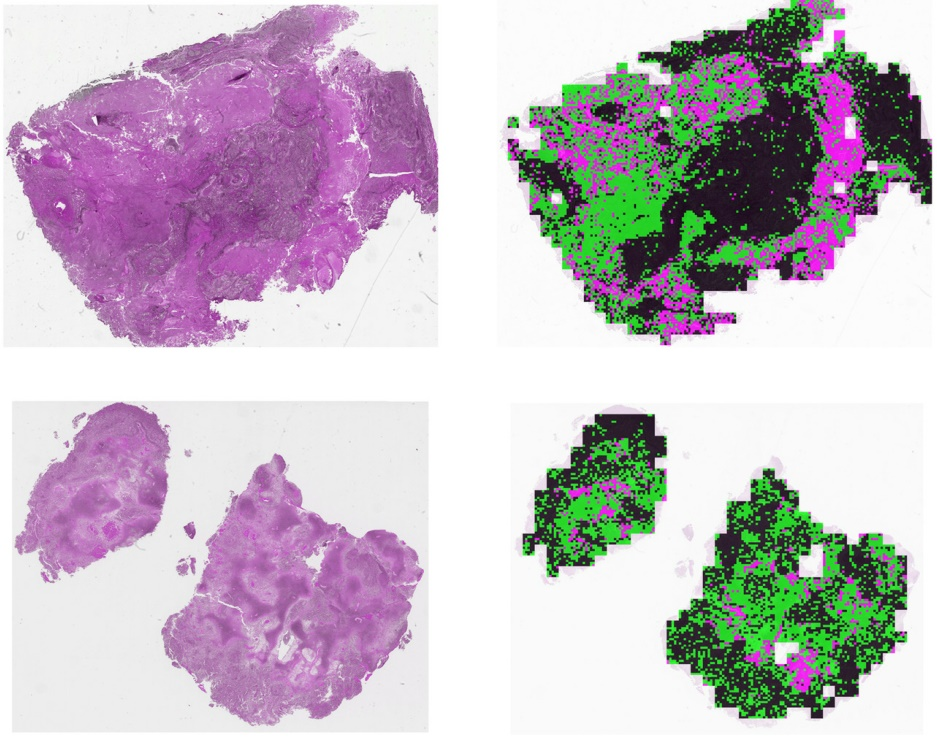}}
\caption{Two examples of classification results of a heterogeneous GBM tissue sections. 
The left and right images correspond to the original and classification results, respectively. 
Color coding is black (tumor), pink (necrosis), and green (transition to necrosis). 
This figure corresponds to Fig.4 in~\cite{Nayak-2013-CTHS}.}
\label{fig:CTHS}
\end{figure}

In~\cite{Nayak-2013-CTHS}, a method of automatically learning features from unlabeled images 
is proposed for WSI classification. The first step is to learn the dictionary from the unlabeled 
images. Then the sparse automatic encoder is used to learn the function. An automatic encoder 
consists of three parts: an encoder, a dictionary, and a set of codes. The encoder is used to 
train the classifier on a small amount of label data. Multi-class regularization support vector 
classification is used, the regularization parameter is 1, and the polynomial kernel is 3. In 
terms of data, two datasets from (i) glioblastoma multiforme (GBM) and (ii) clear cell renal 
cell carcinoma (KIRC) of TCGA are used. The classification accuracy rates of $84.3\%$ and $80.9\%$ 
are obtained respectively. The classification results of heterogeneous GBM tissue sections are 
shown in Figure.~\ref{fig:CTHS}.

In~\cite{Jiao-2013-CCDU}, an SVM based classifier is used to classify colon cancer 
WSI and non-colon cancer WSI. Then, 18 simple features (such as gray-level mean and gray-level 
variance) and 16 texture features extracted by the GLCM method are selected as the feature set. 
The final result shows that when all features are used, the mean values of accuracy, recall and 
F-measure are $96.67\%$, $83.33\%$, and $89.51\%$, respectively.

In~\cite{Lu-2015-AADS}, a technology for automatically diagnosing skin using WSI is proposed. 
There are five steps: epidermal segmentation, keratinocyte segmentation, melanocyte detection, 
feature construction and classification. Based on the segmented ROI, the spatial distribution 
and morphological features are constructed. A multi-class SVM classifier classifies these features representing skin tissue. It provides about $90\%$ accuracy in the classification of melanoma, 
and normal skin.

In~\cite{Weingant-2015-EPTC}, the WSI of prostate tumors is classified. First, adaptive 
cell density estimation is introduced, and H\&E staining normalization is applied in the 
supervised classification framework to improve the robustness of the classifier. RF feature 
selection, class-balanced training sample sub-sampling, and SVM classification are used to 
predict high- and low-grade prostate cancer on image slices. AUC measured the classification 
performance to get 0.703 for HG-PCA and 0.705 for LG-PCA. The results proved the effectiveness 
of cell density and staining standardization for prostate WSIs classification.

In~\cite{Peikari-2015-TDRR}, a texture analysis technique is proposed to simplify the processing 
of H\&E stained WSIs by classifying clinically important areas. The specific method is to randomly 
select image blocks from the entire tissue area, divide them into small blocks, and perform Gaussian 
texture filtering on them. The texture filter responses for each texture are combined and statistical 
indicators are obtained from the histogram of the response. Then, the visual word bag pipeline is 
used to combine extracted features to form a word histogram for each image block. The SVM classifier 
is trained using the computed lexical histograms to distinguish clinically relevant and unrelated 
plaques. Finally, the ROC is 0.87. It can be proved that texture features can be used to classify 
important areas in WSI.

In~\cite{Gadermayr-2016-DACC}, to improve the generalization ability of the classification model 
for kidney WSI, a domain adaptive method is proposed, in which the classifier is trained on the data 
from the source domain to present a small number of user-labeled samples from the target domain. 
Efficient linear SVM is used to avoid waiting time during interaction. In a comparison between interactive 
and non-interactive domain adaptation, it is observed that interactive domain adaptation has a 
positive effect on the classification performance.

In~\cite{Shukla-2017-CFRI}, SVM is used to predict the presence of cancer by WSI of sentinel 
lymph nodes, and an equivalent fuzzy model is developed to improve the interpretability. The SVM model 
consists of 50 support vectors, and the accuracy rate is $94.59\%$. This uses all 50 support vectors 
to create a fuzzy rule-based model (FRBS) equivalent to two IF-then rules, which categorizes each 
WSI as non-cancerous and cancerous. FRBS achieved an accuracy rate of $91.89\%$. Experiments show 
that the work of SVM can be accurately represented in a human interpretable method.

In~\cite{Xu-2018-AACM}, an automatic WSI analysis and classification technology for melanocyte tumors 
is proposed. First, the skin epidermis and dermis regions are segmented through a multi-resolution 
frame. Next, an epidermal analysis is performed, in which a set of epidermal features reflecting the 
nucleus morphology and spatial distribution are calculated. While performing epidermal analysis, dermal 
analysis is also performed, in which the dermal cell nucleus is segmented, and a set of structural 
and cytological features are calculated. Finally, by using a multi-class SVM with extracted epidermal 
and dermal features, the skin melanocyte images are classified into different categories. It is known 
from the experimental results that the classification accuracy of this technology is over $95\%$.

\paragraph{Random Forest-based Classification Method}~{}
\\

RF is a popular machine learning algorithm, often used in classification tasks in various 
fields ~\cite{Cutler-2007-RFCE, Ghimire-2010-CLCC, Gislason-2006-RFLC, Guo-2011-RALM, Chen-2005-PPPI,Ozccift-2011-RFEC, Seera-2014-HISM, Titapiccolo-2013-AIMS}. 
RF is a collection of tree structure classifiers ~\cite{Breiman-2001-RF}. Each tree relies on the 
value of a randomly selected vector distributed in the same way among all trees in the 
forest~\cite{Masetic-2016-CHFD}. Each tree in the forest will vote once, assigning each input to 
the most likely category label. This method is fast and robust to noise, and is a successful ensemble 
that can identify nonlinear patterns in data. It can easily handle numerical and categorical 
data~\cite{Titapiccolo-2013-AIMS}. One of the main advantages of RF is that even if more trees are 
added to the forest, it causes overfitting~\cite{Chaudhary-2016-IRFC}. In the papers we have 
summarized, there are five papers mainly using RF classifiers for related WSI classification.

In~\cite{Homeyer-2013-PQNH}, it is mainly to quantify the necrotic part in histological WSIs. First, 
the threshold principle is used to eliminate the background. Color and texture features are then 
extracted. After the feature classification step is performed, another post-processing step is 
performed to process the misclassified isolated tiles, that is, using the principle of context classification and spatial context information to re-evaluate all uncertainly classified slices.
The sections are then merged, and each section is identified by merging all the non-background slices 
joined by the edges. Finally, the results are evaluated. Naive Bayes classifier, $k$-nearest neighbor 
($k$NN), and RF classifier are used in the post-processing classifier. From the experimental results, 
it can be known that the RF classifier has the best results, and the HSV-based features are better 
than RGB.

In~\cite{Sharma-2016-DCNN}, the main task is to automatically classify cancer from histopathological 
stomach images. The classification performance of traditional image analysis methods and deep 
learning methods are quantitatively compared. First, the data is augmented. In traditional image 
analysis methods, the classifier used is a RF classifier.

Metastatic breast cancer is identified in~\cite{Wang-2016-DLIM}. First, the background is 
automatically detected based on the threshold. And by comparing GoogLeNet, AlexNet, VGG16 and FaceNet. 
GoogLeNet is selected to generate tumor probability heatmaps, and a RF classifier is used to classify 
metastatic WSIs and negative WSIs, and finally the AUC of 0.925 is obtained.

In~\cite{Jamaluddin-2017-TDWS}, normal sections and tumor sections from histological images of 
lymph node tissue are classified. The first step is to remove unnecessary information. The CNN model 
is then reconstructed or trained to segment the tumor area. The specific details are in 
Sec.~\ref{ss:int:DM}. After the tumor area is segmented, the features are extracted, and the RF 
classifier is used for classification. Finally the result with an AUC score of 0.94 is obtained.

In~\cite{Klimov-2019-WSIB}, the risk of recurrence of DCIS is classified. First of all, the color normalization and down sampling tasks is performed. Secondly, the texture features are extracted. Then, input these features into 
the RF classifier to predict the high and low risk of recurrence. It is convenient for the doctor 
to give the corresponding diagnosis and treatment plan. The final result shows that the 
classifier significantly predicts the 10-year risk of recurrence during training 
(accuracy = 0.87, sensitivity = 0.71, and specificity = 0.91).

\paragraph{Others Traditional Machine Learning Classification Method}~{}
\\

In addition to the two commonly used classification methods of SVM and RF, some other methods 
are also used to classify histopathological WSI, such as Bayesian 
classifier~\cite{Domingos-1997-OSBC} and $k$NN~\cite{Hastie-1996-DANN} classifier.

In~\cite{Sertel-2008-CAPN, Sertel-2009-CPNW}, a system has been developed for quantitative analysis 
of neuroblastoma WSIs, which included stromal rich and stromal poor. The developed method is based 
on the Gaussian pyramid method with multi-resolution. WSIs include non-overlapping image blocks and 
parallel processing of image blocks, which is carried out by the parallel computing module developed 
previously. Then, the texture features are extracted and the optimal subset is executed. Next, the 
feature selection of the sequential float forward selection (SFFS) method is adopted, and the confidence 
degree is calculated by $k$NN classification. If the confidence level falls below the set threshold, 
switch to a higher resolution. The experimental results show that the overall classification accuracy 
is $95\%$, and the calculation amount is reduced by $60\%$.

In~\cite{Kong-2009-CAEN}, WSI is graded for neuroblastoma biopsy. The texture features obtained from 
the segmentation components of the tissue are extracted and processed by an automatic classifier group. 
This automatic classifier group Multiple Classifiers: $k$NN, linear discriminant analysis 
(LDA) \& $k$NN, LDA \& nearest mean (NM), correlation LDA (CORRLDA) \& $k$NN, CORRLDA \& NM, LDA \& 
Bayesian and SVM with a linear kernel. The output of multiple classifiers is then selected using a 
simple two-step classifier combination mechanism consisting of voting and weighting processes. 
The automatic classifier group is trained in multi-resolution frame with different levels of 
differentiation. The trained classification system is tested on 33 WSIs. Finally, the classification 
accuracy is $87.88\%$.

In~\cite{Doyle-2010-BBMC}, prostate cancer regions in WSIs are identified. WSIs are first decomposed 
into an image pyramid with multiple resolution levels. Areas identified as cancer by a Bayesian 
classifier at a lower resolution are then identified at a higher resolution. At each resolution level, 
the AdaBoost integration method is used to select 10 image features from more than 900 first order 
statistical, second order co-occurrence and Gaborfilter feature groups. The experimental result shows 
that compared with other classifiers, the Bayesian classifier produces higher AUC and precision.

\subsection{Deep Learning based Classification Method}
In this section, the relevant contents of using deep learning algorithms to classify histopathological 
images with WSIs are summarized.

\begin{figure}[htbp!]
\centerline{\includegraphics[width=0.98\linewidth]{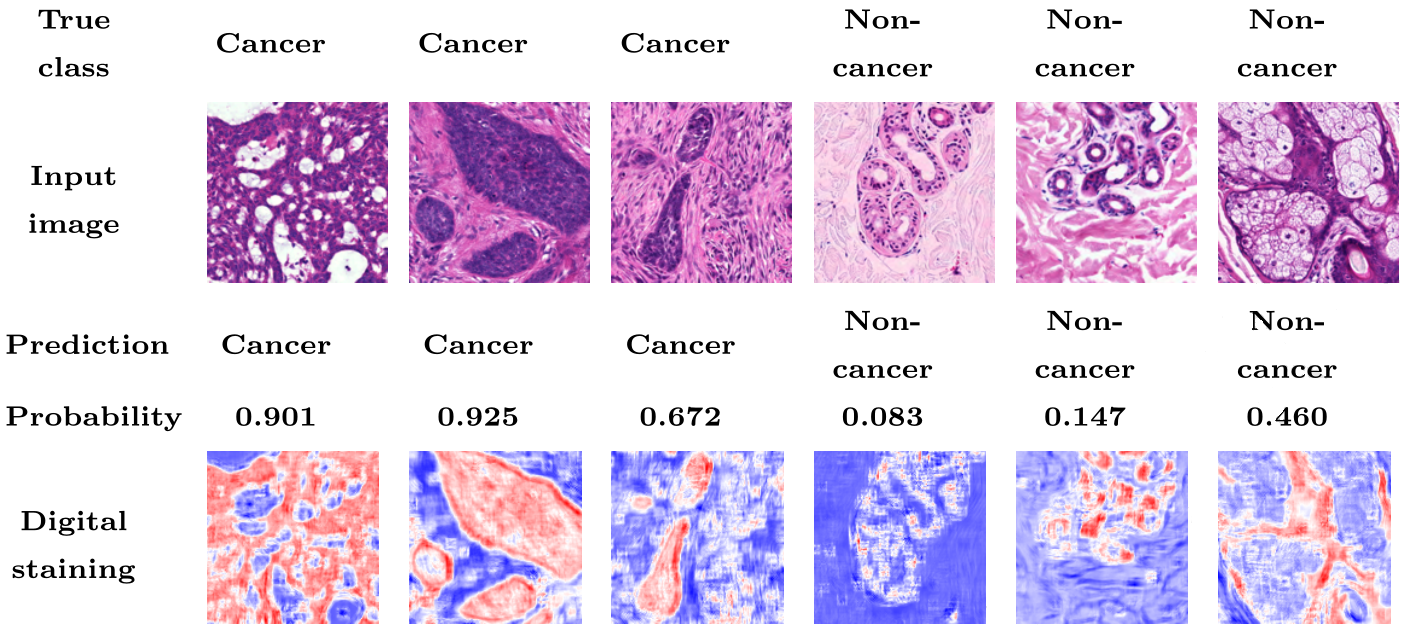}}
\caption{Outputs produced by the system for different cancer and non-cancer input images 
(red stain indicates cancer regions, blue stain indicates normal regions). 
This figure corresponds to Table.2 in~\cite{Arevalo-2015-UFLF}.}
\label{fig:UFLF}
\end{figure}

In~\cite{Arevalo-2015-UFLF}, basal cell carcinoma WSIs are studied as an integrated unsupervised 
characteristic. First, a set of feature detectors are learned from a set of patches randomly sampled 
in the image set. The detectors will capture the most common patterns by simulating the automatic 
encoder neural network. The image is then represented using the convolution or BOF method. This 
representation is achieved using the feature detector learned in the previous step. Next, the 
representation obtained from the convolution or BOF method to train the binary classification 
model, the softmax regression classifier. Basal cell carcinoma includes different categories of 
cancer and non-cancer carcinoma. The final result of the system is shown in Figure.~\ref{fig:UFLF}. 
The best results in AUC are obtained, which are superior to the most advanced $7\%$ and $98.1\%$.

In~\cite{Barker-2016-ACBT}, brain tumor types in WSIs are automatically classified. The main 
method is to analyze local features from coarse to fine in pathological images. Firstly, the 
diversity of rough areas in WSIs is analyzed, including the spatial local characteristics of 
shape, color, and texture of WSIs. Then the clustering-based method is used to create the 
representative group. The individual representative tiles in each group are then concretely 
analyzed. An Elastic Net classifier produces the diagnostic decision value for tile to obtain 
the WSI level diagnosis. Finally, 302 cases of brain cancer included automatically  two possible 
diagnoses (glioblastoma multiforme (182 cases) and glioblastoma low grade (120 cases)) to evaluate 
our method with an accuracy rate of $93.1\%$ ($P \ll 0.001$).

\begin{figure}[htbp!]
\centerline{\includegraphics[width=0.98\linewidth]{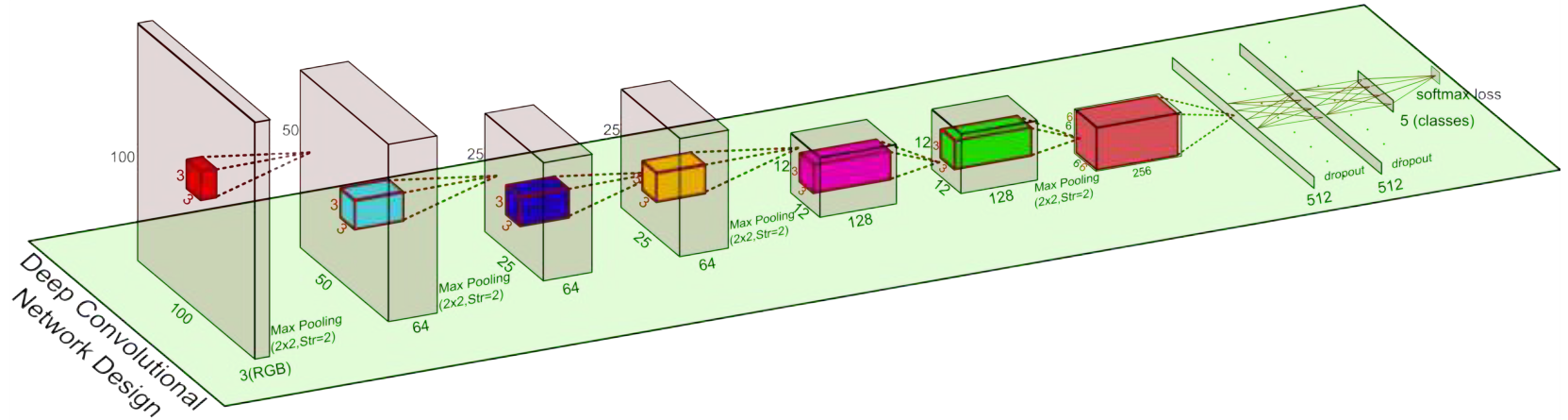}}
\caption{Designs of the CNN. 
This figure corresponds to Figure.4.6 in~\cite{Geccer-2016-DCBC}.}
\label{fig:DCBC}
\end{figure}

In~\cite{Geccer-2016-DCBC} and~\cite{Gecer-2018-DCCW}, deep convolutional networks are used 
to detect and classify WSIs ROI in breast cancer. The detection part is described in detail 
in Sec.~\ref{ss:int:DM}. In this part, the classification is mainly introduced, that is, 
the ROI include five diagnostic categories. The classifier is designed based on CNN, which 
uses the features learned by CNN to classify the detected ROI. The CNN structure diagram 
designed is shown in Figure.~\ref{fig:DCBC}. Then the post-processing of WSI classification 
is carried out. WSIs are classified according to the prediction of most categories in the 
remaining cancer areas. The results show that the efficiency is improved by about 6.6 times 
with sufficient accuracy.

\begin{figure}[htbp!]
\centerline{\includegraphics[width=0.7\linewidth]{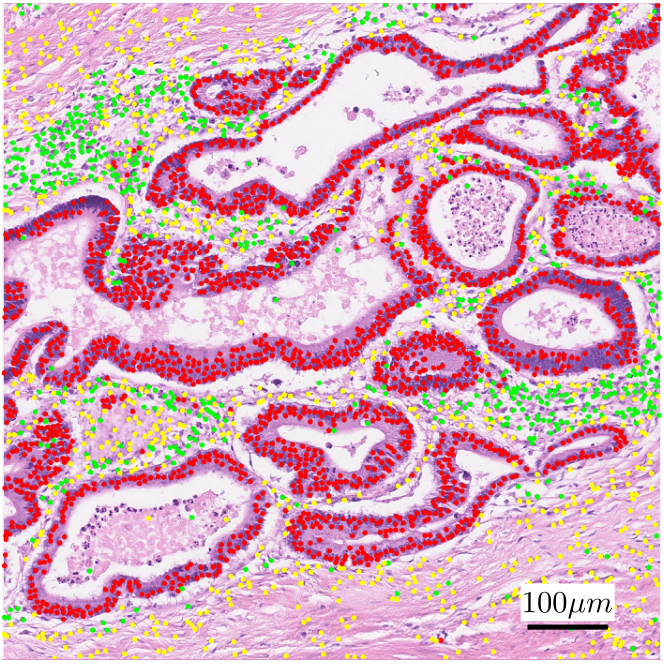}}
\caption{The detection and classification are the result graphs under 20 times magnification. 
This figure corresponds to Figure.9.(c) in~\cite{Sirinukunwattana-2016-LSDL}.}
\label{fig:LSDL}
\end{figure}

In~\cite{Sirinukunwattana-2016-LSDL}, it mainly includes two parts, namely, the detection and 
classification of cell nuclei. The detection part proposes a spatially constrained CNN for 
nuclear detection. The details are in Sec.~\ref{ss:int:DM}. Classification uses a new NEP 
combined with CNN. None of the proposed detection and classification methods require nuclear 
segmentation. And the proposed method is applied to colorectal adenocarcinoma WSIs, and the 
final classification result obtains a higher F1 score. The final detection and classification 
results are shown in the Figure.~\ref{fig:LSDL}.

In~\cite{Hou-2016-PCNN}, an expectation maximization method is proposed. The spatial relationship 
of patches is used to locate the distinguished patches robustly. This method is applied to the 
subtype classification of glioma and non-small cell lung cancer. The classification module uses 
patch-level CNN and trains a decision fusion model as a two-level model. The first level (patch-level) 
model is based on expectation maximization, combined with CNN output patch-level prediction. 
In the second level (image-level), the patch-level predicted histogram is input to the image-level 
multiple logistic regression or SVM.

In~\cite{Babaie-2017-CRDP}, a new dataset, Kimia Path24, is introduced for image classification 
and retrieval in digital pathology. The WSIs of 24 different textures are generated to test patches. 
Especially, the patch classification is based on LBP histograms, bag-of-words approach, and CNN.

In~\cite{Araujo-2017-CBCH}, a method to classify WSI for breast biopsies using CNN is proposed. 
The proposed network architecture can retrieve information on different scales. First, the image 
includes 12 non-overlapping patches, and then the piece-by-piece training CNN and CNN+SVM 
classifier are used to calculate the patch-level probability. Finally, one of three different 
patch probability fusion methods is used to obtain image classification results. These three 
methods are majority voting (choosing the most common patch as the image tag), maximum probability 
(choosing the patch category with high probability as the image tag), and probability (the 
category with the largest sum of patch-level probability). The final results classify the images 
into four categories: invasive, in situ, benign and normal. The proposed system achieves the 
overall sensitivity of about $81\%$ of cancer patch classification.

In~\cite{Bejnordi-2017-CASC}, a context-aware stacked CNN is proposed to classify WSIs as 
normal/benign, DCIS, and invasive ductal carcinoma (IDC). The first is to train a CNN and use 
high pixel resolution to capture cell level information. The characteristic responses generated 
by the model are then inputted to the second CNN and superimposed on the first CNN. The system 
had an AUC of 0.962 for binary classification of non-malignant and malignant slides, and a 
three-level accuracy rate of $81.3\%$ for normal/benign classification of WSIs, DCIS, and IDC.

\begin{figure}[htbp!]
\centerline{\includegraphics[width=0.98\linewidth]{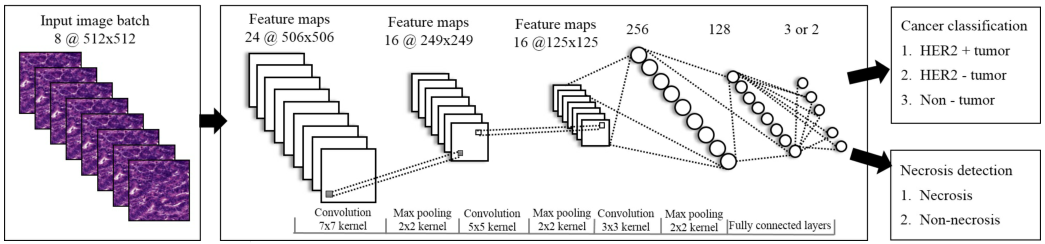}}
\caption{Proposed AlexNet architecture. 
This figure corresponds to Figure.4 in~\cite{Sharma-2017-DCNN}.}
\label{fig:DCNN}
\end{figure}

In~\cite{Sharma-2017-DCNN}, gastric cancer WSI is automatically classified. The traditional 
image analysis method and deep learning method are proposed and compared quantitatively. 
In the traditional analysis method, GLCM, Gabor filter bank response, LBP histogram, gray 
level histogram, HSV histogram, and RGB histogram are used for classification and RF. In terms 
of the deep learning method, AlexNet is proposed as a deep convolution framework. The structure 
of the network is shown in Figure.~\ref{fig:DCNN}. According to the experiment, the overall 
classification accuracy of the cancer classification proposed by AlexNet is 0.6990, and the 
overall classification accuracy of necrosis detection is 0.8144. 

In~\cite{Korbar-2017-DLCC}, different types of colorectal polyps on WSIs are classified to 
help pathologists diagnose them. Here is a modified version of the ResNet structure. The whole 
WSI is divided into patches and then applied to the ResNet. If at least five patches on a WSI 
are recognized as this class, the average confidence level is $70\%$. If there is no cancer type 
in the patch, the WSI is considered normal. Finally, the accuracy is $93.0\%$. The recall is 
$88.3\%$; F1 scores $88.8\%$.

\begin{figure}[htbp!]
\centerline{\includegraphics[width=0.98\linewidth]{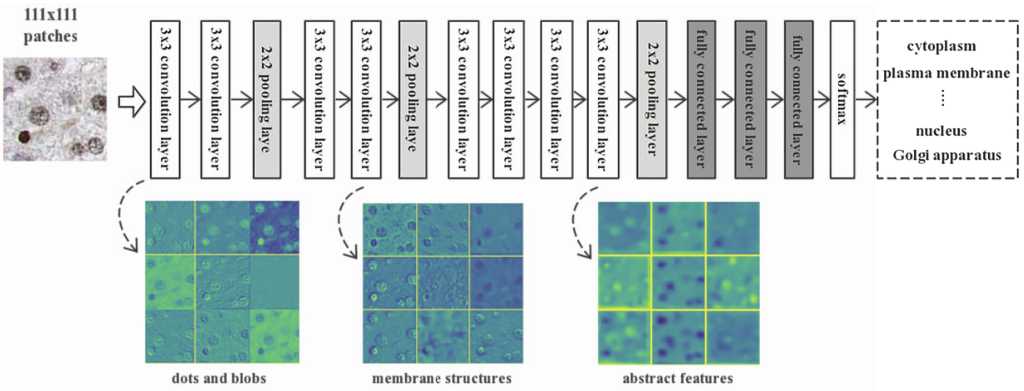}}
\caption{The deep neural network architecture and visualization of intermediate layers features. 
This figure corresponds to Figure.4 in~\cite{Hu-2017-DLBC}.}
\label{fig:DLBC}
\end{figure}

In~\cite{Hu-2017-DLBC}, a deep neural network (an 11-layer CNN model) is trained to automatically 
learn valid features and classify protein images of eight subcellular locations. First, image 
preprocessing and data balance are carried out. Then, the processed data is passed through the 
11-layers of CNN model. Details of the structure are shown in Figure.~\ref{fig:DLBC}. Finally, the 
classification accuracy rate reached $47.31\%$ in the test data and $100\%$ in the training data.

In~\cite{Xu-2017-LSTH}, a deep CNN is proposed to conduct large-scale histopathological image 
classification, segmentation, and visualization. In the part of the classification, WSI is first 
divided into patches, and background is discarded, and then the selected patches are input into 
the network to obtain 4096-dimensional CNN feature vectors. The final feature vectors of the 
image are assembled through softmax. Then feature selection is carried out to remove redundant 
and irrelevant features. Finally, SVM is used for classification. When using the MICCAI challenge 
dataset, the classification accuracy is $97.5\%$.

In~\cite{Korbar-2017-LUHD}, an image analysis method based on deep learning is proposed to 
classify different polyp types on WSIs. This classification model is based on the ResNet 
architecture, with minor modifications to the architecture. Specifically, replace the last 
fully connected layer with a convolutional layer. Finally, the average accuracy rate is $91.3\%$, 
which is better than other deep learning architectures.

\begin{figure}[htbp!]
\centerline{\includegraphics[width=0.98\linewidth]{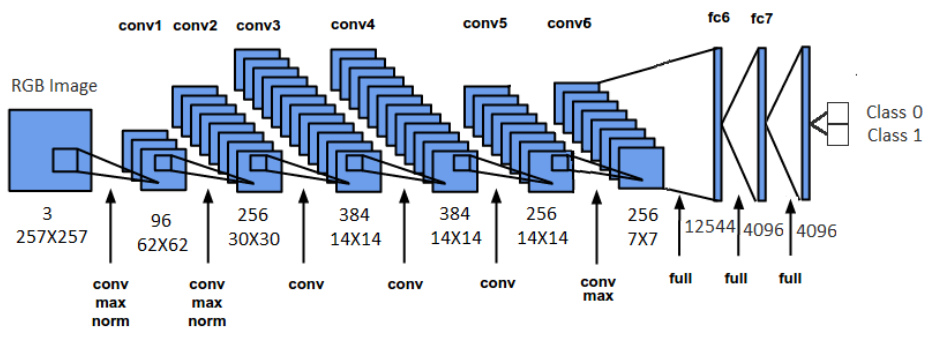}}
\caption{Architecture of the DCN used in approach. 
This figure corresponds to Figure.1 in~\cite{Ghosh-2017-SLCA}.}
\label{fig:SLCA}
\end{figure}

In~\cite{Ghosh-2017-SLCA}, a new deep learning method is proposed to classify white blood cells 
in WSIs. The network uses the average pool level to find the hot spots of the white blood cell 
location in WSI. The first is to fine-tune the pre-trained AlexNet based on the dataset. Then the 
network is trained on the patch dataset, and the trained network structure is shown in 
Figure.~\ref{fig:SLCA}

In~\cite{Das-2017-CHWS}, a network structure based on CNN is proposed. The tissue section area 
of WSI is analyzed with multiple resolution methods. That is, the class posterior estimate of 
each view at a specific magnification is obtained from the CNN at a specific magnification, and 
then the posterior-estimate of random multiple views at a multiple magnification is voted to 
filter to provide a slide-level diagnosis. According to the experimental results, the final 
classification accuracy is $94.67 \pm 14.60\%$, sensitivity $96.00 \pm 8.94\%$, specificity 
$92.00 \pm 17.85\%$, and F-score of $96.24 \pm 5.29\%$.

\begin{figure}[htbp!]
\centerline{\includegraphics[width=0.98\linewidth]{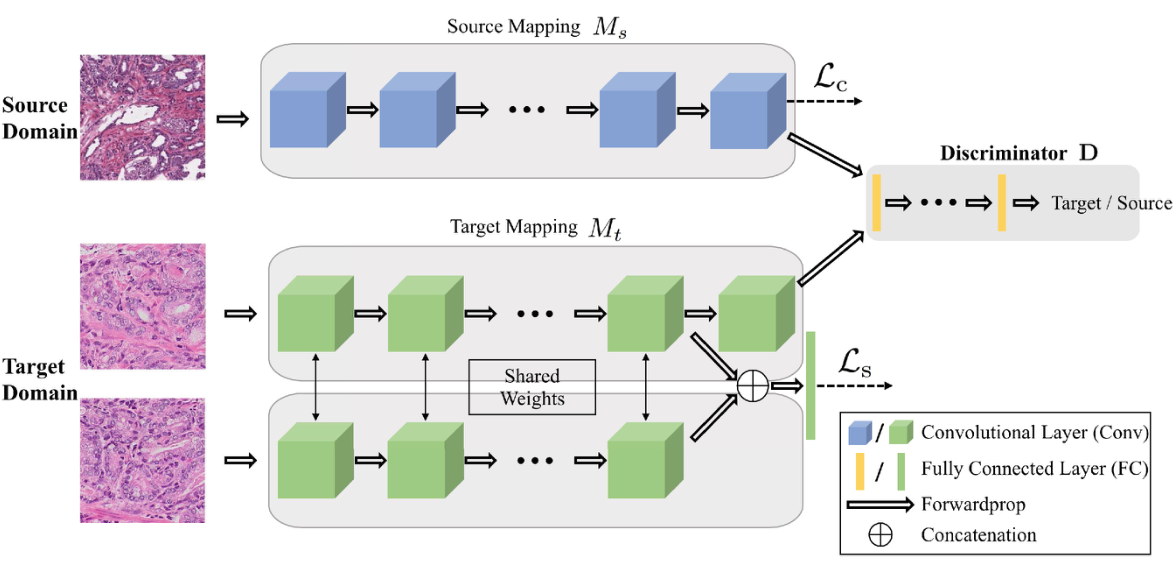}}
\caption{The architecture of the networks for the unsupervised domain adaptation. 
This figure corresponds to Figure.1 in~\cite{Ren-2018-ADAC}.}
\label{fig:ADAC}
\end{figure}

In~\cite{Ren-2018-ADAC}, prostate histopathology WSI is graded. The main method is an unsupervised 
domain adaptive method. The adaptation here is achieved through confrontation training, which can 
minimize the distribution difference of feature space between two domains (annotated source domain 
and untagged target domain) with the same number of high-level classes. The loss function also uses 
Generative Adversarial Network (GAN). Besides, a Siamese architecture is also developed to normalize 
patch in WSIs. A flowchart of the entire method is shown in  the Figure.~\ref{fig:ADAC}. The method 
is then applied to the public prostate dataset for verification. The experimental results show that 
this method improves the classification accuracy of the Gleason score significantly.

\begin{figure}[htbp!]
\centerline{\includegraphics[width=0.98\linewidth]{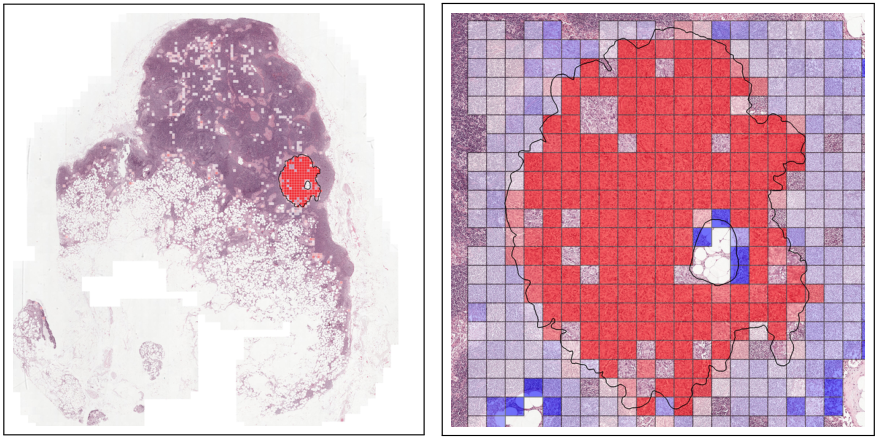}}
\caption{The classification result of the proposed method in Camelyon16 dataset. 
This figure corresponds to Figure.4 in~\cite{Courtiol-2018-CDLH}.}
\label{fig:CDLH}
\end{figure}

In~\cite{Courtiol-2018-CDLH}, a method of classification and localization of diseases with weak 
supervision is proposed. First, WSIs pretreatments are performed, including foreground and background 
segmentation, color normalization, and tile partitioning. Then ResNet is used for feature extraction. 
Then, the WELDON method is proposed by Durand et al. is improved and adjusted~\cite{Durand-2016-WWSL}. 
Modification of the pre-trained deep CNN model, feature embedding, and introduction of an additional 
set of full connection layers for context re-classification from instances. The final classification 
output diagram is generated as shown in Figure.~\ref{fig:CDLH}.

In~\cite{Tellez-2018-GWSI}, a two-part approach is proposed to classify WSIs. Firstly, the encoder 
is trained in an unsupervised way, and the tissue blocks on WSI are mapped to the embedded vector, 
and the sliding window is used to form the stack of the feature map of WSI. Then, the CNN classifier 
is trained based on the compact representation of WSIs. There are three types of encoders trained 
here: convolutional autoencoder (CAE), variational autoencoder (VAE) and a new method based on 
contrast training. Experiments show that the new contrast encoder is superior to CAE and VAE.

In~\cite{Morkunas-2018-MLBC}, WSI colorectal cancer tumor tissue is classified. The simple linear 
iterative clustering (SLIC) algorithm is proposed by Achanta et al.~\cite{Achanta-2012-SSCS} is mainly 
applied to generate superpixels, and superpixels are used to annotate the image. Therefore, the 
selected area is divided into superpixels. Then, color texture features are extracted and dimensionality 
reduction is used to regenerate composite features. Experiment shows that the superpixel method is 
suitable for using different classification algorithms based on machine learning.

\begin{figure}[htbp!]
\centerline{\includegraphics[width=0.98\linewidth]{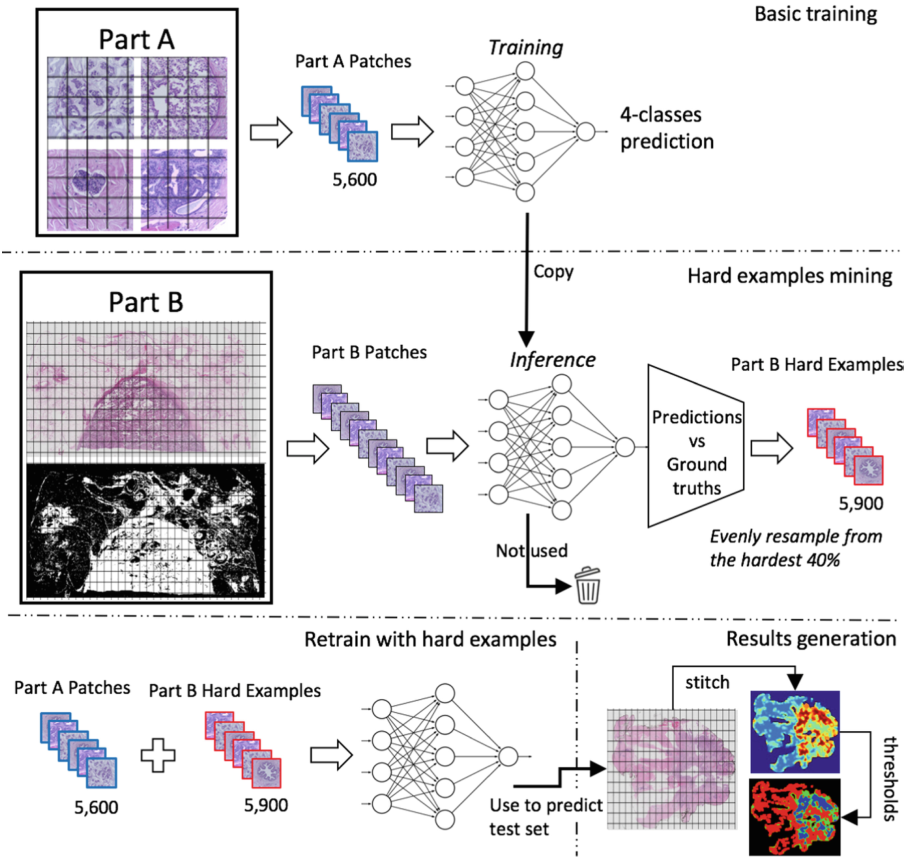}}
\caption{Overview of the framework. 
This figure corresponds to Figure.1 in~\cite{Kwok-2018-MCBC}.}
\label{fig:MCBC}
\end{figure}

In~\cite{Kwok-2018-MCBC}, a multi classification of breast cancer in WSI is presented. They propose a deep learning framework, which is mainly divided into two stages, using microscopic 
images and WSI to achieve the purpose of classification. After the two types of images are patched, 
the microscopic images are used for Inception- ResNet-v2 to train the classifier. The WSIs are then 
subsampled and converted from RGB to CIE-LAB color space and segments the foreground from and background. 
Then the extraction of hard examples and patch classifier is retrained. The prediction results are 
aggregated from the block by block prediction back to image prediction and WSI annotation. This method 
is applied to ICIAR 2018 Grand Challenge on breast cancer histology images, with an accuracy rate of 
$87\%$, far exceeding the second place. The specific working process is shown in Figure.~\ref{fig:MCBC}.

In~\cite{Mercan-2017-MIML}, a weakly supervised method for multi-classification of breast 
histopathology WSIs is proposed. Firstly, ROI is extracted by zooming, translation, and fixation. 
Then, the color and texture information and structural features are extracted in CIE-LAB space. 
Next, four different multi-instance multi-label (MIML) learning algorithms are used to predict 
the slide level and ROI level in the image. The result is an average classification accuracy 
of $78\%$ at slide level 5-Class.

The author in ~\cite{Das-2018-MILD} also proposes a multiple instance learning (MIL) framework based on breast 
cancer WSI. This framework is based on CNN and introduces a new pooling layer that enables the 
patch in WSI to aggregate the most informative features. This pooling layer is a new layer of the 
multi-instance pool (MIP), which introduces MIL as an end-to-end learning process into DNN to 
realize WSIs classification. At the end, high classification sensitivity of $93.87\%$, $95.81\%$, 
$93.17\%$, and $88.95\%$ are achieved with four different magnifications using the public dataset.

The study in ~\cite{Campanella-2018-TSDM} is also based on the MIL classification for prostate cancer. Slide 
tiling is the first run at different magnification and its bags are generated. Then, model training 
is carried out to find the tiling with the highest positive and negative probability in the slide, 
which is used to give more importance to the under-representative examples. Then, based on AlexNet, 
ResNet, VGG classification, given a threshold, if at least one instance is positive, then the WSI 
is called positive and the slide is negative if all instances are negative. The optimal models are Resnet34 
and VGG11-BN, with AUC of 0.976 and 0.977, respectively. The method of~\cite{Campanella-2019-CGCP} 
is roughly the same as that of~\cite{Campanella-2018-TSDM}.

In~\cite{Wang-2018-WSLW}, a weakly supervised learning approach is used to classify WSI lung cancer. 
Firstly, the improved FCN based on patch-level from WSI is used for cancer prediction to find the 
different regions as the prediction model of patch-level. When the probability of a block exceeds 
the threshold, it is retrieved. Then, context-based feature selection and aggregation from the retrieved 
parts are performed to construct the global feature descriptor. Finally, the global feature descriptor 
is input into the standard RF classifier. Finally, a high classification accuracy of $97.1\%$ is obtained.

In~\cite{Shou-2018-WSIC}, CNN is mainly used to classify gastric cancer in WSI. Firstly, the difference 
of threshold value and color feature is used to extract the tissue and conduct morphological processing. 
Then, it is separated into patches, and the data is expanded by flipping. Then, the existing CNN 
architecture is utilized to conduct experiments on patch-level and slide-level. Finally, good results 
are obtained on DenseNet-201.

In~\cite{Li-2019-AMRM}, a multi-resolution classification of prostate WSIs based on attention mechanism 
is performed. The MIL model based on attention is used to extract transient features. The overall 
process is separated into two parts. The first part is to classify cancer and non-cancer, using 
attention-based clustering tile selection. The second part is to conduct cancer classification research 
with higher resolution. Finally, an average classification accuracy rate of $85.11\%$ is obtained, and 
the new performance of prostate cancer classification is realized.

\begin{figure}[htbp!]
\centerline{\includegraphics[width=0.98\linewidth]{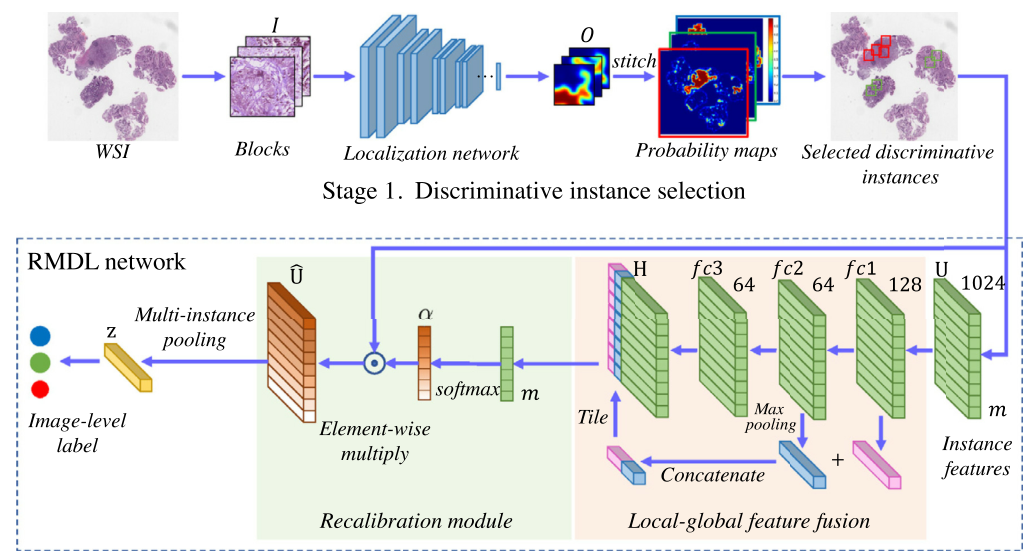}}
\caption{Overview of the framework. 
This figure corresponds to Figure.2 in~\cite{Wang-2019-RRMI}.}
\label{fig:RRMI}
\end{figure}

In~\cite{Wang-2019-RRMI}, multi-instance deep learning is also used to classify WSI gastric cancer images. 
A method named RMDL method is proposed. Similar to the above MIL method, it separated into two stages. 
The first phase also trains a localization network to select the discriminative instances. In the second 
stage, the RMDL network is used for image-level label prediction. The network is composed of local-global 
feature fusion, instance recalibration, and multi-instance pooling modules. The specific working process 
is shown in Figure.~\ref{fig:RRMI}.

\subsection{Others Classification Method}
Among the papers we have reviewed, seven papers are based on classification~\cite{Diamond-2004-UMCT, Harder-2016-CFCG, Petushi-2006-LSCH, Sertel-2009-CCCS, Yeh-2014-MSDP, Yoshida-2017-AHCWC, Yoshida-2018-AHCWG}, 
and they involve techniques related to machine learning.

In~\cite{Diamond-2004-UMCT}, a machine vision system is developed, which uses its morphological 
features and texture analysis to classify the sub-regions in WSIs. However, it focuses on the 
application of features and does not introduce the classification part of the machine vision 
system in detail. The study in ~\cite{Harder-2016-CFCG} also do not make a special description of the classifier, 
but only focused on the features.

In~\cite{Petushi-2006-LSCH}, the LNKNET software package is used to classify breast cancer WSIs. 
LNKNet integrates neural networks, statistical and machine learning 
classification, clustering, and feature selection algorithms into a modular package.

In~\cite{Sertel-2009-CCCS}, a classification framework is proposed to classify the degree of 
neuroblastoma differentiation. The main point here is that a new method of structuring structural 
features are introduced, and its classification method is also based on probability, namely 
mapping decision rules.

In~\cite{Yeh-2014-MSDP}, the pixel-based stain classification is carried out to classify the 
IHC stain and the nearest neighbor classification, mainly using the nearest neighbor and 
morphological methods. Then, the density distribution of the identified IHC stains are calculated 
by the kernel density estimator. After that the staining distribution on WSI is obtained.

Both~\cite{Yoshida-2017-AHCWC} and ~\cite{Yoshida-2018-AHCWG} are used to evaluate the 
classification accuracy of an automated image analysis system, E-pathologist. Among them, 
\cite{Yoshida-2017-AHCWC} is for colorectal biopsy, and~\cite{Yoshida-2018-AHCWG} is for 
gastric biopsy.

\subsection{Summary}
From what we have summarized above, we can see that the combination of CAD and WSI technology 
is used for classification methods, including traditional machine learning methods, deep learning 
methods, and some other methods.

In traditional machine learning algorithms, the commonly used algorithms used in WSI 
classification are SVM and RF. There are other algorithms, such as $k$NN. Since 2008, traditional 
machine learning algorithms have taken the mainstream position in WSI classification. In 2015, deep learning algorithms began to be widely used, obtaining classification results with higher 
accuracy than traditional machine learning algorithms. Among them, MIL has many applications. 
Table.~\ref {SCMC} is a summary of the CAD method of classification technology in WSI.
\onecolumn
\begin{center}\tiny
\renewcommand\arraystretch{1.85}
\setlength{\tabcolsep}{0.001 pt}
\newcommand{\tabincell}[2]{\begin{tabular}{@{}#1@{}}#2\end{tabular}}
\centering
\begin{longtable}{ccccc}
\caption{Summary of the CAD methods used for classification in WSI.} \\
\endfirsthead
\caption[l]{Continue: Summary of the CAD methods used for classification in WSI(Deep learning (DL)).}\\
\hline
\endhead
 \hline
\endfoot
\hline
Method                         & Reference                  & Year & Team                                                                         & Details                                                                                                                                                                                                                                                            \\ \hline
SVM                            & ~\cite{Difranco-2011-ESWS}         & 2011 & Matthew D. DiFranco                                                          & \tabincell{c}{RF, linear SVMs, RBF SVMs are used to classify\\ a collection of 19 B * channel features                                                                                                                                                                          } \\
SVM                            & ~\cite{Nayak-2013-CTHS}            & 2013 & Nandita Nayak                                                                & \tabincell{c}{Multi-class regularized SVM classification\\ with regularized parameters 1 and 3 polynomial kernels                                                                                                                                                               } \\
SVM                            & ~\cite{Jiao-2013-CCDU}             & 2013 & Liping Jiao                                                                  & SVM                                                                                                                                                                                                                                                                \\
\end{longtable}
\end{center}

\section{Detection Methods}
\label{ss:int:DM}
To completely understand an image, one should accurately estimate the objects' concept and location in each image~\cite{Zhao-2019-ODDL}. Detection determines whether one or more specific category instances exist in an image or not~\cite{Vedaldi-2009-MKOD}. Detection can provide valuable information for various fields. For example, remote sensing 
image detection~\cite{Zhang-2013-HRSI} can provide useful information related to geology, 
meteorology, water conservancy, and face recognition~\cite{Hsu-2002FDCI}. It is widely used in 
military and public security criminal investigations, and biomedical image 
detection~\cite{Szczypinski-2017MFBI} make it convenient for doctors in clinical diagnosis and 
pathological research.

\begin{figure}[htbp!]
\centerline{\includegraphics[width=0.65\linewidth]{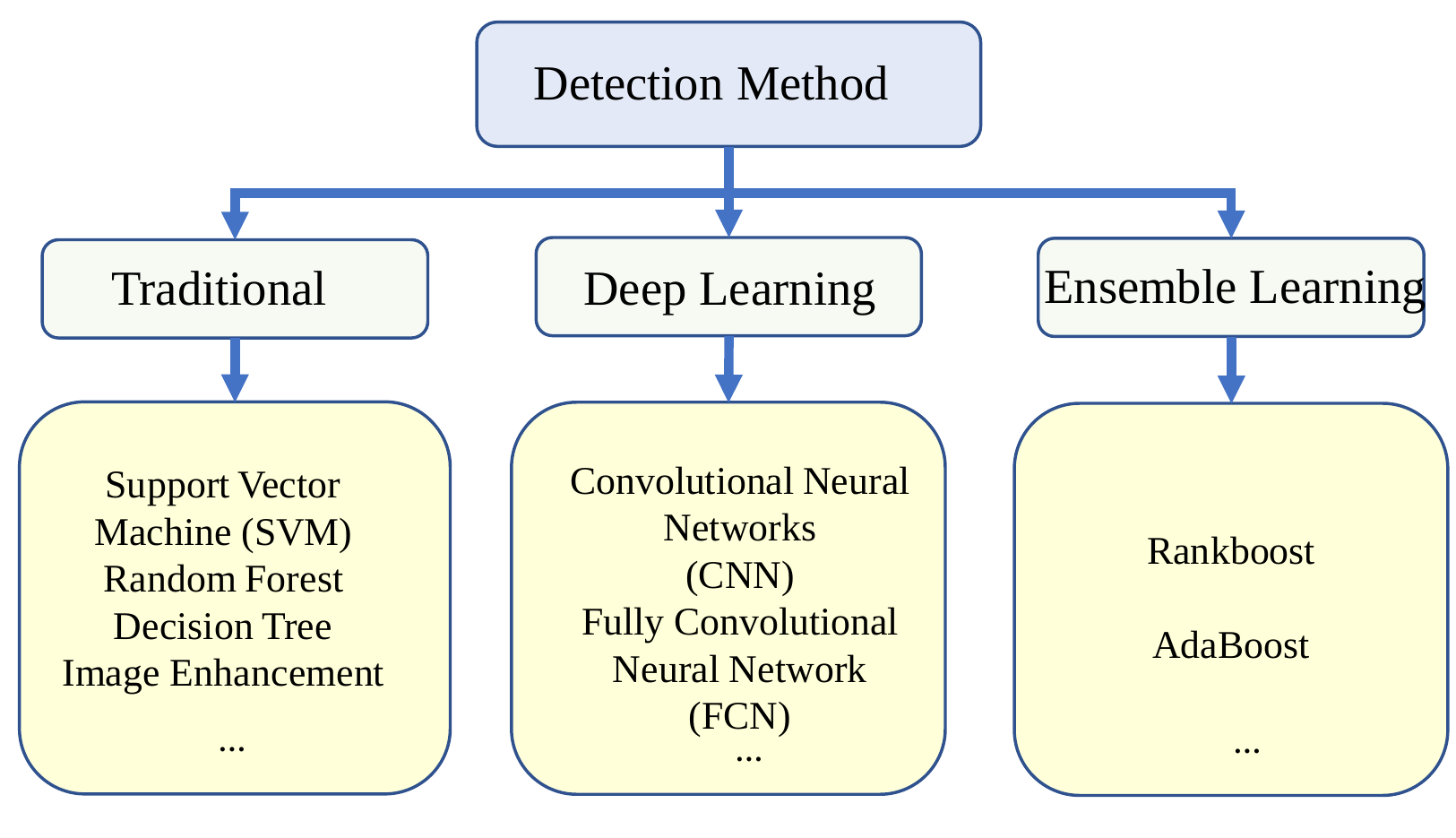}}
\caption{The structure of diseases detection methods in CAD.}
\label{fig:fig8}
\end{figure}

As one of the most common tasks for CAD pathologists to view WSI, the detection method has 
developed rapidly in recent years. Figure.~\ref{fig:trend}, the number of detection cases is increasing from 2009 to 2019, which reflects the development of detection technology. Besides, the basic content of the CAD view WSI detection 
method is shown in Figure.~\ref{fig:fig8}. As shown in Figure.~\ref{fig:fig8}, all CAD WSI methods for disease detection can be roughly divided into three categories. The 
first category is traditional detection methods (such as SVM, image enhancement, etc.), the 
second category is an ensemble learning method, and the last category is deep learning method.

\subsection{Traditional Detection Method}
In this part, we select the traditional detection methods that appear in the papers 
and summarize these methods.

\subsubsection{SVMs based Detection Methods}
In~\cite{Jiao-2013-CCDU}, colon cancer is detected by WSI, use SVM to develop a classifier, 
and select 18 simple features (such as gray-level variance and gray-level mean) and 16 texture 
features (such as GLCM) to form features. Finally, the experiment uses 3-fold cross-validation 
and achieves an average precision of $96.67\%$, recall of $83.33\%$, and F-measure of $89.51\%$.

In~\cite{Sharma-2015-ABND}, the texture feature is extracted to represent the necrotic area 
in the tissue, and the complex dataset is subjected to differentiated threshold processing 
and SVM is used for machine learning. In the end, the average cross-validation rate is 
$85.31\%$. It detects the heterogeneity of necrosis in WSI effectively.

In~\cite{Han-2018-ACDL}, calculate the tissue composition map of the WSI and extract the 
first-order and second-order texture features. Using SVM for machine learning, the offline area 
of the work area is 0.95.

In~\cite{Simon-2018-MRLF}, detect glomeruli on WSI of thin kidney tissue biopsy, LBP feature 
vector adaptation is used to train the SVM model, the experimental results have obtained high 
precision ($\textgreater 90\%$) and reasonable recall rate ($\textgreater 70\%$).

In~\cite{Swiderska-2015-TMMH}, WSI is used to determine the sample map and calculate the texture 
features of each sample pixel and its surroundings, and use SVM classifier to learn, divide 
the pixels into normal areas and bleeding areas, and then detect in WSI by applying threshold 
method and extended area minimum method in the area of tumor proliferation, the experimental 
result shows that automatic evaluation is better than manual evaluation.

In~\cite{Nguyen-2011-PCDF}, since there is no previous information about cancer, a fusion of 
cell and structural features are used, and a patch-based method is used to detect prostate cancer 
using an SVM classifier with a RBF and a parameter c of 1. When the threshold is 90 pixels, it 
provides the most satisfactory detection results (TPR = $78\%$, FPR = $6\%$).

\subsubsection{RF based Detection Methods}
In~\cite{Valkonen-2017-MDWS}, the WSI rough segmentation simplifies the background, and the 
color is normalized, the adaptive threshold method is used to obtain the 
binary image. Moreover, the watershed method is used to divide the nucleus. Then, LBP features are extracted. 
After, different models are used for metastasis detection. According to the experimental results, 
the RF achieves an AUC value of 0.97.

In~\cite{Litjens-2015-ADPC}, the SLIC algorithm is used to calculate superpixels, and after removing 
the pixel background, the color and LBP features of the remaining tissue pixels are calculated. 
Then, the RF classifier is used to detect the superpixels that are most likely to be cancer, and use 
the RF to input the highest resolution graphics and gland features to detect the area  
that may have cancer. The experimental result The experimental result obtaines an AUC of 0.96, and reaches 0.4 specificity at 1.0 sensitivity.

In~\cite{Bejnordi-2015-MSSC}, Similar to ~\cite{Litjens-2015-ADPC}, both use the multi-scale 
superpixel classification method. The experimental result of detecting breast cancer is that the 
ROC is 0.958, and the AUC for the tile analysis in comparison is 0.932. In ~\cite{Bejnordi-2016-ADDW}, 
similar to~\cite{Litjens-2015-ADPC} and~\cite{Bejnordi-2015-MSSC}, it detects DCIS in the WSIs of 
breast histopathology.

In~\cite{Li-2015-FRID}, the superpixel classification method is also used to obtain the initial 
recognition of the ROI by gathering superpixels at low magnification. Then, by marking the 
corresponding pixels, the superpixels are mapped to the higher magnification image. This process 
is repeated several times until the segmentation is stable. Finally, the RF classifier and SVM 
respectively mark the superpixels represented by the selected features and quickly detect the ROI 
in the WSI. Experiments prove that the superpixel results suggested in the article are better than 
SLIC and non-superpixel.

\subsubsection{Other Traditional Detection Methods}
There is an article each using DT classifier, clustering algorithm, linear discriminator and machine learning organization classifier~\cite{Lopez-2013-ABDM,Lopez-2012-CMAD,Veta-2013-DMFB,Shirinifard-2016-DPAU}.

The author in ~\cite{Lopez-2013-ABDM} uses the multiple sharpness feature method to identify the blur area on 
the WSI. Using different blur features (such as features related to the co-occurrence matrix 
and image gradient), the best blur detection results are obtained by using a DT classifier 
(i.e., $98.56\%$ and $96.63\%$ classification accuracy and small hardware investment).

In~\cite{Lopez-2012-CMAD}, Ki67 immunohistochemistry-based WSI technology is used to detect tumor 
areas with high proliferation activity, a hybrid clustering method, referred to as Seedlink, is 
developed in the paper. This tool greatly improves the pathologist’s identification of hot-spots 
consistency.

In~\cite{Veta-2013-DMFB}, the author uses the level set method to extract candidate objects, and use two 
sets of features (the baseline set describing the size, shape, and color of the extracted candidate 
objects and the extension of texture information in addition to the baseline set) input into 
the linear discriminator, divide candidate objects into mitotic figures or fake objects to detect mitosis.

In~\cite{Shirinifard-2016-DPAU}, the staining and texture characteristics of Ki67 stained 
glass slides is used to divide the tissues in WSI into living tumor tissue, necrotic tissue, and background. 
And use a tissue classifier based on machine learning, which is trained five times magnified 
images to be applied to living tumor tissue to detect phenotypic changes. In the end, an accuracy 
of $95\%$ in each area is achieved.

We found three papers about non-machine learning. In~\cite{Bautista-2009-DTFW}, by introducing a 
movement operator, it can effectively amplify the chromaticity difference between tissue folds and 
other tissue components to detect tissue folds in WSI. In~\cite{Bautista-2010-IVDT}, similar 
to~\cite{Bautista-2009-DTFW}, the weighted difference between the color saturation and brightness 
of the image pixels is used as the offset factor of the original RGB color of the image to enlarge 
tissue folds.

In~\cite{Vyas-2016-CWSD}, the author compares the histopathological characteristics of dermatitis cases detected 
by WSI and traditional microscopy. Although WSI is not as effective as traditional microscopes, it 
is sufficient to examine the pathological features often encounter in dermatitis cases and get the 
effect that can be used.

\subsection{Deep Learning based Detection Methods}
In this section, the relevant content of using deep learning algorithms to detect histopathological 
images with WSI is briefly summarized.

\subsubsection{CNN-based Deep Learning Detection Method}
The following~\cite{Zanjani-2018-CDHW,Cruz-2017-ARIB,Bejnordi-2017-DADL,Tellez-2018-WSMD,Kohlberger-2019-WSIF,Jamaluddin-2017-TDWS,Cruz-2014-ADID} are all detection classifiers based on CNN. 
In~\cite{Zanjani-2018-CDHW}, CRF is applied to the latent space of the trained deep CNN, and the 
compact features extracted from the middle layer of the CNN are regarded as observations in the 
fully connected CRF model to detect invasive breast cancer. Experiments show that the average FROC 
score of tumor area detection in histopathology WSIs increased by about $3.9\%$. The proposed model 
is trained on the Camelyon17 ISBI challenge dataset and won second place with a kappa score of 0.8759.

In~\cite{Cruz-2017-ARIB}, involved images from five different cohorts from different 
institutions/ pathologylabs in the United States of America and TCGA. The training dataset 
had 349 estrogen receptor-positive (ER+) invasive breast cancer patients. The approach 
yielded a Dice-coefficient of $75.86\%$, a positive predictive value of $71.62\%$ and a negative 
predictive value of $96.77\%$ in terms of pixel-by-pixel evaluation compared to manually annotated 
regions of IDC.

In~\cite{Bejnordi-2017-DADL}, the author evaluates the performance of automated deep learning algorithms in detecting metastasis in lymph node H\&E tissue sections of women with breast cancer and compares it with a pathologist's diagnosis in a diagnostic environment. The dataset is collected from 
399 patients who underwent breast cancer surgery at the Radboud University Medical Center (RUMC) 
and Utrecht University Medical Center (UMCU) in the Netherlands. The algorithm in the paper is 
significantly better than the pathologist's artificial algorithm.

In~\cite{Tellez-2018-WSMD}, a method for training and evaluating CNNs in breast cancer WSIs 
mitosis detection is proposed. In the three tasks challenged by Tupac, the performance of the 
proposed method is independently evaluated.

In~\cite{Kohlberger-2019-WSIF}, an autofocus quality detector called ConFocus is developed to 
detect and quantify the out-of-focus area and severity on WSIs. When compared to pathologist-graded 
focus quality, ConvFocus achieved Spearman rank coefficients of 0.81 and 0.94 on two scanners and 
reproduced the expected OOF patterns from Z-stack scanning. The article also evaluates the impact 
of OOF on the accuracy of the most advanced metastatic breast cancer detector, and finds that as 
OOF increases, the performance continues to decline.

\begin{figure}[htbp!]
\centerline{\includegraphics[width=0.98\linewidth]{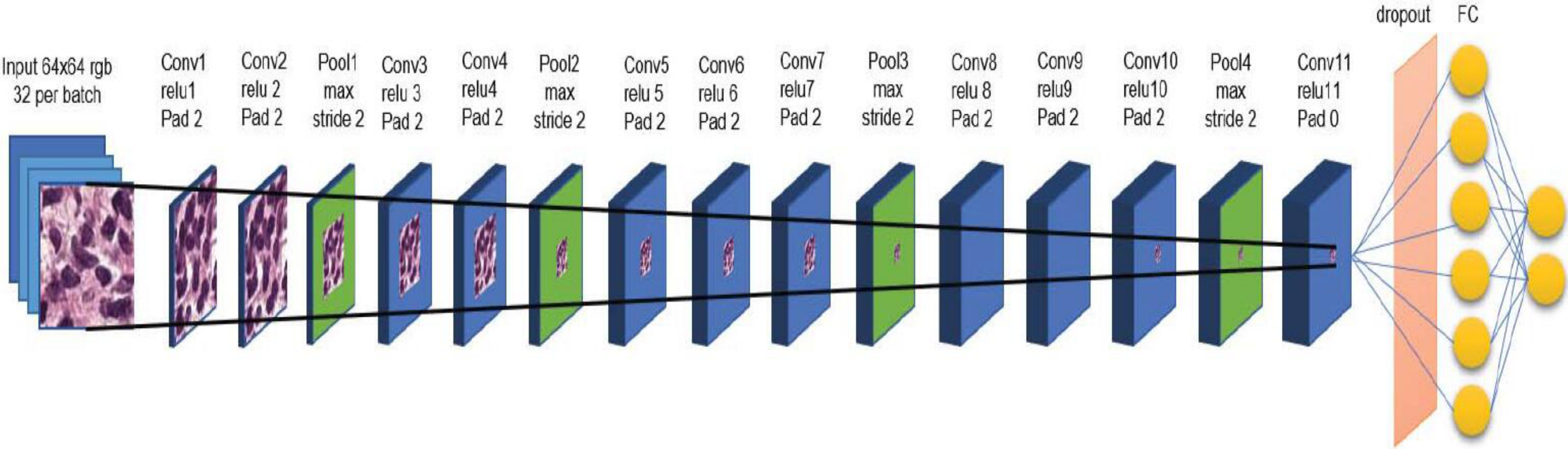}}
\caption{12 convolutional layer including the fully connected layer. 
This model was also inspired from VGG model.}
\label{fig:fig9}
\end{figure}

\cite{Jamaluddin-2017-TDWS} is divided into two parts. The first part is to use CNN to detect 
possible tumor locations in WSI, and the second part is to use the detected results to extract 
features to classify normal or tumors. Using Camelyon16 dataset, which consists of 160 negatives 
and 110 positives WSIs for training, and 50 positives and 80 negatives for testing. The method 
has a better AUC result at 0.94 than the winner of Camelyon16 Challenge with an AUC of 0.925. The 
CNN structure designed in ~\cite{Jamaluddin-2017-TDWS} is shown in Figure.~\ref{fig:fig9}.

\begin{figure}[htbp!]
\centerline{\includegraphics[width=0.98\linewidth]{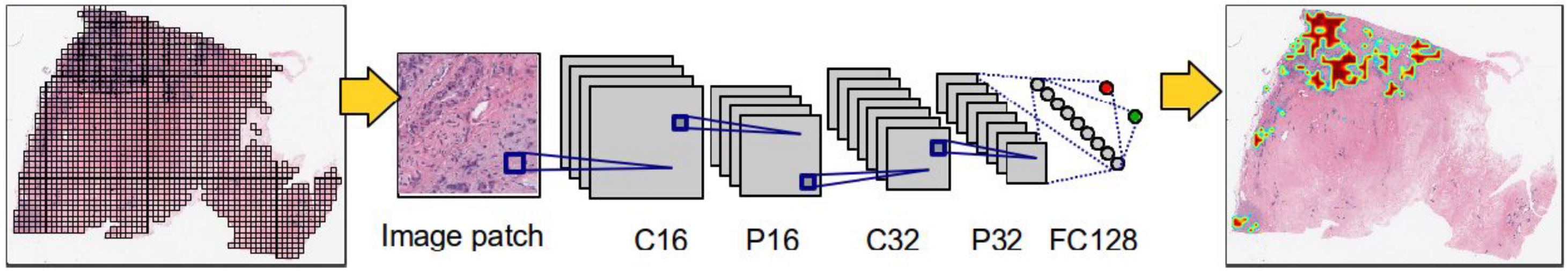}}
\caption{Overall detection framework of ~\cite{Cruz-2014-ADID}.}
\label{fig:fig10}
\end{figure}

In~\cite{Cruz-2014-ADID}, CNN is trained to detect IDC in WSI.  At the end, $71.80\%$ of 
F-measure (F1) and $84.23\%$ of balanced accuracy are obtained. The overall detection framework 
of~\cite{Cruz-2014-ADID} is shown in Figure.~\ref{fig:fig10}.

\subsubsection{FCN-based deep learning detection method}
The following~\cite{Geccer-2016-DCBC,Gecer-2018-DCCW,Lin-2018-SFDS,Lin-2019-FSFD} are processed 
with 4-layers of FCN to achieve the purpose of detection.

The study in ~\cite{Geccer-2016-DCBC} is composed of two tasks, retrieval and classification. The retrieval 
part is composed of four layers of FCN. Through the feedforward processing of FCN-1, the salient 
area is detected from WSI, and each connected component above the threshold is enlarged on the 
input image and processed by FCN-2. This process lasts four times, and a significant area of WSI 
can be detected. Then CNN classifies and finally determines the breast cancer diagnosis results.

In~\cite{Gecer-2018-DCCW}, a system is proposed for diagnosing WSI of breast biopsy. Firstly, 
four FCNs are used for saliency detection and multi-scale localization of ROI. Then, convolutional network is used for cancer classification. Finally, the saliency map and the classification 
map are combined. The test result shows that its accuracy rate is roughly indistinguishable from 
the pathologist's prediction.

In~\cite{Lin-2018-SFDS}, an improved FCN layer is used to input the WSI of any size, and the standard FCN layers are converted into the anchor layer. Fast and dense ScanNet inference 
with the anchor layer makes the network faster. The result of its detection of cancer metastasis 
shows excellent performance on the Camelyon16 Challenge dataset.Similar method is applied in ~\cite{Lin-2018-SFDS} and ~\cite{Lin-2019-FSFD} 
.

\subsubsection{Other deep learning detection methods}
The following references~\cite{Cruz-2018-HTAS,Sirinukunwattana-2016-LSDL,Bilaloglu-2019-EPCW} are used for 
detection after combining or simplifying the CNN structure with other methods.

In~\cite{Cruz-2018-HTAS}, a new type of efficient adaptive sampling based on probability 
gradient and quasi-Monte Carlo sampling is used, combined with a CNN-based classifier. Applied 
to the detection of invasive breast cancer on WSI. The experimental result shows that the 
Dice-coefficient is $76\%$, which is an efficient strategy.

In~\cite{Sirinukunwattana-2016-LSDL}, in conventional colon cancer WSI proposes a space-constrained 
CNN for nuclear detection. The new NEP is used in conjunction with CNN to more accurately predict 
the type of cell detection. The test result shows that this article produces an higher average F1 score of detection.

In~\cite{Bilaloglu-2019-EPCW}, a simplified CNN, named PathCNN, is used to detect outliers of WSI 
in whole cancer. The WSI used in the experiment is downloaded from the genome data sharing database 
of the TCGA.

In~\cite{Liu-2019-AIBC}, Lymph Node Assistant (LYNA), a tool for breast cancer lymph node metastasis 
detection based on Inception-v3, is evaluated for its application and clinical practice. The data is obtained from two sources: the Camelyon16 challenge containing 399 slides, and a 
private dataset containing 108 slides from 20 patients 
(86 tissue blocks). When applying the second dataset, LYNA achieved an AUC of $99.6\%$.

\subsection{Ensemble Learning based Detection Method}
This section provides a summary of the contents related to the use of WSIs by the ensemble 
learning algorithm to detect histopathological images.

In~\cite{Huang-2017-AHGP}, the image is sampled based on the kernel density, and the 
taint deconvolution and feature description are used to extract image features. Then, an enhanced version of the rank boost integrated method (using multiple weak classifiers to obtain better performance of the final rank) is utilized to rank and detect high-level prostate cancer. The experiment shows the mean AUC 
is $0.9486 \pm 0.005$ and the mean accuracy achieves $95.57\% \pm 2.1\%$. 

In the first step in~\cite{Doyle-2010-BBMC}, WSIs are decomposed into an image pyramid containing multiple resolution levels. At a lower resolution level, the Bayesian classifier identified the infected areas like cancer, and then a higher resolution level is used for more detailed examination. At each resolution level, 
the AdaBoost ensemble method is used to collect more than 900 first-order statistics, ten image features 
are selected from the second-order co-occurrence and Gaborfilter feature pool, and an enhanced Bayesian 
multi-resolution (BBMR) system is used to detect the coronary artery lesion area on the digital 
biopsy slice.

\subsection{Summary}
Among the papers we have summarized, it can be divided into traditional detection 
methods, detection methods based on deep learning, and detection methods based on ensemble learning.

There are traditional machine learning algorithms and some non-machine learning algorithms 
in traditional detection methods. Among the traditional detection method, six 
articles~\cite{Jiao-2013-CCDU,Sharma-2015-ABND,Han-2018-ACDL,Simon-2018-MRLF,Swiderska-2015-TMMH,Nguyen-2011-PCDF} 
use SVM methods and five 
articles~\cite{Valkonen-2017-MDWS,Litjens-2015-ADPC,Bejnordi-2015-MSSC,Bejnordi-2016-ADDW,Li-2015-FRID} 
use RF, DT, clustering, linear discriminator, and organization classifier algorithms. Two non-machine learning papers~\cite{Bautista-2009-DTFW,Bautista-2010-IVDT} construct a color transfer 
factor to enhance the image folds for detection. There are 15 
papers~\cite{Zanjani-2018-CDHW,Cruz-2017-ARIB,Bejnordi-2017-DADL,Tellez-2018-WSMD,Kohlberger-2019-WSIF,Jamaluddin-2017-TDWS,Cruz-2014-ADID,Cruz-2018-HTAS,Sirinukunwattana-2016-LSDL,Liu-2019-AIBC,Geccer-2016-DCBC,Gecer-2018-DCCW,Lin-2018-SFDS,Lin-2019-FSFD} 
related to detection based on deep learning, most of which use CNN. In ~\cite{Huang-2017-AHGP,Doyle-2010-BBMC}, an ensemble learning based detection, 
one uses RankBoost and the other uses AdaBoost. It can be seen from the summary paper that before 2010, 
traditional non-machine learning algorithms are used for detection. Since 2011, machine learning has 
entered the public arena, and methods such as SVM and RF have been widely used. Until 2016, the 
emergence of deep learning enables better results. From the 
detection results, we can also see that CNN obtain high accuracy. Table.~\ref{SCAD} is 
a summary of the CAD methods used for detection in WSI. 
\onecolumn
\begin{center}\tiny
\renewcommand\arraystretch{1.85}
\setlength{\tabcolsep}{0.001 pt}
\newcommand{\tabincell}[2]{\begin{tabular}{@{}#1@{}}#2\end{tabular}}
\centering
\begin{longtable}{cclclccc}
\caption{Summary of the CAD methods used for detection in WSI (Traditional (T), Deep learning (DL), Ensemble Learning (EL), Breast Cancer Surveillance Consortium (BCSC)).} \\
\endfirsthead
\caption[l]{Continue: Summary of the CAD methods used for detection in WSI.}\\
\hline
\endhead
 \hline
\endfoot
\hline
Type & Reference                          & \multicolumn{1}{c}{Year} & \multicolumn{2}{c}{Team}                      & Data                                                                   & \multicolumn{2}{c}{Details}                                                             \\ \hline
T    & ~\cite{Nguyen-2011-PCDF}           & 2011                     & \multicolumn{2}{c}{Kien Nguyen}               &                                                                    $\backslash$  & SVM with RBF kernel and c $=$ 1                                      \\
T    & ~\cite{Jiao-2013-CCDU}             & 2013                     & \multicolumn{2}{c}{Liping Jiao}               &                                                                       $\backslash$ & SVM                                                                \\
T    & ~\cite{Sharma-2015-ABND}           & 2015                     & \multicolumn{2}{c}{Harshita Sharma}           &                                                                       $\backslash$ & SVM                                                                \\
T    & ~\cite{Swiderska-2015-TMMH}        & 2015                     & \multicolumn{2}{c}{Zaneta Swiderska}          &                                                                       $\backslash$ & SVM with Gaussian kernel function                                 \\
T    & ~\cite{Han-2018-ACDL}              & 2018                     & \multicolumn{2}{c}{W. Han}                    &                                                                       $\backslash$ & SVM                                                                \\
T    & ~\cite{Simon-2018-MRLF}            & 2018                     & \multicolumn{2}{c}{Olivier Simon}             &                                                                       $\backslash$ & SVM                                                                \\
T    & ~\cite{Litjens-2015-ADPC}          & 2015                     & \multicolumn{2}{c}{Litjens, G}                &                                                                       $\backslash$ & RF                                                                 \\
T    & ~\cite{Li-2015-FRID}               & 2015                     & \multicolumn{2}{c}{Ruoyu Li}                  & NLST                                                                   & RF                                                                 \\
T    & ~\cite{Bejnordi-2015-MSSC}         & 2015                     & \multicolumn{2}{c}{Bejnordi, B. E}            &                                                                       $\backslash$ & RF                                                                 \\
T    & ~\cite{Bejnordi-2016-ADDW}         & 2016                     & \multicolumn{2}{c}{Bejnordi, B. E}            &                                                                       $\backslash$ & RF                                                                 \\
T    & ~\cite{Valkonen-2017-MDWS}         & 2017                     & \multicolumn{2}{c}{Valkonen M}                &                                                                       $\backslash$ & RF                                                                 \\
T    & ~\cite{Bautista-2009-DTFW}         & 2009                     & \multicolumn{2}{c}{PA Bautista}               & Massachusetts General Hospital                                         & Color shift factor                           \\
T    & ~\cite{Bautista-2010-IVDT}         & 2010                     & \multicolumn{2}{c}{PA Bautista}               & Massachusetts  General   Hospital                                      & Color shift factor                           \\
T    & ~\cite{Veta-2013-DMFB}             & 2012                     & \multicolumn{2}{c}{M. Veta}                   & UMCU                                   & Linear discriminator                                               \\
T    & ~\cite{Lopez-2012-CMAD}            & 2012                     & \multicolumn{2}{c}{Lopez X M}                 &                                                                       $\backslash$ & A hybrid clustering method                                         \\
T    & ~\cite{Lopez-2013-ABDM}            & 2013                     & \multicolumn{2}{c}{Lopez X M}                 &                                                                       $\backslash$ & DT                                                                 \\
T    & ~\cite{Shirinifard-2016-DPAU}      & 2016                     & \multicolumn{2}{c}{Shirinifard A}             &                                                                       $\backslash$ & A tissue classifier based on machine learning          \\
T    & ~\cite{Vyas-2016-CWSD}             & 2016                     & \multicolumn{2}{c}{Vyas N S}                  &                                                                       $\backslash$ & Compare WSI and traditional microscopy                            \\
DL   & ~\cite{Cruz-2014-ADID}             & 2014                     & \multicolumn{2}{c}{Angel Cruz-Roa}            &                                                                       $\backslash$ & CNN                                                                \\
DL   & ~\cite{Jamaluddin-2017-TDWS}       & 2017                     & \multicolumn{2}{c}{Jamaluddin M F}  & Camelyon16  & CNN                                                                \\
DL   & ~\cite{Bejnordi-2017-DADL}         & 2017                     & \multicolumn{2}{c}{Bejnordi, B. E}            & RUMC and UMCU                                                          & CNN                                                                \\
DL   & ~\cite{Cruz-2017-ARIB}             & 2017                     & \multicolumn{2}{c}{Angel Cruz-Roa}            & TCGA                                                                   & CNN                                                                \\
DL   & ~\cite{Zanjani-2018-CDHW}          & 2018                     & \multicolumn{2}{c}{Farhad Ghazvinian Zanjani} & Camelyon17  & CNN                                                                \\
DL   & ~\cite{Tellez-2018-WSMD}           & 2018                     & \multicolumn{2}{c}{David Tellez}              & TNBC,TUPAC                                                             & CNN                                                                \\
DL   & ~\cite{Kohlberger-2019-WSIF}       & 2019                     & \multicolumn{2}{c}{T. Kohlberger}             &                                                                       $\backslash$ & CNN                                                                \\
DL   & ~\cite{Geccer-2016-DCBC}           & 2016                     & \multicolumn{2}{c}{Geçer B}                   &                                                                       $\backslash$ & FCN                                                                \\
DL   & ~\cite{Gecer-2018-DCCW}            & 2018                     & \multicolumn{2}{c}{Geçer B}                   & \tabincell{c}{registries associated with \\the BCSC Consortium} & FCN                                                                \\
DL   & ~\cite{Lin-2018-SFDS}              & 2018                     & \multicolumn{2}{c}{Huangjing Lin}             & Camelyon16  & FCN                                                                \\
DL   & ~\cite{Lin-2019-FSFD}              & 2019                     & \multicolumn{2}{c}{Huangjing Lin}             & Camelyon16                                                      & FCN                                                                \\
DL   & ~\cite{Sirinukunwattana-2016-LSDL} & 2016                     & \multicolumn{2}{c}{Sirinukunwattana K}        &                                                                       $\backslash$ & \tabincell{c}{New NEP is used in \\conjunction with CNN} \\
DL   & ~\cite{Cruz-2018-HTAS}             & 2018                     & \multicolumn{2}{c}{Angel Cruz-Roa}            & HUP,CWRU,CINJ,TCGA                                                     & CNN and adaptive sampling                                          \\
DL   & ~\cite{Bilaloglu-2019-EPCW}        & 2019                     & \multicolumn{2}{c}{S Bilaloglu}               & TCGA                                                                   & PathCNN \\
DL   & ~\cite{Liu-2019-AIBC}              & 2019                     & \multicolumn{2}{c}{Liu Y}                     & \tabincell{c}{Camelyon16, \\a separate dataset}                        & LYNA (Based on Inception-v3) \\
EL   & ~\cite{Doyle-2010-BBMC}            & 2010                     & \multicolumn{2}{c}{S. Doyle}                  &                                                                       $\backslash$ & AdaBoost \\
EL   & ~\cite{Huang-2017-AHGP}            & 2017                     & \multicolumn{2}{c}{Huang C H}                 & TCGA                                                                   & Rankboost                                                          \\ 
\label{SCAD}\\
\end{longtable}
\end{center}

\section{Methodology Analysis}
\label{s:MA}
This section analyzes prominent methods in different tasks.

\subsection{Analysis of Segmentation Applications in WSI}
\begin{figure}[htbp!]
\centerline{\includegraphics[width=0.98\linewidth]{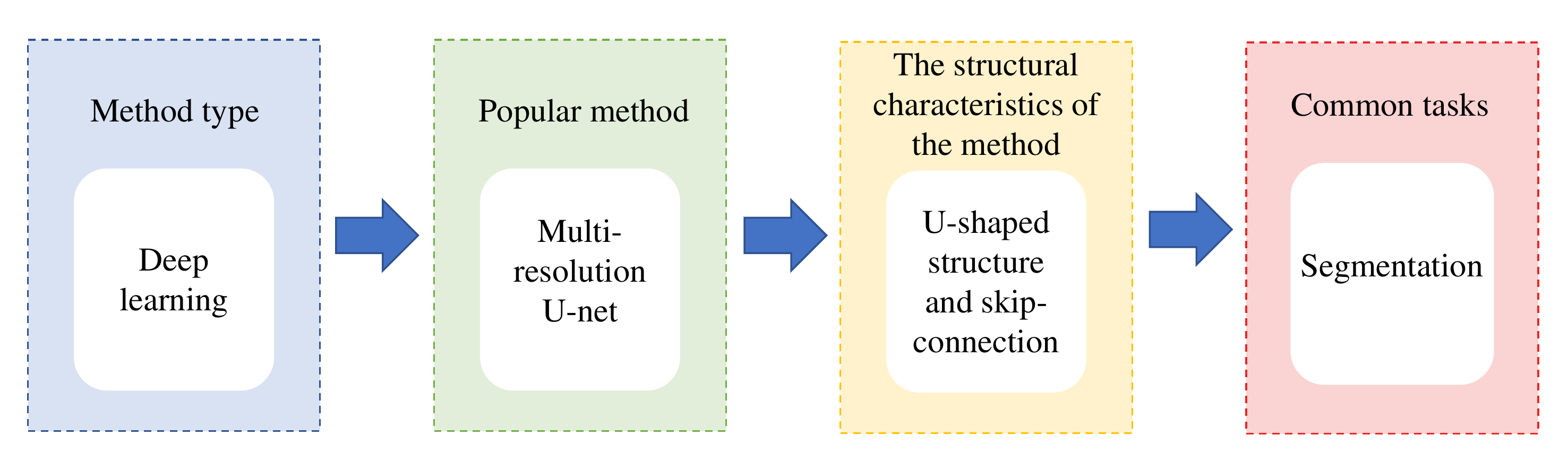}}
\caption{The popular methods for segmentation in WSIs.}
\label{fig:Methodology-as}
\end{figure}

In recent years, medical WSIs have been used for segmentation to assist doctors in their diagnosis 
and treatment. Most of them use deep learning algorithms. Because of the different datasets used in each effort, it is impossible to evaluate each segmentation method's effectiveness vertically. 
The use of multi-resolution U-net architecture is undoubtedly a common one, as shown in 
Figure.~\ref{fig:Methodology-as}.

Its U-shaped structure and its skip-connection are its structural characteristics. The U-shaped 
structure can be used to extract deep features by down-sampling and down-dimension-reduction, 
and then up-sampling to obtain more accurate output images. This kind of end-to-end network can 
achieve good results in medical image segmentation. There is another multi-resolution encoder-decoder network for breast cancer segmentation~\cite{Mehta-2018-LSBB}, 
which is similar to the U-net method. However, U-net structure can only be predicted on a single 
scale, so it cannot cope with scale changes well. Moreover, training does not have good generalization 
ability when the convolutional layer is increased~\cite{Du-2020-MISU}. Breast cancer is the most commonly used~\cite{Dong-2018-RANT, Mehta-2018-LSBB, Apou-2014-FSTC, Veta-2013-ANSH, Veta-2012-PVAE, Roullier-2010-GMRS, Roullier-2010-MEBC, Seth-2019-ALBD, Seth-2019-ASDW} 
tasks in segmenting WSI to assist pathologists in diagnosis.

\subsection{Analysis of Classification and Detection Applications in WSI}
According to the review of classification applications of WSI in CAD, it can be seen that SVM is 
the most frequently used technique in traditional machine learning. However, basic SVM cannot achieve excellent classification results. For example, in~\cite{Nayak-2013-CTHS}, multi-class 
regularized SVM is used to classify tumors, and the accuracy rate is only $84\%$. 
In~\cite{Peikari-2015-TDRR}, the RBF SVM is used for classification, and only 0.87 AUC is obtained. 
However, if SVM is combined with SVM of other kernel functions or other kinds of classifiers to 
train ensemble learning classifiers, the results will be highly improved. For example, 
in~\cite{Difranco-2011-ESWS}, the AUC of prostate cancer is 0.95 when it is classified. However, 
this ensemble learning method has low efficiency, slow running speed, and requires a large number 
of parameters. For some more complex tasks, a large number of calculations are needed~\cite{Weihua-2013-RCMA}.

In the traditional machine learning algorithm, the Gaussian pyramid method is also commonly combined 
with $k$NN  to improve the classification accuracy. The Gaussian pyramid is based on simple 
down-sampling plus Gaussian filtering. The original image is continuously sampled by decreasing order, 
and a series of images of different sizes are obtained, from large to small, from bottom to top, to 
form a tower model. In this way, images of different resolutions can be obtained, thus improving the 
accuracy of classification. For example, in~\cite{Sertel-2008-CAPN}, the improved $k$NN combined with 
the Gaussian pyramid is used to classify neuroblastoma, and the classification accuracy of $95\%$ is 
obtained. However, in the classification problem of unbalanced datasets, the defects of $k$NN are obvious. 
Due to the influence of sample distribution, the minority class will be more biased towards the majority 
class discrimination~\cite{Li-2011-IKNN}.

\begin{figure}[htbp!]
\centerline{\includegraphics[width=0.98\linewidth]{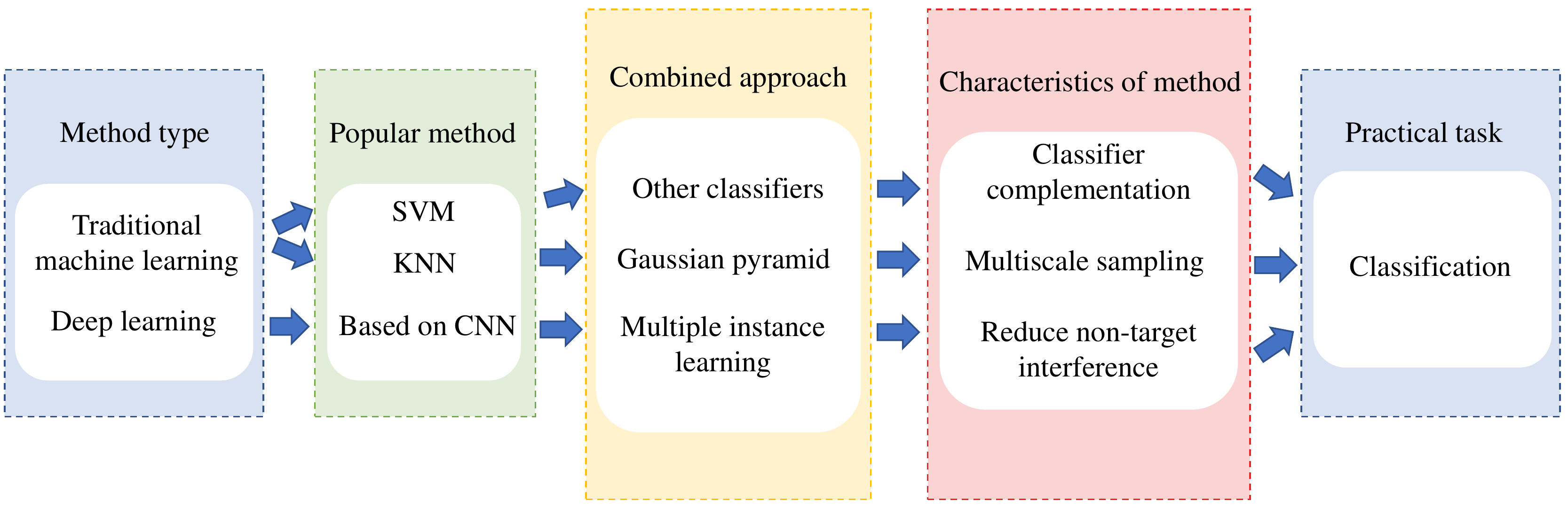}}
\caption{The popular methods in classification for WSIs.}
\label{fig:Methodology-ac}
\end{figure}

In the deep learning classification algorithm, most of them are based on CNN. CNN is better than 
traditional machine learning methods in processing high-dimensional data, and because of the 
convolutional layer, it can automatically extract features for learning. In recent years, MIL and 
neural networks are often used to carry out classification tasks for medical WSIs. For 
example~\cite{Mercan-2017-MIML, Das-2018-MILD, Campanella-2018-TSDM, Campanella-2019-CGCP, Wang-2019-RRMI}. 
MIL is a learning problem with Multiple example packages as training units. A bag is marked as a 
positive-class multi-example package if it contains at least one positive instance. Conversely, 
the negative example is also true. The multi-example learning method can effectively reduce the 
noise and improve the classification accuracy of the prediction. 
In~\cite{Campanella-2018-TSDM, Campanella-2019-CGCP}, the AUC of its classification is 0.98, a good 
result. However, the CNN-based neural network is an end-to-end architecture, similar to a black box, which 
is weak in interpretation~\cite{Zhang-2013-CML}. The popular methods in classification for WSIs are 
as shown in Figure.~\ref{fig:Methodology-ac}.

In the task of WSIs classification, the classification of 
breast cancer~\cite{Peikari-2015-TDRR, Wang-2016-DLIM, Geccer-2016-DCBC,Araujo-2017-CBCH, Bejnordi-2017-CASC, Gecer-2018-DCCW, Kwok-2018-MCBC, Das-2018-MILD}, 
prostate cancer~\cite{Difranco-2011-ESWS, Doyle-2010-BBMC, Li-2019-AMRM, Weingant-2015-EPTC, Ren-2018-ADAC}, 
and colon cancer~\cite{Jiao-2013-CCDU, Sirinukunwattana-2016-LSDL, Morkunas-2018-MLBC, Korbar-2017-DLCC} 
is commonly used. 
Gastric cancer~\cite{Sharma-2016-DCNN, Sharma-2017-DCNN, Shou-2018-WSIC}, 
neuroblastoma~\cite{Sertel-2008-CAPN, Kong-2009-CAEN}, 
and melanocytic tumor on skin ~\cite{Lu-2015-AADS, Xu-2018-AACM} have also been studied.

There are many researches on medical WSIs detection that are related to classification. For example, 
\cite{Geccer-2016-DCBC, Sirinukunwattana-2016-LSDL, Jamaluddin-2017-TDWS, Gecer-2018-DCCW}, they are 
all detected first and then classified. Most of the tasks that use WSIs alone as an adjunct to 
treatment are for breast cancer 
detection~\cite{Cruz-2017-ARIB, Cruz-2018-HTAS, Valkonen-2017-MDWS, Lin-2018-SFDS, Tellez-2018-WSMD, Zanjani-2018-CDHW, Veta-2013-DMFB, Liu-2019-AIBC}.

\subsection{Potential Methods for WSI}
In this section, we discuss some of the potential approaches for WSI technology.

\subsubsection{Potential Methods for Feature Extraction Applied to WSI Technology}
In addition to the feature extraction methods we reviewed, there are some other feature extraction 
methods that grab our attention and can be used in WSI technology.

In~\cite{Wang-2020-AEMS}, a method is proposed to extract the structural features of building 
facades through texture fusion. After texture fusion, the gradient amplitude of elements reduced, 
and the gradient amplitude of structural features can be kept constant. The interference of texture 
to structural feature extraction can be eliminated by this method of texture fusion. If we apply 
this method to the extraction of traditional features, it may improve the availability of features 
and get better results.

In~\cite{Tougaccar-2020-DLCC}, LeNet, AlexNet, and VGG-16 based deep learning models are used to realize 
the detection of lung cancer. This experiment is applied to the Computed Tomography (CT) image dataset. 
The combination of the AlexNet model and $k$NN classifier is used to obtain the best accuracy of $98.74\%$. 
Then, the minimum redundancy and maximum correlation feature selection method proposed in this paper 
is applied to the deep learning features, and the pruning operation is carried out to select the most 
effective features. This method improves accuracy to $99.51\%$. Next, we can try to apply this approach 
to WSI datasets as well.

In~\cite{Zheng-2017-FEHI}, a CNN-based nuclear guided feature extraction framework is proposed to do the classification. The detected nucleus can guide the training of neural network and reduce the noise caused 
by matrix. Feature extraction (including image-level features of nuclear pattern and spatial distribution) 
is carried out by the neural networks trained by nucleus guidance. The framework can also be used for 
automatic WSI analysis and can be extended to other different cancer types.

In~\cite{Zhao-2016-IQAC}, an image quality evaluation method is proposed, which is based on log-Gabor 
wavelet features and texture information description. The statistical local representation of these 
two complementary sources can quantify the differences in these extracted features between the distorted 
image and the reference image. We can apply this method to WSI datasets for data selection. This can 
reduce the impact of data on experimental results.

\subsubsection{Potential Methods for Segmentation Applied to WSI Technology}
There are some segmentation methods that are worth noting in other fields and can be tried 
with WSI for CAD.

In~\cite{Sun-2020-GHIS}, a method of gastric histopathological image segmentation based on the layered 
CRF is introduced. This method can automatically locate the cancer nest information in the stomach image, 
and because the CRF can represent the spatial relationship, a higher order term can be established on 
this basis and applied to the image based post-processing, which further improves the segmentation 
performance. The model shows high subdivision performance and effectiveness. In ~\cite{Zhang-2020-MCFE}, CRF is also applied to the segmentation of environmental microorganism Images. So we can also apply it 
to WSI technology.

In~\cite{Galvao-2020-ISDS}, an image segmentation method is presented that is faster than superpixel is proposed. 
It separates into dense and sparse methods. Then, a new intensive method can achieve superior boundary 
adherence by exploring alternative mid-level segmentation strategies are proposed. This method is a 
very effective hierarchical segmentation method. But in this case, it applied to natural images, and 
we can also try to apply it to WSI.

In~\cite{Jiang-2019-MMMI}, a multi-channel weighted region scalable fitting (M-WRSF) segmentation model 
for medical image segmentation is proposed. In this M-WRSF model, a new penalty term is introduced to 
improve the numerical stability and the time interval is increased to improve the iteration efficiency. 
The new edge detection function is used to improve the segmentation performance. Based on the original 
model, the  Gaussian kernel function is added to enhance the robustness.

In~\cite{Cui-2020-FSCA}, a method of automatic segmentation of coronary artery based on growing 
algorithm is proposed. Firstly, 2D U-net is used to automatically locate the initial seed points and 
the growth strategy, and then a growth algorithm combined with the 3D triangulation network is proposed. 
The improved 3D U-net is used for coronary artery segmentation. This method adopts residual block and 
two-phase training. The input data of the network is set as the neighborhood block of the seed point. 
And according to the Iterative termination condition, it determines whether the segmentation is stopped.

\subsubsection{Potential Methods for Classification Applied to WSI Technology}
The following is an introduction to some classification methods used in other fields, which can be used 
in WSI for CAD.

In~\cite{Mu-2020-HICA}, a spectral-spatial classification algorithm based on spectral-spatial feature 
fusion of spatial coordinates is proposed to classify hyperspectral images. Active learning is introduced 
to improve performance. The method of combining spectral information with spatial information can solve 
the noise interference.

In~\cite{Devi-2020-NPSC}, two new Privacy Supporting Binary Classifier Systems are proposed to classify 
the Magnetic Resonance Imaging (MRI) images of the brain. LSB Substitution Steganographic method is used 
to protect the privacy of the patient. We can apply this approach to histopathological WSI. 
\cite{Okwuashi-2020-DSVM} also classifies hyperspectral images. Deep SVM is used, and the results obtained 
by this method are better than those obtained by other classifiers. Hopefully, this method can be extended 
to WSI datasets.

In~\cite{Li-2019-SACM}, a Content-Based Microscopic Image Analysis (CBMIA) approaches are proposed 
to classify microscopic images of microorganisms. This method is based on the computer semi-automatic 
or automatic method, so it is very effective and saves manpower and material resources. We can also 
apply this approach to medicine.

In~\cite{Manyala-2019-CGCN}, a CNN-based gender classification method for near-infrared periocular 
images is proposed. In other words, the neural network is used to extract features and SVM is used 
for classification. In other words, directly using neural network classification to achieve advanced performance.

\subsubsection{Potential Detection Methods on WSI Technology}
There are a number of tests that are used in other areas that can be attempted for WSI for CAD.

In~\cite{Park-2020-DASO}, the target in the video is detected. The method used is based on 
traditional background subtraction and artificial intelligence detection. Mask R-CNN is used to 
judge whether there is a segmentation object in the candidate region and to segment it.

In~\cite{Mukherjee-2020-SCAT}, a soft-computing based approach for automatic detection of pulmonary 
nodules is proposed. This method is applied to the CT images. Firstly, threshold processing, 
gray-scale morphology, and other preprocessing are used, and then random undersampling is used 
to deal with unbalanced problems. Then, a combination of particle swarm optimization (PSO) and 
stacking integration is used to detect.

In~\cite{Mohammadi-2020-CCAG}, to deal with the complex scene of the target, a Feature Guide Network 
is proposed. The multi-scale feature extraction module (MFEM) is used to obtain multi-scale context information for each level of abstraction. Finally, a loss function that outperforms the widely used 
cross-entropy loss is designed. This method does not require pretreatment and is efficient. 
\cite{Singh-2019-SBNF} is also used for significance detection, an artificial neural network regressor is 
trained to refine the significance map. If applied to the histopathological WSI for significance 
detection is very promising.

In~\cite{Hamadi-2020-USCM}, semantic context is used to carry out multiple concept detection of 
still images. The first is to generate semantic descriptors using a set of test scores for a single 
concept. This advanced feature is pushed as input to the target multi-concept detector. The second 
method detects the target multiple concepts and their categories, and then aggregates the results 
of the two treatments. Combining semantic context with the use of features based on deep learning 
yields good results.

\section{Conclusions and Future Work}
\label{s:CFW}
In this paper, image analysis methods based on machine learning using WSI technology for CAD are 
summarized. The applied datasets, evaluation methods, feature extraction, segmentation, classification, 
and detection in the task are analyzed and summarized. By reviewing all the related works, we can find 
that the most frequently-used datasets in Sect.~\ref{s:DEM},feature extraction  in Sect.~\ref{s:FE}, 
segmentation methods in Sect.~\ref{s:SM}, classification methods in Sect.~\ref{s:CM}, and detection 
methods in Sect.~\ref{ss:int:DM}, respectively.

Through the review of relevant work, we can find the commonly used methods of these three tasks. With time and the progress of science and technology, deep learning algorithm has gradually 
replaced the traditional machine learning algorithm.

TCGA~\cite{TCGA} and Camelyon~\cite{Litjens-2018-1HSS} are the two commonly used datasets in the 
common datasets summarized by us. In terms of feature extraction, color features, texture features, 
shape features, and deep learning features are the most commonly used. In the segmentation work, it 
separated into thresholding-based segmentation, region-based segmentation, graph-based segmentation, 
clustering-based segmentation, deep learning related segmentation and other methods. These traditional 
methods are simple to calculate, but sensitive to noise, so they are not robust. And the segmentation 
method based on U-net has become the mainstream in recent years. Classification work is the most studied. 
In the classification work, the combination of ensemble learning for the traditional classifier, MIL, 
and neural network has better recognition ability. Most of the testing work is carried out together with 
the classification work. In addition, the deep learning method based on CNN has achieved excellent 
performance in segmentation, classification, and detection tasks, which will contribute to the early 
detection, diagnosis, and treatment of patients.

In the future, the combination of WSI technology and machine learning to help pathologists assist in 
diagnosis is promising. In recent years, CAD research has mainly focused on the breast, stomach, colon, 
and nervous systems, etc., and the research field can be expanded to a wider extent in the future. 
Second, there is still a lack of large-scale, comprehensive, and fully annotated WSI datasets. Finally, 
it would be very useful to develop a network that requires less computation, requires less hardware and 
can be interpreted.

\begin{acknowledgements}
This work is supported by National Natural Science Foundation of China (No. 61806047). 
We thank Miss Zixian Li and Mr. Guoxian Li for their important discussion. 
We also thank B.E. Xiaoming Zhou, B.E. Jinghua Zhang and B.E. Jining Li, for their 
Important technical supports.
\end{acknowledgements}

%
 \section*{Conflict of Interest}
 The authors declare that they have no conflict of interest.

\bibliographystyle{spphys}       
\bibliography{lxt}   
\end{document}